\documentclass[buch,nomenklatur]{ita-report}

\author{Simon Ehlers}
\title{Traffic Queue Length and Pressure Estimation for Road Networks with Geometric Deep Learning Algorithms}
\subject{Project Thesis}
\date{\today}

\bearbeiter{cand.~mach.~Simon Ehlers, B.Sc.}
\matrikelnummer{}
\zeitraum{07.08.2018 bis 07.02.2019}
\betreuer{Prof. R. Horowitz, M. Wright, M.Sc, S. Sohrt, M.Sc.}
\pruefer{Prof.~Dr.-Ing. Ludger Overmeyer}

\DeclareSIUnit{\Bit}{Bit}

\begin{document}

%-------------------------------------------------------------------------------
% \vorspann erstellt aus obigen Angaben die Titelseite und die Erklärung, bindet 
% Aufgabenstellung (Datei aufgabe.pdf) und Abstract (Datei abstract.tex) ein und 
% erzeugt Inhalts-, Abbildungs- und Symbolverzeichnis
%-------------------------------------------------------------------------------
\vorspann
%\usepackage{float} %selbst eingebunden für Option [H] bei Figure
%-------------------------------------------------------------------------------
% Ab hier beginnt der eigentliche Text, der Inhalt. Empfehlung als Beispiel: 
% Kapitel mit zugehörigen Abbildungen in einzelnen Unterverzeichnissen sammeln 
% und an dieser Stelle per 
%
%    \include{verzeichnisname/dokument-ohne-endung}
%
% einbinden.
%-------------------------------------------------------------------------------

\chapter{Introduction}

%Here should be described why there is a need for queue length estimation (for traffic control) and what the challenges are. Describe the research questions and the goal of this work.  

The continuous trend of urbanization and the increase of vehicle sales world wide, cause in almost every large metropolitan region a high amount of congestion and inefficient traffic delays. Congestion causes an increase of the travel time for the passenger and a waste of working hours for the industry. For the year 2011 the extra travel hours in the US due to congestion is estimated to 5.5 billion \cite{schrank_ttis_2012}. Furthermore, the extra fuel consumption caused by inefficient traffic and congestion in the US  was for the year 2011 2.9 billion gallons of fuel (\$121 billion) \cite{schrank_ttis_2012}. The ecological effects regarding air pollution and the emit of greenhouse gases are even more alarming. According to \cite{kurzhanskiy_traffic_2015}, one of the main reasons for congestion is poor traffic control, mostly caused by a lack of collected traffic data. 
One solution to reduce congestion in road networks is to improve intelligent traffic control systems, which use for instance traffic light signals to control the traffic. These control algorithms rely on measurements of the traffic state. One such measured quantity is the queue length in front of signalized intersections. 

Unfortunately, a lot of today's freeways and urban road networks don't have the required sensors to implement a reliable control algorithm. The most common sensor is an inductive loop detector, which is hard to maintain and also limited in its reliability compared to camera systems for instance, \cite{kurzhanskiy_traffic_2015}. A problem is that, if loop detectors are available, it is challenging to estimate the congestion and queue length in every traffic situation only based on loop detector data. Loop-detectors can only measure passing vehicles. So they cannot provide reliable information in the situation where traffic is congested and the traffic flow is limited so that fewer vehicles pass the detector. 

This thesis addresses the problem of the trade-off between the usage of common loop-detectors and still a reliable and accurate estimation of the traffic behavior (in this case traffic queue length). Accurate measurements and estimates of the traffic state are required for traffic control, such as the adaptive traffic light control scheme of \cite{varaiya_max_2013}.
The main goal of this thesis is to present an algorithm which is able to estimate the queue length in front of intersections in road networks reliably and accurately by only using inductive loop detector data. Furthermore, it is required that the estimation method is also able to handle more complicated traffic scenarios like randomized traffic arrival, lane changing and stop-and-go traffic. 

Therefore, in this approach a geometric deep learning algorithm is presented, which uses loop detector data not only from one specific lane but also from its spatial neighbors in the road network to find correlations between lanes. This is possible due to graph attention (GAT) layers \cite{velickovic_graph_2017}, which uses spatial attention mechanisms between single lanes. Therefore, it is necessary to represent the topology of a road network as an abstracted graph consisting of nodes (lanes) and edges (intersections). Furthermore, an encoder-decoder structure \cite{bahdanau_neural_2014} is used, which is part of a subclass of methods concerning to "sequence to sequence" (seq2seq) problems, where the input as well as the output for the deep learning model is a sequence. The encoder maps the input sequence of detector data into hidden states. Another attention mechanism searches in these encoded hidden states for temporal attention between single time steps of the input sequence. Finally, the decoder maps the hidden states under attention of relevant previous time steps into the output sequence, which is the estimation of the queue length. Due to the spatial and temporal attention mechanism, which takes not only data from one specific lane but all neighboring lanes into consideration, the deep learning approach is potentially superior to conventional methods, which do not use spatial and temporal attention. So it is expected, that also more complex traffic situations as mentioned above can be handled by the deep learning model. As reference for the deep learning approach, a conventional queue length estimation method \cite{liu_real-time_2009} (Liu-method) is implemented as preliminary work, which uses second-by-second loop detector data to detect shockwaves in the traffic queue. The detected shockwaves are then used to calculate the queue length in front of signalized intersections. The Liu-method does not use spatial or temporal correlations, since it has only traffic data from one lane of only the last three traffic light cycles. Finally, the conventional method can be used as reference for the deep learning approach, but also as additional input information for the deep learning model to use prior knowledge of the traffic behavior to stabilize the estimation process.

\section{About this work}

%Present here the structure of the thesis and why you choose this.
At the beginning of this thesis in section \ref{sec.basics_traffic_data}, the basics of traffic data including the measurement and calculation of fundamental variables are introduced. These basics are applied to introduce the theoretical background necessary for the conventional queue length estimation method in section \ref{sec.liu_method}. Following, the fundamentals of artificial neural networks and geometric deep learning are presented in section \ref{sec.Fundamentals_GDL}.
Based on the theoretical background, the main scientific goal but also secondary goals are defined in chapter \ref{sec.scientific_goals}. 
In chapter \ref{sec.simulation_model} the traffic simulation software SUMO as well as the created road networks for the simulations are presented. 
The implementation of the conventional queue length estimation algorithms for a single test intersection but also for an entire grid-road-network is presented in chapter \ref{sec.Implementation_liu}. 
Following, in chapter \ref{sec.geometric_deep_learning} the implementation of the geometric deep learning approach is introduced. Furthermore, the results of both the conventional and the deep learning method are shown and compared. 
Finally, the conclusion regarding the deep learning approach is made in chapter \ref{sec.conclusion}.

\chapter{State of the Art} \label{sec.stateofart}

In this chapter the state of the art for real-time queue length estimation is presented. The focus is on queues that occur due to red traffic light signals in front of an intersection. For accurate estimation of the queue length, it is necessary to have reliable data about the traffic situation. Therefore, at the beginning of this chapter the fundamentals of measuring and processing of traffic data as well as the fundamental diagram, which describes the correlation between traffic flow and density, is introduced. 
Based on this fundamentals, a well known method for queue length estimation in research will be introduced \cite{liu_real-time_2009}. Since the method is based on shock waves in the queue, a brief introduction in the shock wave theory will be made. Afterwards the basic model, the expansion model and the specific modifications regarding this work is presented. 
Following, the fundamentals of geometric deep learning are introduced, starting with feedforward neural networks, continuing with graph attention layer and the RNN encoder-decoder structure. 

\section{Basics of Traffic Data} \label{sec.basics_traffic_data}

There are several ways of measuring road traffic data, such as radar sensors, light barriers, GPS data or cameras with image processing. However, the most common way to measure traffic data is by using inductive loop detectors \cite{kurzhanskiy_traffic_2015}. 
They are cheap and easy to install, as well as a data source which detects every vehicle that passes the detector. Based on their measurements and considering some assumptions, it is possible to calculate values which are important for traffic estimation, simulation and control. This is not true for every measuring system. For instance by using GPS data, only a certain group of people is taken into account (those who use a navigation service which is uploading the speed and position). So a detection of the whole traffic situation is not possible, because there is no standardized way to collect this data. Furthermore, GPS data cannot obtain necessary measurements like traffic flow and density, since there is no information about the travel direction for a particular location \cite{kurzhanskiy_traffic_2015}. 
%reasons against GPS data: -> not very precise, especially in cities with high buildings -> not everything can be solved by intelligent sself driving vehicles: see paper intelligent intersections 
% -> there is no standardized system for collecting and processing GPS data, yet
% but GPS data can be used to train spatial correlations for train neural networks

\subsection{Measuring and Processing of Loop Detector Data} \label{sec.loop_detector_data}

\begin{figure}[h]
	\centering
	\includegraphics[width=0.9\textwidth]{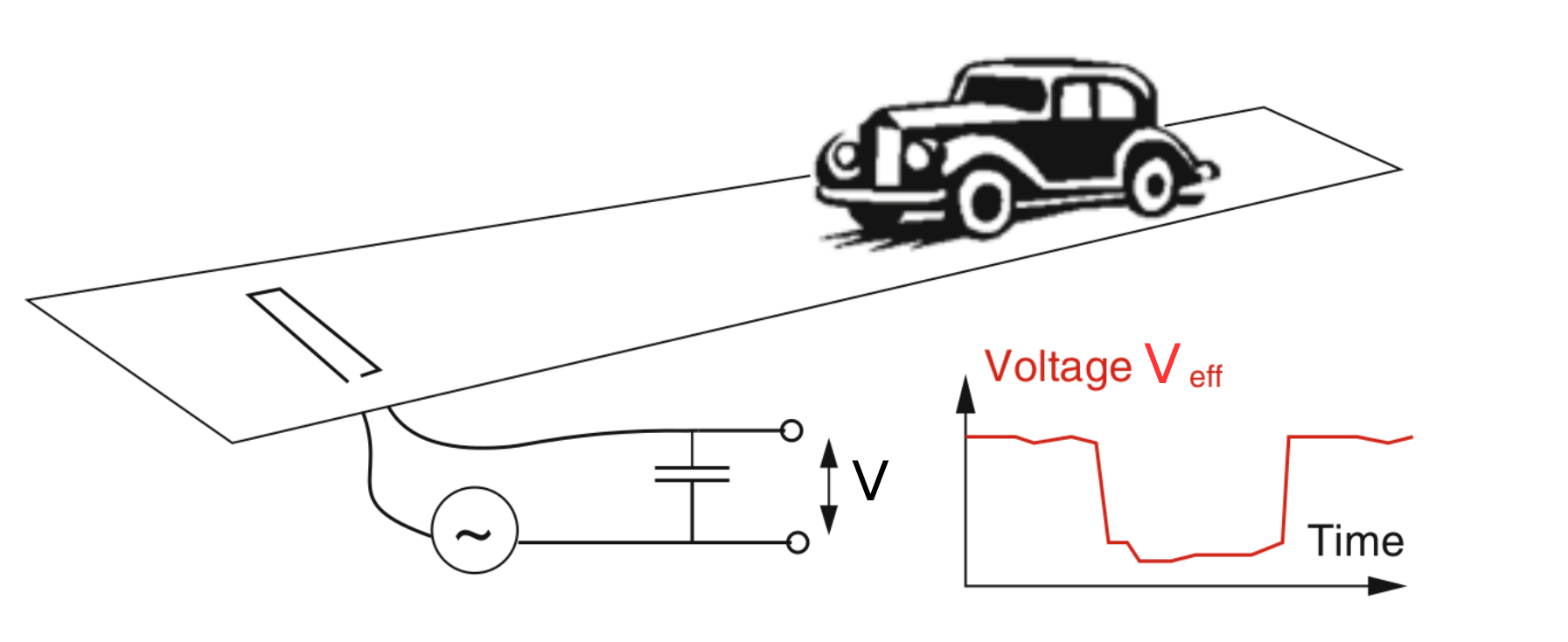}
	\caption[Principle of loop detectors]{Installation of a single loop detector beneath the street with LC circuit and voltage drop \cite{treiber_traffic_2013}.}\label{bild.principle_detector}
\end{figure}

The loop detector is based on the following principle: An induction loop is built under the street surface and included in an LC circuit with a capacitor and an AC voltage source. When the detector is unoccupied, the inductance of the loop is low resulting in a high voltage V. If the area above the detector is occupied by an metallic vehicle (truck, car, motorcycle or bike), the inductance of the loop is increased and voltage V drops \cite{treiber_traffic_2013}. The principle of this binary detector is shown in Fig. \ref{bild.principle_detector}.

One disadvantage of the single-loop detector (detector that consists  only of one inductive loop) is, that the vehicle speed is not directly measurable, since only a binary signal (occupied/unoccupied) over time is provided. The speed can be estimated by detecting the detector occupancy time $t_o$ (period of time between the front/ rear end of the vehicle passes the loop detector) and by assuming an effective vehicle length $L_e$ \cite{liu_real-time_2009}. However, the assumption of $L_e$ can lead to errors. A better way to measure the single vehicle speed is to use a double-loop detector, where two loops are in a well-defined distance behind each other. The time gap between the occupancy of both detectors and the detector distance yields a direct calculation of the vehicle speed $u_i$. In this work only single loop detectors are used, since they are more common \cite{kurzhanskiy_traffic_2015}. 
Generally, loop detectors are positioned in front of an intersection to observe the traffic behavior. There are two different ways of positioning: The \textit{advanced detector} is positioned at a certain distance ahead of the intersection (e.g. 100 m) while the \textit{stop bar detector} is directly at the intersection behind the stop bar \cite{liu_real-time_2009}. Both are important for the implementation of queue length estimation in the following chapter. In reality, the implementation and number of loop-detectors depends on the location (city, country) of the measuring system \cite{kurzhanskiy_traffic_2015}. 
Usually the raw data from loop detectors are time-averaged before they are sent to a main server. Thereby, the amount of data and necessary memory is decreased to a minimum. Usually the reporting frequency of the preprocessed data is about 30 to 60 seconds \cite{chen_freeway_2001}. However, in this work a reporting frequency  of 1 second is necessary. Otherwise, a successful implementation of the queue length estimation based on \cite{liu_real-time_2009} is not possible, because this method requires high frequency data for characterization of changes in traffic behavior.

In the following, important variables like \textit{speed}, \textit{flow} and \textit{density} are introduced. They are fundamental to describe traffic situations and are essential for the estimation algorithm in \cite{liu_real-time_2009}.

\subsubsection{Speed}
As mentioned above, the speed of a single vehicle can be calculated from single-loop detector data by measuring the detector occupancy time $t_{o,i}$ for every vehicle $i$ and assuming an effective vehicle length $L_e$ 
\begin{equation} \label{eq:indiviual_speed}
	u_i = L_e/t_{o,i}. 
\end{equation}

The space mean speed $u_s$ over $n$ vehicles of the same traffic situation can be then calculated based on \cite{liu_real-time_2009} as 

\begin{equation} \label{eq:space_mean_speed}
	 u_s = 1/\Big(\frac{1}{n} \sum_{i=1}^{n} \frac{1}{u_i}\Big).
\end{equation}

\subsubsection{Flow}
The flow $q$ is the number of vehicles that pass the detector in a certain period of time \cite{gomes_optimization_2004}. Therefore, the dimension of flow is \textit{vehicles per hour} or \textit{vehicles per minute}. The flow can be calculated as
\begin{equation} \label{eq:flow}
		q = 1/\Big(\frac{1}{n} \sum_{i=1}^{n} (t_{o,i}+t_{g,i})\Big) 
\end{equation}

with the detector occupancy time $t_{o,i}$ and the time gap $t_{g,i}$ for every vehicle \cite{liu_real-time_2009}. 

\subsubsection{Density}

The density describes how many vehicles are located in a certain section of the road at a specific point of time \cite{treiber_traffic_2013}. The dimension is \textit{vehicles per distance} and can be calculated based on the hydrodynamic relation

\begin{equation} \label{eq:density}
	k = \frac{flow}{speed} = \frac{q}{u_s}.
\end{equation}

With these basic definitions it is possible to describe the traffic situation based on single-loop detector data. Flow, density, time gap between vehicles and detector occupancy time are the main units to implement the queue length estimation based on \cite{liu_real-time_2009} but also to train the deep neural networks presented in chapter \ref{sec.geometric_deep_learning}. Based on the traffic flow and density, the fundamental diagram can be introduced. 

\subsection{The Fundamental Diagram}\label{sec.fundamentaldiagram}
\begin{figure}[h]
	\centering
	\includegraphics[width=0.7\textwidth]{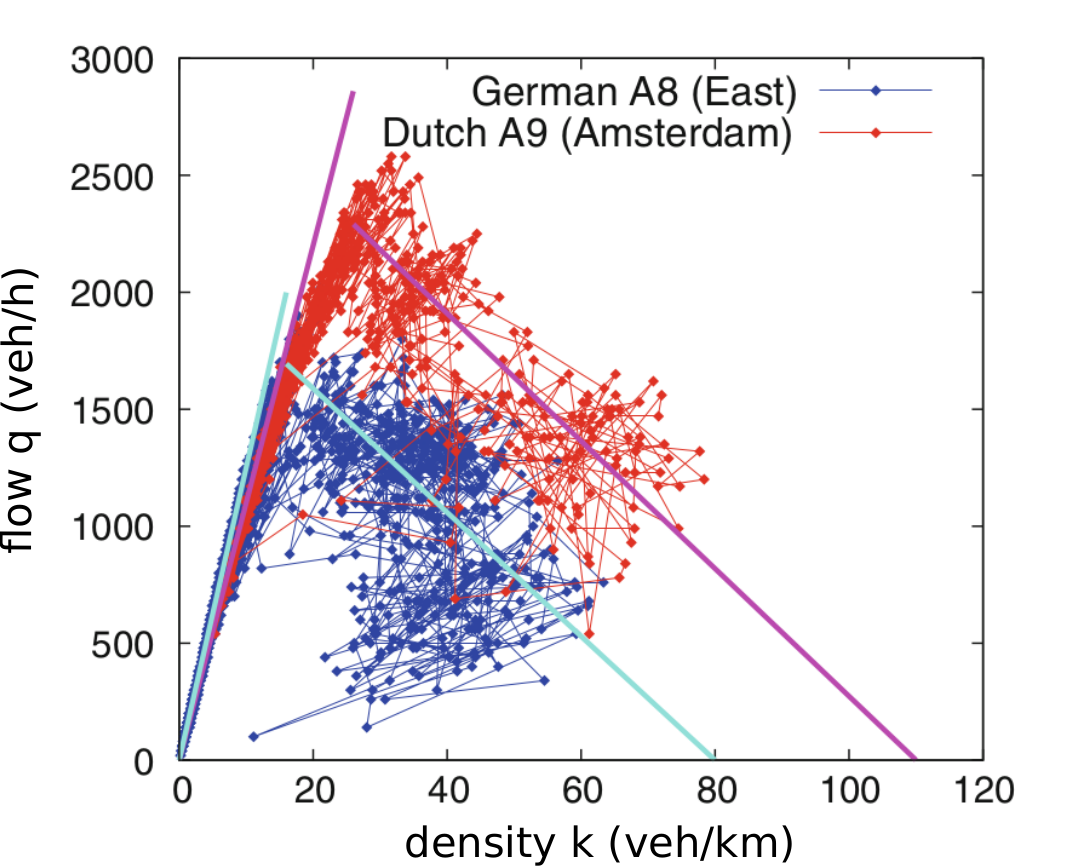}
	\caption[Flow-density diagram]{Example of a flow-density diagram for the Dutch A9 and the German A8. The difference between the data for free flow and congestion is emphasized \cite{treiber_traffic_2013}.}\label{bild.flow_density_diagram}
\end{figure}

In Fig. \ref{bild.flow_density_diagram} a flow-density diagram is shown based on traffic data which rely on non-stationary traffic situation.
By plotting aggregated empirical data of the flow and density in a graph, a characteristically relationship can be found. At lower density the flow begins to increase (positive slope) with increasing density, but after a critical point the flow decreases with increasing density. This critical point is called \textit{critical density} $k_{max}$ and it separates the traffic-flow diagram in two sections. Before $k_{max}$ the traffic is in \textit{free flow} and for higher densities it is in \textit{congestion}. 

\begin{figure}[h]
	\centering
	\includegraphics[width=0.7\textwidth]{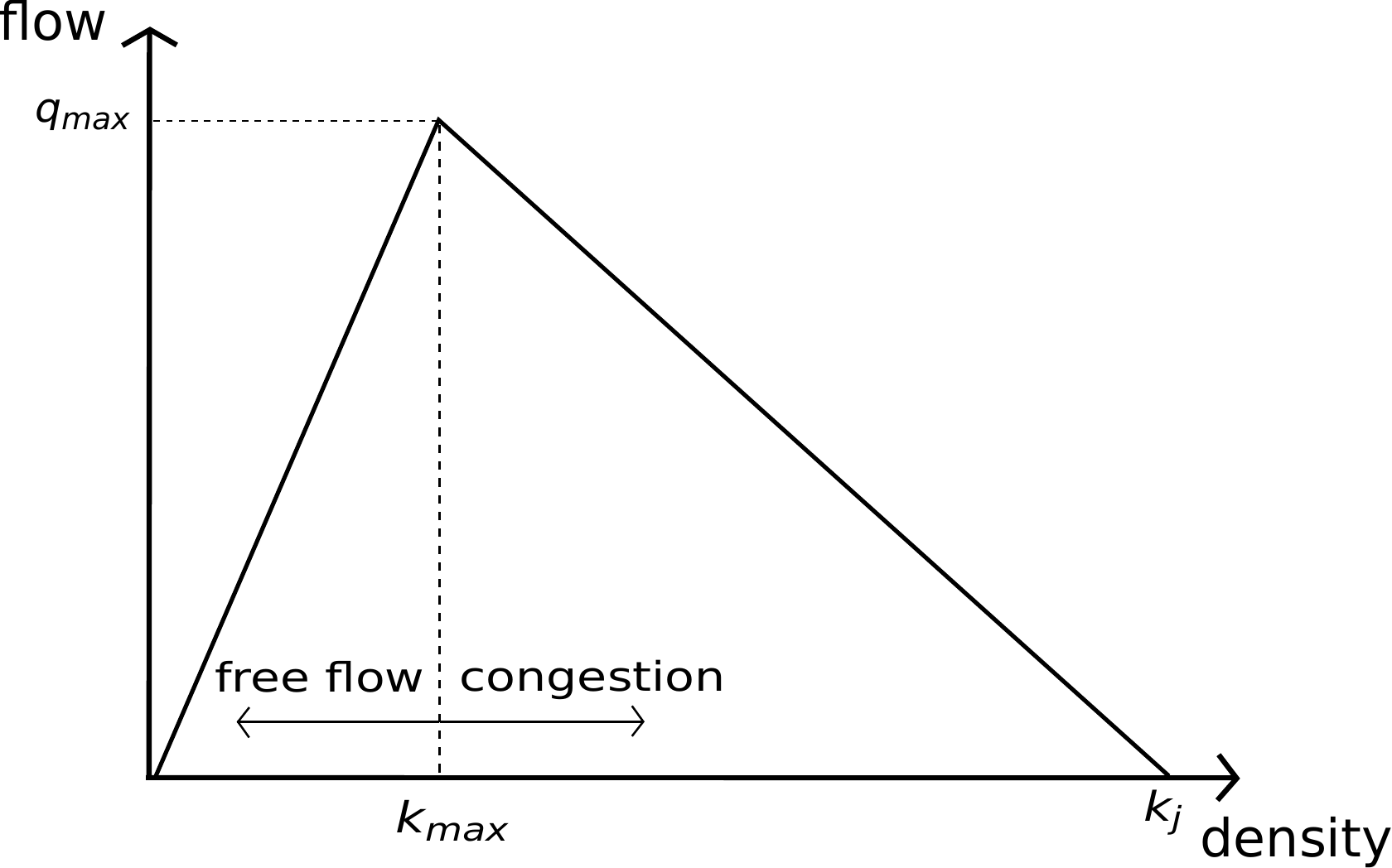}
	\caption[Fundamental diagram]{Fundamental diagram with critical density $k_{max}$ and flow $q_{max}$ \cite{gomes_optimization_2004}}\label{bild.fundamental_diagram}
\end{figure}

Before the critical density is reached, the flow increases linearly ($q = u_0 k$, where $u_0$ is the desired speed), while for the data after critical point $k_{max}$ a wide scattering occurs. The main reasons for this phenomenon are that the traffic is not in an equilibrium and local traffic inhomogeneities as well as different vehicle types are present \cite{treiber_traffic_2013}. Different vehicle types corresponds to different data points in the graph. The consideration of theoretically ideal data of homogeneous traffic situations in an equilibrium leads to the \textit{fundamental diagram} which is shown in Fig. \ref{bild.fundamental_diagram}. 

The fundamental diagram can be interpreted as follows: At the density $k=0$ the flow $q$ is zero as well, because no vehicle pass the detector. With increasing density the flow is increasing, too. More vehicles are on the lane, but the traffic is still in free flow with the speed $u_0$ so that the correlation is linear. At a certain density $k_{max}$ the maximum flow is reached. Now the traffic state changes in congestion mode. With increasing density the flow decreases. There are now more vehicles on the lane (high density) but the average velocity is lower so that the amount of vehicles that pass the detector within a certain period of time decreases. As soon as the jam density $k_j$ is reached, the flow is zero again \cite{treiber_traffic_2013}\cite{gomes_optimization_2004}. The fundamental diagram is the basis for the shockwave theory presented in the following section. 

\subsection{Calculation of the Traffic Pressure} \label{sec.pressure}
The main reason why it is of interest to estimate queue length in road networks, is to use this information for a following control algorithm to reduce congestion.
When it comes to control of signalized intersections in road networks, an important value is the \textit{traffic pressure}. The traffic pressure can be defined as "a response if the local ensemble of vehicles or drivers on density gradients" \cite{treiber_traffic_2013}. Since the queue length on a lane is proportional to the vehicle density regarding a lane, the difference between queue lengths can also be used to calculate the traffic pressure. In the past, approaches for traffic control based on pressure were made \cite{varaiya_max_2013} \cite{wongpiromsarn_distributed_2012}. In \cite{varaiya_max_2013}, the pressure for a single lane $l$ and a specific traffic light phase can be calculated as
\begin{equation}
	P_l = q_{m,l} \Big(x_l - \sum_{k \in \mathcal{O}_l} r_k x_k\Big) 
\end{equation}
where $q_{m,l}$ is the saturation flow of the lane, $x_l$ the queue length for lane $l$, $\mathcal{O}_l$ is the set of output lanes connected to $l$ with the corresponding proportion of vehicles leaving $r_k$ and the queue length on the output lanes $x_r$. In other words, the pressure can be calculated as the difference of the queue length between the input and the output lanes. In \cite{varaiya_max_2013}, pre-determined routing proportions $r_k$ are assumed. These can be estimated either by a person who observes the traffic and determines the proportions manually or by more advanced measurement systems which are able to track specific vehicles. The saturation flow $q_{m,l}$ can be estimated by loop detectors as presented in section \ref{sec.loop_detector_data}.

In the following, the main focus of this thesis is to estimate the queue length accurately, rather than calculating the pressure. However, once the queue length are estimated, the pressure can be directly calculated based in the estimations. 

\section{Real-Time Queue Length Estimation for Intersections} \label{sec.liu_method}
%overview 

%why are we using the liu method? It is fairly simple compared to other estimation methods and could also be applied in reality because it only uses loop detector data. Furthermore, it only uses infromations from one lane. In contrast to that, the geometric deep learning approch uses spatial correlations. So if a benefit of using spatial correlations occur, it must turn out in the comparison of both approaches. 

Several approaches for queue length estimation in front of intersections have been proposed in the past. Generally, there are two different types of methods. The first one is based on analysis of input-output of traffic for an intersection \cite{sharma_inputoutput_2007}\cite{vigos_real-time_2008}. The amount of incoming and outgoing vehicles in a specific section in front of an intersection is counted and the queue length is calculated. The biggest disadvantage regarding this methods is, that if the queue length exceeds the distance between stop bar and loop detector, the methods cannot estimate the queue length anymore. In this particular case, the number of outgoing and incoming vehicles have no information about the queue length behind the  input-output section. That limits the field of application. The second group of methods relies on the shock wave theory for traffic queues from Lighthill and Whitham \cite{lighthill_kinematic_1955} and Richards \cite{richards_shock_1956}. Queue lengths are estimated by tracking shockwaves through the traffic, this effect is used in \cite{stephanopoulos_modelling_1979} and \cite{skabardonis_real-time_2008} for instance. Unfortunately, these methods require ideal conditions like knowledge of vehicle arrival and constant driver behavior.
For this thesis, the implemented method after \cite{liu_real-time_2009} belongs to the second group which relies on the detection of shockwaves. This method is chosen, since it is fairly simple compared to other estimation methods. Furthermore, it is applicable in reality without high cost and effort, because it uses loop detector data. In contrast to the deep learning approach, presented in chapter \ref{sec.geometric_deep_learning}, it uses only loop detector data from a single lane and the last three traffic light cycles in the past. The geometric deep learning approach is able to find spatial and temporal correlations in the road network, which makes it potential superior and emphasizes the importance of these correlations. The main goals for the implementation of the method after \cite{liu_real-time_2009} are, to have a reference for comparison to the deep learning approach. Furthermore, it is of interest to analyze, if the estimation results based on the method are beneficial for the performance of the deep learning approach, when it is used as input for the deep learning model.

The real-time estimation method after \cite{liu_real-time_2009} is based on the Lighthill-Whitham-Richards (LWR) model. The advantage of this method is, that a reliable estimation becomes feasible also for long queues (queues where the length exceeds distance between stop bar and detector). Since high resolution detector data with a reporting frequency of one second in combination with traffic light data are used, it is possible to detect characteristic breakpoints within the queue dynamics. These breakpoints make an estimate possible based on usual stop bar or advanced loop detectors.
In this section the shockwave theory and breakpoint characterization is introduced. Afterwards, the basic model and an expansion model, which is more eligible for second-by-second data, will be presented. Finally, the modifications regarding this work and the implementation first on a single intersection and later on a entire road network will be shown.

\subsection{Shockwave Theory and Breakpoint Characterization}

The description of shockwaves is based on the Lighthill-Whitham-Richards (LWR) model which assumes that flow is a function of density. The LWR model is inspired by fluid mechanics and express the conservation of vehicles on the road \cite{gomes_optimization_2004}. The fundamental and simplest form of the LWR model can be written as

\begin{equation} \label{eq:LWR_model}
	  \frac{\partial k}{\partial t} + \frac{d q(k)}{d k} \frac{\partial k}{\partial x} = 0.
\end{equation}

The traffic shockwave theory can be deduced from solving the partial differential equation. A traffic shockwave is defined as a spatial motion of sudden change in density. The shockwave velocity can be calculated by the change of flow and density

\begin{equation}
	v_w = \frac{\Delta q}{\Delta k} = \frac{q_{downstream} - q_{upstream}}{k_{downstream} - k_{upstream}}.
\end{equation}

\begin{figure}[h]
	\centering
	\includegraphics[width=0.7\textwidth]{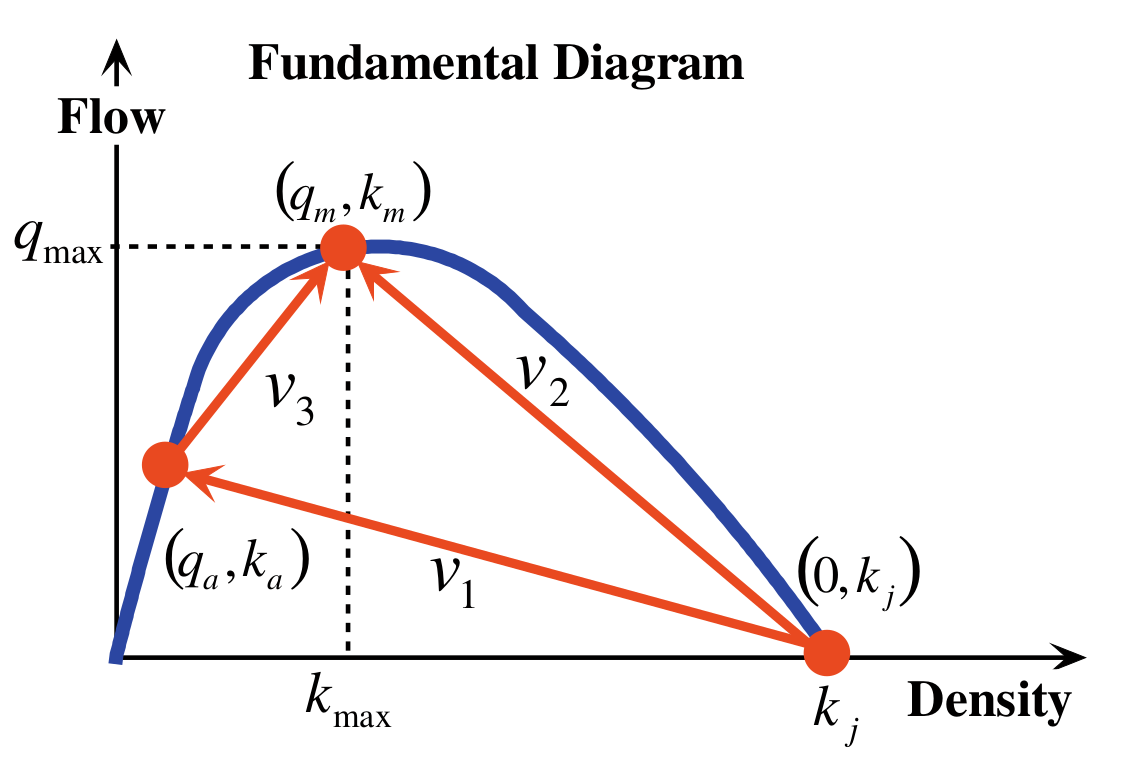}
	\caption[Fundamental diagram including shockwaves]{Shockwave velocities shown in the fundamental diagram \cite{liu_real-time_2009}.}\label{bild.shockwave_fundamental_diagram}
\end{figure}

This correlation can be also shown in the fundamental diagram (Fig. \ref{bild.shockwave_fundamental_diagram}), where the tangent of the chord between the two traffic states represents the shockwave velocity.

\begin{figure}[h]
	\centering
	\includegraphics[width=1\textwidth]{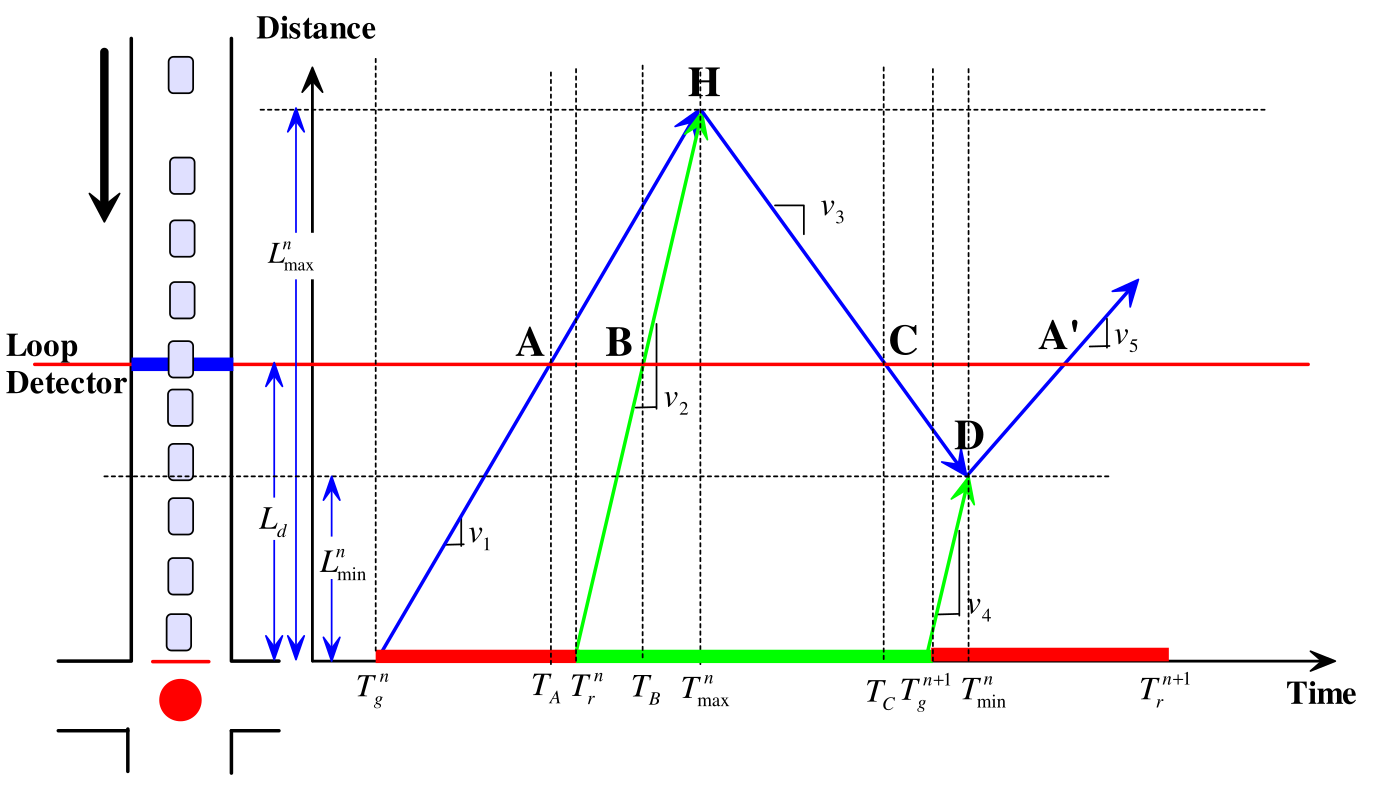}
	\caption[Distance-time diagram for a traffic light cycle]{Distance-time diagram for a traffic light cycle \cite{liu_real-time_2009}.}\label{bild.queue_length}
\end{figure}

The points in the diagram represent the different traffic situations with the  traffic flow $q$ and density $k$: $(0, k_j)$ is the point for the vehicles waiting in queue, $(q_m, k_m)$ is the traffic situation where vehicles are discharging from lane and the point $(q_a, k_a)$ describes the arrival traffic.
During a traffic light cycle different situations with different shockwaves occur. For estimating the queue length, it is necessary to determine these shockwaves. Assuming an empty lane in front of an intersection with a red light starting at point of time $T_n^g$, the arriving cars have to stop and wait for the green interval starting at time $T_n^r$. Thus a queue forms upstream and the vehicles change their state from free arrival to a jam. The queuing shockwave velocity $v_1$ can be calculated by

\begin{equation} \label{eq:v_1}
	v_1 = \frac{0-q_a^n}{k_j-k_a^n}
\end{equation}

where 0 is the jammed flow, $k_j$ the jammed density, $q_a^n$ the average arrival flow and $k_a^n$ the arrival density for the $n$th cycle. As soon as the light turns green ($T_r^n$), the cars begin to discharge from the lane and a second shockwave $v_2$ propagates upstream with the velocity 

\begin{equation} \label{eq:v_2}
	v_2 = \frac{q_m-0}{k_m-k_j}
\end{equation}

where $q_m$ and $k_m$ are the discharge flow and density. In Fig. \ref{bild.queue_length} is shown the queue length over time. The shockwave $v_1$ starts with the red light interval and the discharge shockwave $v_2$ starts at the beginning of the green interval. Thus $v_2$ is higher than $v_1$, the intersection between these both shockwaves is the point H of maximum queue length $L_{max}^n$ at point of time $T_{max}^n$. 

After the meeting of shockwave one and two, a third wave propagates downstream. It is called the departure shockwave and has the velocity 

\begin{equation} \label{eq:v_3}
	v_3 = \frac{q_m -q_a^n}{k_m-k_a^n}.
\end{equation}

Since the vehicles discharge during the green phase ($T_r^n$ to $T_g^{n+1}$), the queue length decreases again.
When the red light interval starts again for the ($n+1$)th cycle, a fourth shockwave is propagating upstream with the velocity $v_4$ which has the same value as $v_2$:

\begin{equation} \label{eq:v_4}
	v_4 = \frac{0-q_m}{k_j-k_m}.
\end{equation}

The point D of lowest queue length $L_{min}^n$ at time $T_{min}^n$ is where wave three and four meet. From point D on, the queuing shockwave for the ($n+1$)th cycle with the speed

\begin{equation}
	v_5 = \frac{0-q_a^{n+1}}{k_j-k_a^{n+1}}
\end{equation}

occurs, and has a similar value to $v_1$. The for the velocity calculation required flow and density values can be estimated by the equations \ref{eq:flow} and \ref{eq:density} from section \ref{sec.loop_detector_data}.
For every traffic light cycle the process starts again. At the points A, B and C in Fig. \ref{bild.queue_length} the traffic flow changes at the position of the loop detector. The points A, B and C are called \textit{breakpoints} and they are essential for the estimation of queue length. Once the data for one cycle $n$ are collected, the characterization of breakpoints can be processed. 

\begin{figure}[h]
	\centering
	\includegraphics[width=1\textwidth]{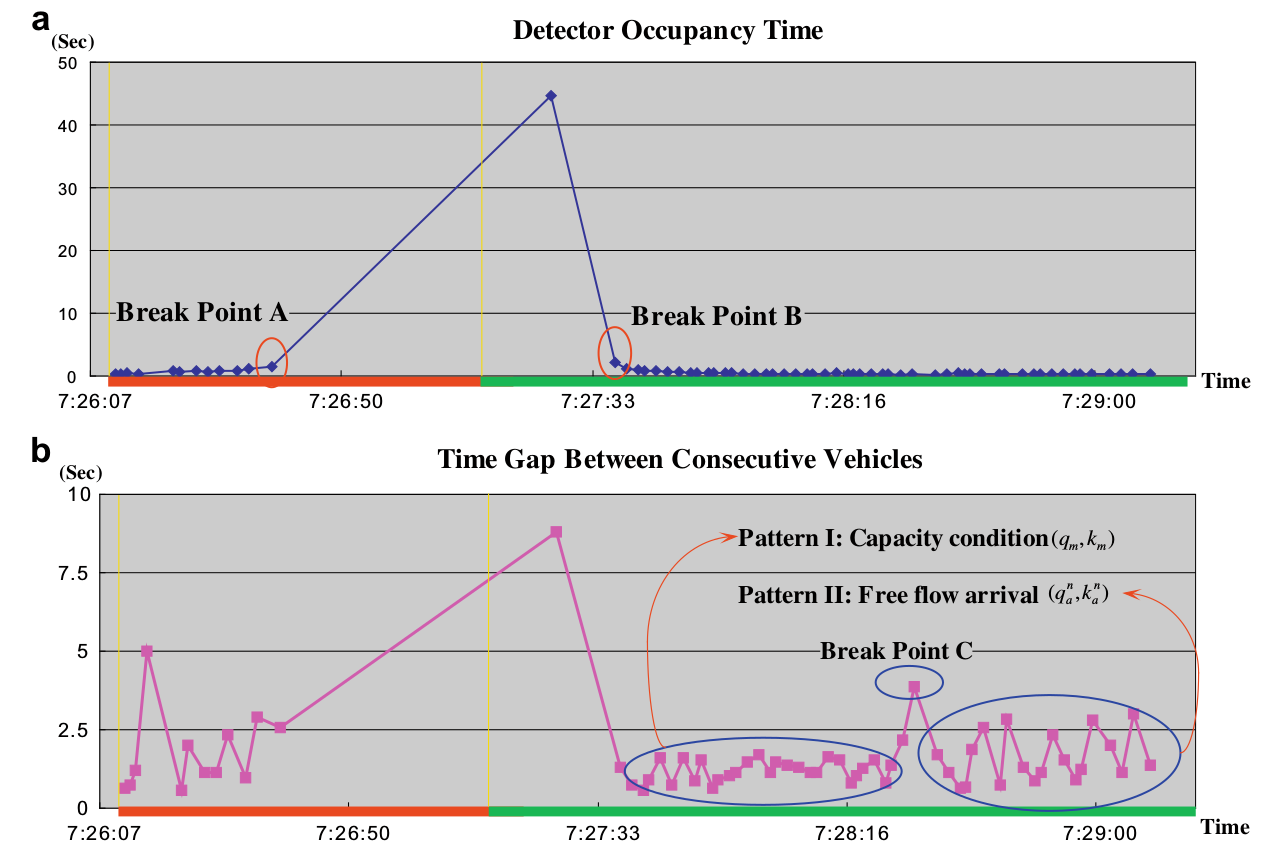}
	\caption[Breakpoint detection in loop-detector data]{(a) Detector occupancy time with breakpoint A and B; (b) shows the time gap between consecutive vehicles with breakpoint C \cite{liu_real-time_2009}.}\label{bild.breakpoints}
\end{figure}

For the detection of the breakpoints A and B, the detector occupancy time is the relevant value. Breakpoint A only exists when the queue length exceeds the distance between stop bar and detector. In this case the detector occupancy time increases for a specific period of time. The beginning of this period of time is breakpoint A ($T_A$), and the end breakpoint B ($T_B$) (see Fig. \ref{bild.breakpoints} (a)). If no breakpoint A and B exist, the queue length is short and does not exceed the detector. Threshold for detecting the breakpoints is based on \cite{liu_real-time_2009} around three seconds. Therefore, it is necessary to have access to high resolution loop detector data with a reporting frequency of less or equal than one second. 
For characterization of breakpoint C ($T_C$) the time gap of consecutive vehicles is crucial. The Point C can only exist when this particular cycle has a long queue so A and B also exist. Breakpoint C is the point of time in a cycle, when after the breakpoint B a higher time gap occurs between consecutive vehicles. This high time gap is caused by a change of traffic pattern passing the detector. After the start of the green phase, vehicles discharge and the detector senses low time gaps. But as soon as all vehicle of the former queue pass the detector, newly arriving vehicles with different time gaps begin to pass the detector (see Fig. \ref{bild.breakpoints}(b)). In other words the breakpoint C describes the point of time ($T_C$) when the last vehicle of the queue passes the detector. C is the most important breakpoint and essential for the queue length estimation. In \cite{liu_real-time_2009} a threshold of 2.5 seconds for the time gap was chosen. 
If breakpoint A and B exist but C cannot be found, \textit{oversaturation} occurs. In this case the queue is so long that in result the rear of queue always exceeds the location of detector. Thus no pattern changes at the loop detector, breakpoint C cannot be determined and the estimation of queue length becomes problematic. 
The shockwave velocities and breakpoints that were introduced in this chapter, are fundamental for the following basic model.

%replace Abbildung with Figure!!!

\subsection{Basic Model} \label{sec.basic_model}
%implement online estimation of v3!
The aim of the basic model is to estimate the spatial information ($L_{max}^n$) and the temporal information ($T_{max}^n$) about the maximum queue length. Based on the calculated breakpoints A, B, C and shockwaves in the previous chapter, the basic model can be applied. Either the estimation of shockwave velocities is processed and updated for every cycle based on new detector data by calculating flow and density, or only one time at beginning of the whole estimation process. It depends whether the traffic demand changes during the estimation period or not. Especially $v_3$ depends on traffic arrival which could probably change, whereas $v_2$ is almost constant. The shockwave velocity $v_2$ could be either calculated by eq. \ref{eq:v_2} or by 

\begin{equation} \label{eq:v_2_alterative}
	v_2 = \frac{L_d}{T_B-T_r^n}
\end{equation}

where the velocity is distance $L_d$ between detector and stop bar divided by period of time between green light start and breakpoint B. The maximum queue length can be then calculated by

\begin{equation} \label{eq:L_max_basic}
	L_{max}^n = L_d + \frac{T_C-T_B}{\frac{1}{\mid v_2 \mid} + \frac{1}{v_3}}.
\end{equation}
It should be noted that the absolute value of $v_2$ is used, since by applying eq. \ref{eq:v_2} the sign of value is always negative. The time of maximal queue length is
\begin{equation}\label{eq:T_max_basic}
	T_{max}^n = T_B + \frac{L_{max}^n-L_d}{\mid v_2 \mid}.
\end{equation}

With this spatial and temporal information the point H in Fig. \ref{bild.queue_length} could be estimated. Calculation of the minimum queue length in point D is 
\begin{equation}
	L_{min}^n = \frac{\frac{L_{max}^n}{v_3} + T_{max}^n - T_g^{n+1}}{\frac{1}{v_3}+\frac{1}{\mid v_4 \mid}}
\end{equation}
at the point of time

\begin{equation}
	T_{min}^n = T_g^{n+1} + \frac{L_{min}^n}{\mid v_4 \mid}.
\end{equation}

Every characteristic point of Fig. \ref{bild.queue_length} is determined, now the queuing process can be entirely described \cite{liu_real-time_2009}.

\subsection{Expansion Model} \label{sec.expasion_model}

\begin{figure}[h] 
	\centering
	\includegraphics[width=1\textwidth]{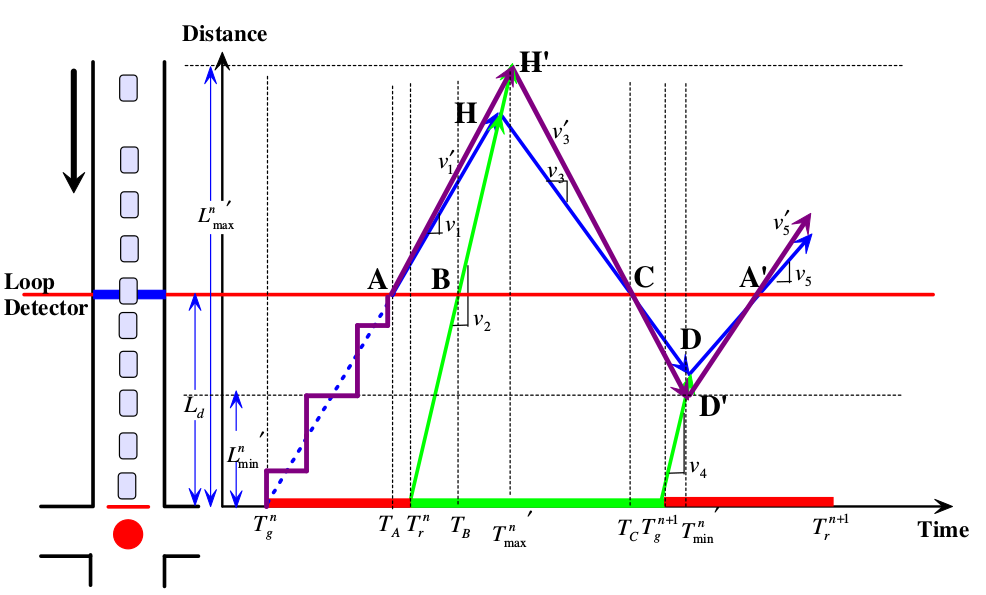}
	\caption[Expansion model]{Expansion model with estimated point H' \cite{liu_real-time_2009}.} \label{bild.expansion_model}
\end{figure}

The expansion model is here presented as an alternative for the basic model, especially when only second-by-second data are available. This model is based on the idea to count the number of vehicles that pass the detector between green light start $T_r^n$ and $T_C$ and use the jam density $k_j$ to calculate the maximum queue length. This is feasible under the assumption that most of the counted vehicles passing before $T_C$ were part of the maximum queue length. But generally this method overestimates, because a few vehicles stack on the queue after $T_{max}^n$ and were counted by the loop detector as well. This slight overestimate leads to point H' like in Fig. \ref{bild.expansion_model} shown. 

The maximum queue length can be estimated with 

\begin{equation}\label{eq:L_max_expansion}
	L_{max}^{n'} = \frac{N}{k_j} + L_d
\end{equation}

where $N$ is the number of vehicle counted by the detector within the period of time from $T_r^n$ to $T_C$. The resulting point of time is

\begin{equation}\label{eq:T_max_expansion}
	T_{max}^{n'} = T_r^n + \frac{L_{max}^{n'}}{v_2}.
\end{equation}

The presented methods of \cite{liu_real-time_2009} are conventional, since they use loop-detector data from only one specific lane. In the following section the theoretical basics for the geometric deep learning model are introduced, which is able to find spatial and temporal correlations in measurements from multiple lanes.

%---------- Abschnitt neu Eingefügt aus deep learning ------------
\section{Fundamentals of Geometric Deep Learning} \label{sec.Fundamentals_GDL}

Deep learning approaches became very popular during the past decades, since they were successfully applied in many different fields to solve real-world problems. Especially deep learning had big achievements in the past in fields like human language translation, computer vision, robotics, search engines and many more. Deep learning is a name for a particular class of machine learning techniques. Many of the recent advancements in the field of artificial intelligence (AI) are based in deep learning techniques. In classic machine learning approaches the input data are manually preprocessed by humans, which is called feature engineering. These hand-designed features are the input for the artificial neural network (ANN) or any other sort of statistical model which predicts the output. The process of feature engineering can be a big effort and requires domain knowledge about the data from the application engineer. For more complex problems it is sometimes not even possible to create the features by hand, for example in the field of face recognition or speech-to-text applications where the inputs are pictures or time sequences. Deep learning has the approach to preprocess the input data by itself and abstract the features by using several layers of neural networks. Thereby, a lot of time and effort in the application process can be saved and more complex problems can be solved. A disadvantage is, that the higher amount of layers and parameters increases the computational effort \cite{goodfellow_deep_2016}. In this work, the input data are the detector data from the e1 (loop) detectors which are fed into the deep learning model. The hypothesis is that the deep learning model should abstract the relevant features by itself by using the knowledge of the spatial correlations within the road network (usage of graph attention layers). This makes this model to a \textit{geometric} deep learning model \cite{bronstein_geometric_2017}.

In this section the theoretical fundamentals about deep learning are presented, which are necessary to successfully implement geometric deep learning networks for queue length estimation. Section \ref{sec:basics_NN} starts with the basics and general introduction of feedforward neural networks. To extract spatial information from the inputs, graph attention layer is introduced in section \ref{sec:gat_layer}. Recurrent neural networks (RNNs), presented in section \ref{sec:RNN}, are used to store relevant information over multiple time steps and represent the main component for the  RNN encoder-decoder model introduced in section \ref{sec:enc-dec}.

\subsection{Basics of Feedforward Neural Networks} \label{sec:basics_NN}
Feedforward neural networks are the most simple and fundamental kind of deep learning models.
When it comes to solving real-world problems, neural networks have great characteristics like nonlinearity, parallelism, handling of uncertain and fuzzy information as well as the ability to generalize knowledge \cite{basheer_artificial_2000}. Generally, the aim of these models is to approximate a function $f^*$. The general formulation for a regression would be $ \boldsymbol{\hat{y}} = f(\boldsymbol{x}, \boldsymbol{W})$, where $\boldsymbol{x}$ is the input, $\boldsymbol{\hat{y}}$ the predicted output and $\boldsymbol{W}$ the weight matrix. During the training process, the optimization algorithm learns the parameters that $f$ approximates $f^*$ the best. These models are called feedforward neural networks, since the input is fed trough the network and results an output without a feedback loop. Once a feedback loop is implemented, the model is called recurrent neural network (see section \ref{sec:RNN}). 

\subsubsection{Structure} \label{sec.structure}

\begin{figure}[h]
	\centering
	\includegraphics[width=1\textwidth]{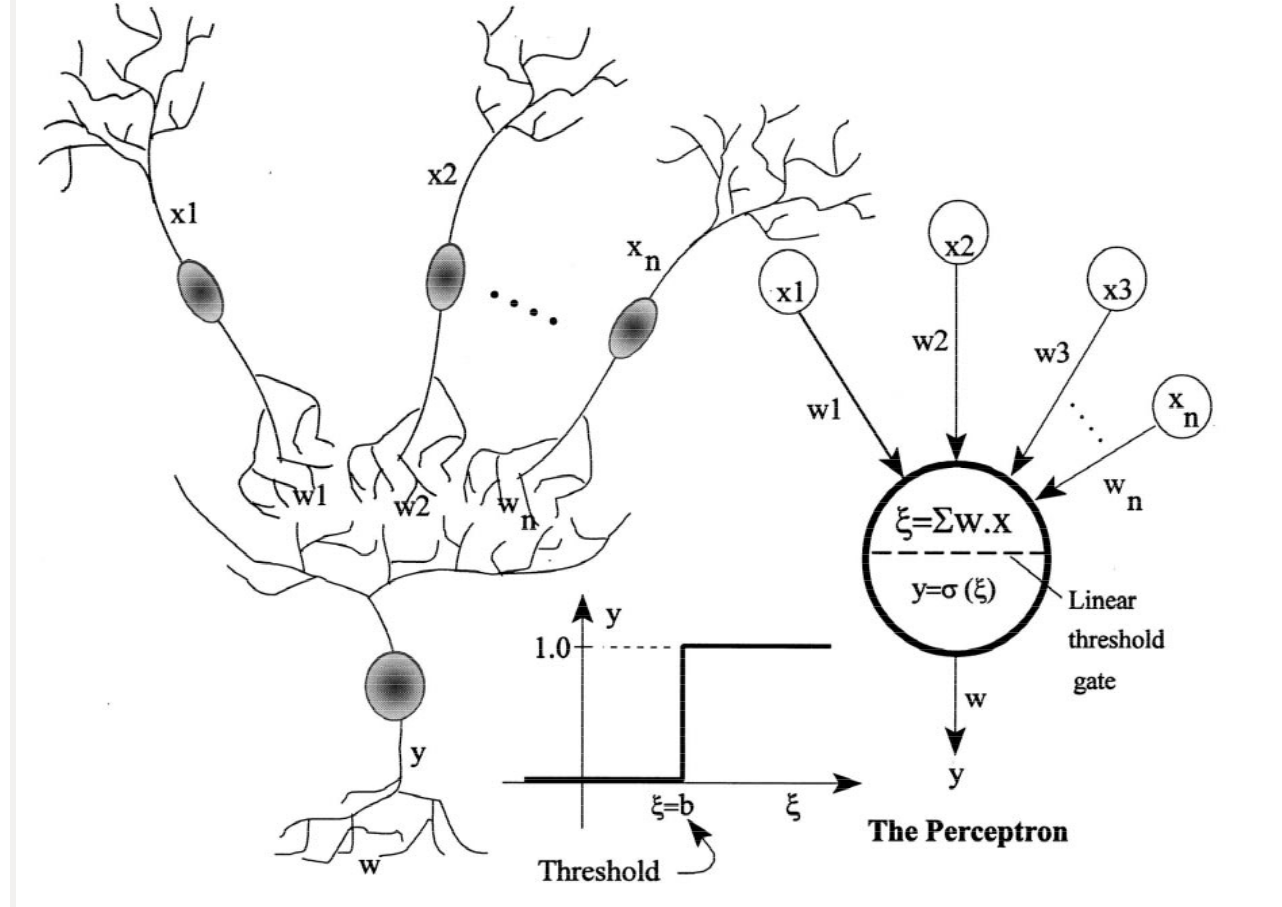}
	\caption[Analogy between biologial and artificial neurons]{Analogy between biologial (left) and artificial neurons (right) \cite{basheer_artificial_2000}.} \label{bild.neurons}
\end{figure}

Inspired by biological neural networks, a complex artificial neural network consists of many simple elements (artificial neurons or nodes). The analogy between the biological neurons to the artificial neurons is presented in Fig. \ref{bild.neurons}.

\begin{figure}[h]
	\centering
	\includegraphics[width=1\textwidth]{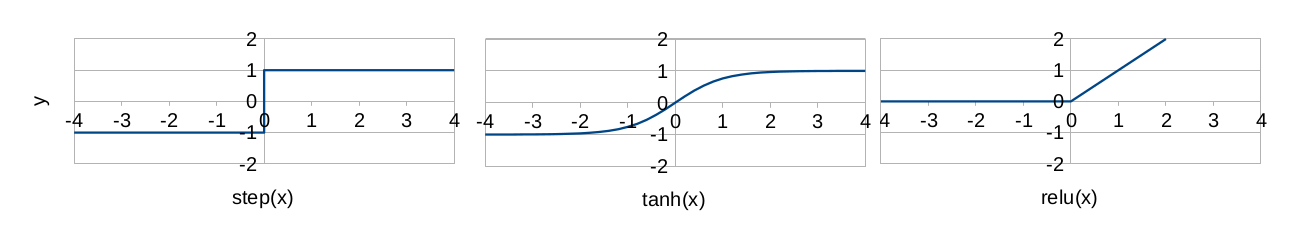}
	\caption[Activation functions for deep learning models]{Examples for activation functions: step-function (left), tanh function (middle) and the rectified linear unit (ReLU) function (right) \cite{neubig_neural_2017}.} \label{bild.activation_functions}
\end{figure}

\begin{figure}[h]
	\centering
	\includegraphics[width=1\textwidth]{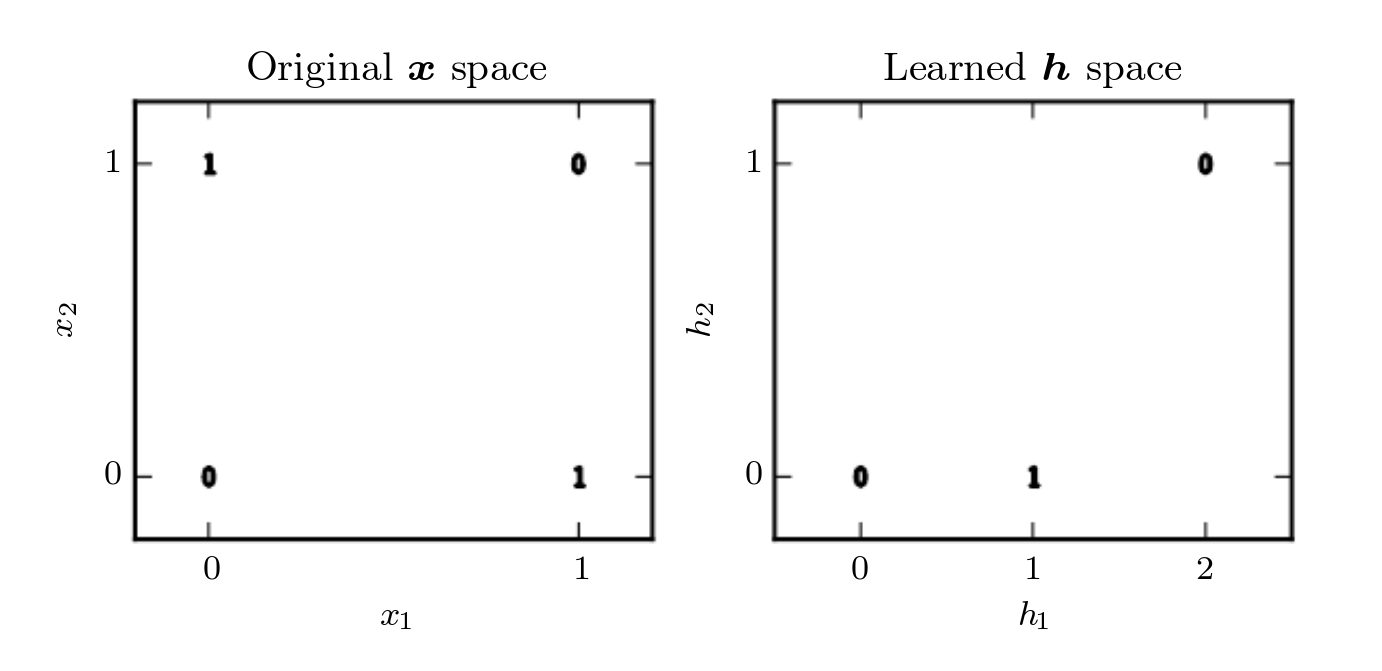}
	\caption[The XOR problem in deep learning] {The XOR problem with the original $\boldsymbol{x}$ space of the XOR logic on the left and the transformed representation after the first hidden layer (learned $\boldsymbol{h}$ space) on the right \cite{goodfellow_deep_2016}.} \label{bild.XOR_problem}
\end{figure}

\begin{figure}[h]
	\centering
	\includegraphics[width=0.45\textwidth]{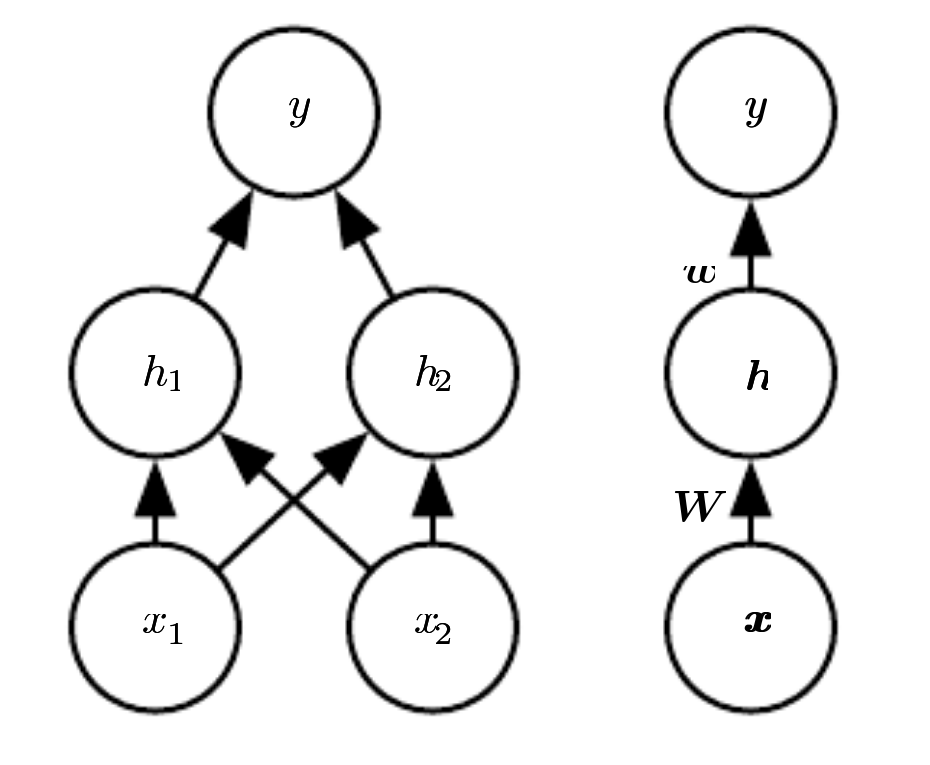}
	\caption[Structure of a model with one hidden layer] {Structure of the model to solve the XOR problem with one hidden layer. On the left side every node is drawn explicit and on the right side an simplified overview with the weights is given \cite{goodfellow_deep_2016}.} \label{bild.XOR_structure}
\end{figure}

At the artificial neuron, the inputs $x_1$ to $x_n$ are multiplied with the corresponding weights $w_1$ to $w_n$ and get summed-up (this process is called \textit{affine transformation}). Afterwards, the sum is an input for the activation function, which adds the nonlinearity to the model. In this particular case, the activation function is a step function, which only lets the value pass, if a threshold is reached. But there are also other functions possible (shown in Fig. \ref{bild.activation_functions}).
According to \cite{goodfellow_deep_2016}, the most recommended default activation function for deep learning applications is the ReLU function, which is defined by $\sigma(z) = max\{0,z\}$. Due to its piecewise linear characteristic, it has good optimization properties for gradient-based optimization algorithms.   
Generally, neural networks chain together many simple affine transformations and activation functions to represent more complex functions. Due to this simple structure, it is easier and more efficient to train.
A very good example for explaining how deep neural networks in principle work, is solving the XOR problem. The visualization of the XOR problem is shown in Fig. \ref{bild.XOR_problem} on the left.
For the original $\boldsymbol{x}$ space of the XOR logic, it is not possible to classify between 1 and 0 just by dividing the space into two parts by a line, since there is no configuration of a line that would separate the zeros from the ones. According to the XOR logic, the value becomes only one if exactly one of the values $x_1$ or $x_2$ is one. So a basic linear model like

\begin{equation}
y = \boldsymbol{w^T}\boldsymbol{x}+\boldsymbol{b}
\end{equation} 

with the weights $\boldsymbol{w}$ and the bias $\boldsymbol{b}$ is not able to model the XOR logic and divide the zeros from the ones. This problem can be solved by implementing a slightly more complicated structure by adding a \textit{hidden layer} and splitting the calculation in two steps.

\begin{equation} \label{eq:hidden_layer}
\boldsymbol{h} = ReLu(\boldsymbol{W}_{xh}\boldsymbol{x}+\boldsymbol{b}_h)
\end{equation}
\begin{equation} \label{eq:output_layer}
y = \boldsymbol{w}_{hy}\boldsymbol{h}+b_y
\end{equation}
Eq. \ref{eq:hidden_layer} represents the hidden layer which takes as inputs the vector $\boldsymbol{x}$ and outputs the vector $\boldsymbol{h}$. The transformed space $\boldsymbol{h}$ is shown in Fig. \ref{bild.XOR_problem} on the right. During the training process, the parameters $\boldsymbol{W}_{xh}$ and $\boldsymbol{b}_h$ are learned so that after the transformation it is possible to apply a linear model, which separates both patterns. This is done by the \textit{output layer} (eq. \ref{eq:output_layer}) to calculate the result $y$. In other words: The points in the $x$ space are transformed in the $h$ space by a matrix multiplication. In this new representation in the $h$ space, it is possible to separate the zeros from the ones by a line. The classification problem is solved by adding an additional hidden layer (eq. \ref{eq:hidden_layer}) to the linear model. The new structure of the model is shown in Fig. \ref{bild.XOR_structure}. 

Generally, in a deep learning model are usually a lot more affine transformations and hidden layers so that more complex, multidimensional problems can be solved. But all this is based on very simple and easy computable affine transformations with following activation functions. So the first layer could be generally described as
\begin{equation} \label{eq:normal_NN}
\boldsymbol{h}^{(1)} = \sigma^{(1)}(\boldsymbol{W}^{(1)}\boldsymbol{x}+\boldsymbol{b}^{(1)})
\end{equation}
with $\boldsymbol{x}$ as input and $\boldsymbol{W}^{(1)}$ as well as $\boldsymbol{b}^{(1)}$ as trainable parameters. The sum is input for the activation function $\sigma(\cdot)$. The second hidden layer is then given by
\begin{equation}
\boldsymbol{h}^{(2)} = \sigma^{(2)}(\boldsymbol{W}^{(2)}\boldsymbol{h}^{(1)}+\boldsymbol{b}^{(2)})
\end{equation}
which takes now $\boldsymbol{h}^{(1)}$ as input. The activation functions $\sigma^{(i)}$ do not necessarily have to be the same. The next hidden layer takes $\boldsymbol{h}^{(2)}$ as input with its own weights and activation function and so on. The last layer of this series is called the \textit{output layer}. The \textit{depth} of the network is defined by number of layers that are chained together. The \textit{width} of the model is determined by the size of the outputs of the hidden layers $\boldsymbol{h}^{(i)}$.

\begin{figure}[h]
	\centering
	\includegraphics[width=0.8\textwidth]{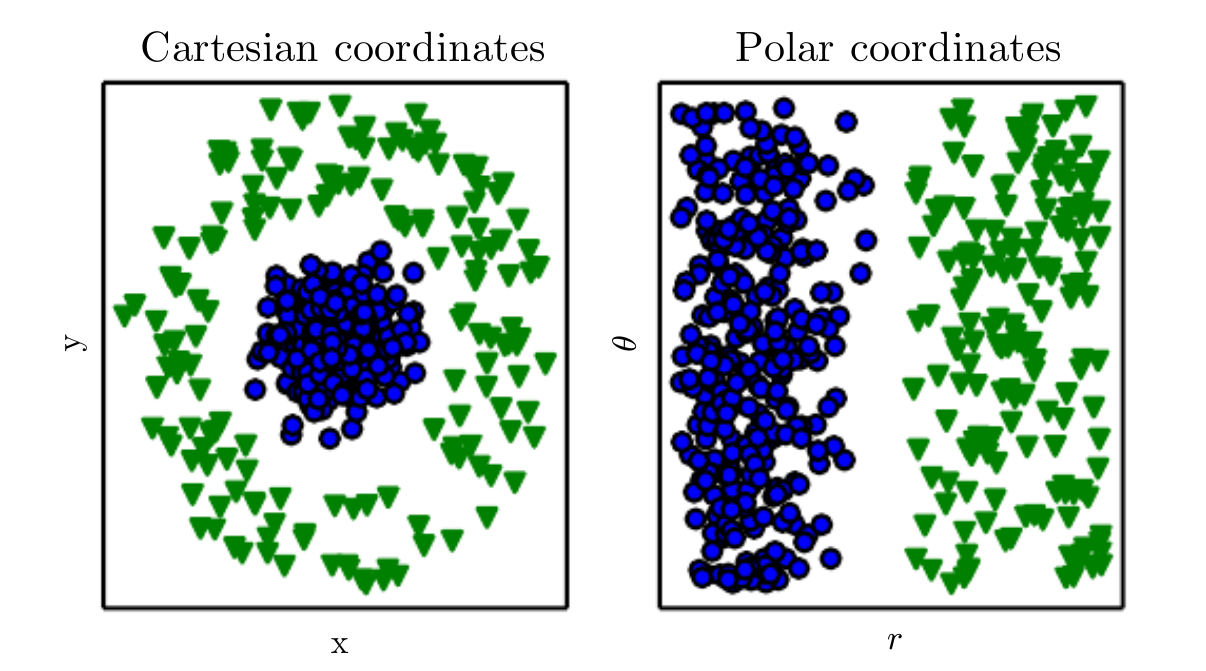}
	\caption[Comparison between patterns in cartesian and polar coordinates] {Comparison between patterns in cartesian and polar coordinates \cite{goodfellow_deep_2016}.} \label{bild.cartesian_polar}
\end{figure}

How powerful it can be to transform the inputs into another representation is emphasized in Fig. \ref{bild.cartesian_polar}. 
Similar to the XOR-Problem, for this example it is also impossible to separate both patterns with a linear model when the representation is in cartesian coordinates. But it becomes possible once the points are transformed in polar coordinates by a hidden layer. Generally, the main task of layers in a neural network is to transform the inputs (features) into another coordinate space, where it is possible to fulfill the original task which could be classification or linear regression. 

The optimal structure of a network with the depth, width and kind of activation functions can only be found in an experimental way, monitoring the errors for the training, test and validation set \cite{goodfellow_deep_2016}. To reach this goal, it is essential to understand the fundamentals how neural networks are trained and how optimization algorithms work. The following chapter provides a brief introduction in this field.

\subsubsection{Training and Optimization}
%intoduce words like supervised learning, targets, freatures
Generally, neural networks are able to solve a variety of tasks, but they have to be specifically trained for this. These tasks can be for instance \textit{classification}, where the network determines the class the input data belongs to. An example would be an deep neural network which has animal pictures as input data and the task is to classify, whether in the picture is a cat, a dog or another animal. A different task would be \textit{regression}, where the output data are a  continuous graph. In this thesis, the presented neural network solves a regression problem, since it has to estimate the time sequence of the queue length based on time sequences of detector data as input. 
Another way to categorize learning algorithms is to differentiate between \textit{supervised} and \textit{unsupervised} learning. In case of machine/deep learning the input data are called \textit{features} $\boldsymbol{x}$ and the ground-truth output data are \textit{targets} or \textit{labels} $\boldsymbol{y}$. \textit{Unsupervised learning algorithms} though, use only the features for training without having corresponding targets. Usually the dataset has many features and the learning algorithm tries to learn correlations within the data. This can be used for tasks like denoising or clustering, where the dataset gets divided into several clusters containing data with similar properties .
For \textit{supervised learning algorithms} exist features $\boldsymbol{x}$ as well as corresponding targets $\boldsymbol{y}$. During training, the neural network model learns to predict $\boldsymbol{y}$ given the the input $\boldsymbol{x}$. This applies to this thesis, with the detector data as features $\boldsymbol{x}$ and the queue length as target $\boldsymbol{y}$. After training, the model should be able to predict an unknown $\boldsymbol{y}$ for a new $\boldsymbol{x}$. 
%mention briefly reinforcement learning, then introduce dataset, loss functions and then backprop
Another popular training method is \textit{reinforcement learning}, where the training model does not learn from a given dataset but from experience by interacting with an environment. But this is beyond the scope of this thesis \cite{goodfellow_deep_2016}. 

Summarized, in case of queue length estimation the approach is a regression problem which is solved by a supervised learning algorithm. For training of a deep neural network it is essential to have a large dataset, which contains of many examples with $n$ features ($\boldsymbol{x} \in \R^n$) and $m$ targets ($\boldsymbol{y} \in \R^m$). To describe the entire dataset with $k$ examples ($\boldsymbol{x}_1 ... \boldsymbol{x}_k$ and $\boldsymbol{y}_1 ... \boldsymbol{y}_k$), all of these examples are concatenated to a matrix $\boldsymbol{X} \in \R^{k \times n}$ (\textit{design matrix}) and $\boldsymbol{Y} \in \R^{k \times m}$.The symbol $k$ is called number of \textit{samples}. %now loss functions
During one training step in a supervised learning algorithm, the model gets one example $\boldsymbol{x}$ and predicts the output $\boldsymbol{\hat{y}}$. After the prediction, $\boldsymbol{\hat{y}}$ is compared to the provided ground-truth $\boldsymbol{y}$. This is usually made by \textit{loss functions}. The most common function for regression problems is the \textit{mean squared error} which is for $n$ features

\begin{equation}
MSE = \frac{1}{n} \sum_{i=1}^{n} (\boldsymbol{\hat{y}}_i^2 - \boldsymbol{y}_i^2).
\end{equation}

If in one training step more than one example is used, the average of every single MSE is calculated. The loss determines how good the prediction is regarding the target. The goal of the training algorithm is to adjust the weights of the neural network, so that the loss gets reduced and the prediction converges to the target. Generally, to minimize the loss is a classic numerical optimization problem, which can be solved by well known numerical optimization algorithms. The most common optimization algorithms are \textit{gradient-based} algorithms. They use the derivative of the function (in this case the deep learning model represents the function) to get information about how to adjust $\boldsymbol{x}$ (or hidden states $\boldsymbol{h}$) in order to minimize the MSE between $\boldsymbol{\hat{y}}$ and $\boldsymbol{y}$. The simplest method is the \textit{gradient descent}, where $\boldsymbol{x}$ (or $\boldsymbol{h}$) is moved in the direction of the opposite sign of the derivative \cite{goodfellow_deep_2016}. 
For a successful optimization it is essential to determine the derivatives in the neural network model. 
% explain principle of back-propagation and why it is necessary. How is it implemented in computer?
In the previous subsection \ref{sec.structure} the fundamental structure is presented. It is shown, how the input $\boldsymbol{x}$ propagates trough the network and is computed, so that the output $\boldsymbol{\hat{y}}$ results. This process is called \textit{forward propagation}. During training, the forward propagation of the training data $\boldsymbol{x}$ results in the estimated output $\boldsymbol{\hat{y}}$. For convenience, the simple model from eq. \ref{eq:hidden_layer} and \ref{eq:output_layer} are shown again extended with the loss $l$ and modeled the ReLu function as a single step:

\begin{equation}
\begin{split} \label{eq:simple_model1}
\boldsymbol{h^{\prime}} &= \boldsymbol{W}_{xh}\boldsymbol{x}+\boldsymbol{b}_h \\
\boldsymbol{h} &= ReLu(\boldsymbol{h^{\prime}}) \\
\hat{y} &= \boldsymbol{w}_{hy}\boldsymbol{h}+b_y \\
l &= (\hat{y} - y)^2\\
\end{split}
\end{equation}

The derivatives of this model can be calculated by a method called \textit{back propagation}. With the chain rule the derivatives can be calculated for every set of weights. 

\begin{equation}
\begin{split} \label{eq:derivatives}
\frac{dl}{db_y} &= \frac{dl}{d\hat{y}}\frac{d\hat{y}}{db_y} \\	
\frac{dl}{d\boldsymbol{w}_{hy}} &= \frac{dl}{d\hat{y}}\frac{d\hat{y}}{d\boldsymbol{w}_{hy}} \\	
\frac{dl}{d\boldsymbol{b}_h} &= \frac{dl}{d\hat{y}}\frac{d\hat{y}}{d\boldsymbol{h}}\frac{d\boldsymbol{h}}{d\boldsymbol{h^{\prime}}}\frac{d\boldsymbol{h^{\prime}}}{d\boldsymbol{b}_h} \\	
\frac{dl}{d\boldsymbol{W}_{xh}} &= \frac{dl}{d\hat{y}}\frac{d\hat{y}}{d\boldsymbol{h}}\frac{d\boldsymbol{h}}{d\boldsymbol{h^{\prime}}}\frac{d\boldsymbol{h^{\prime}}}{d\boldsymbol{W}_{xh}} \\	
\end{split}
\end{equation}

In eq. \ref{eq:derivatives} the derivatives for all the weights regarding the loss function are presented. The chain rule allows to calculate the derivatives for every single step in the back propagation and multiply the results to get the derivatives for every single weight.
The derivations can become very complicated for larger networks so that it is impossible to calculate them by hand. Fortunately, it is possible to let the computer calculate the derivatives by using \textit{automatic differentiation (autodiff)}, where the feedforward propagation is modeled in many single steps (e.g addition, multiplication, square, ...) and the single derivatives are computed. So it is possible to create large networks and provide for every single neuron the forward function and the backward derivative computation \cite{neubig_neural_2017}.  

\begin{figure}[h]
	\centering
	\includegraphics[width=0.8\textwidth]{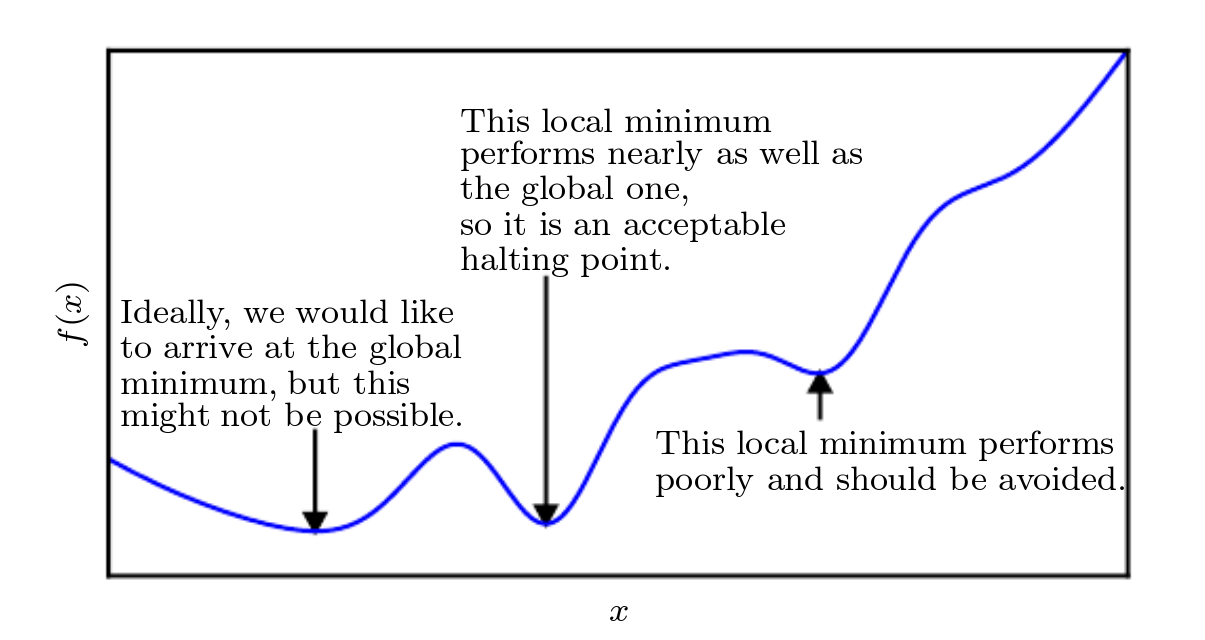}
	\caption[Multiple minima in an optimization problem] {Multiple minima in an optimization problem \cite{goodfellow_deep_2016}.} \label{bild.multiple_minima}
\end{figure}

Once the forward propagation and the derivatives can be provided, the numerical gradient-based optimization algorithm is able to adjust the weights and minimize the loss. 
The fundamental form of the gradient descent algorithm is 
\begin{equation}
\boldsymbol{x^{\prime}} = \boldsymbol{x} - \epsilon\boldsymbol{\nabla}_{\boldsymbol{x}}f(\boldsymbol{x})
\end{equation}

where $\boldsymbol{x^{\prime}}$ is the new approximation for the minimum, $\epsilon$ is the \textit{learning rate} and $\boldsymbol{\nabla}_{\boldsymbol{x}}f(\boldsymbol{x})$ is the gradient of $f$ which contains the partial derivatives. The learning rate is a fundamental \textit{hyperparameter}. Hyperparameters are parameters which control the learning algorithm. Further down in this section a few more will be introduced. 
The learning rate is important, since it has high influence on the training algorithm. If $\epsilon$ is too low, it takes a lot of training steps to reach the global minimum, which is highly inefficient. Furthermore it is possible, that the algorithm only finds a local minimum but not a global. Multidimensional optimization problems often have more than one minimum. This is visualized for a simple one dimensional case in fig. \ref{bild.multiple_minima}.

If the learning rate is too high, it is possible that the algorithm cannot find an acceptable minimum either, since the algorithm "jumps" over the minimum. 
The probably most common and used optimization algorithm for machine and deep learning is \textit{stochastic gradient descent} (SGD), which is a modification of the gradient descent method. Instead of computing the gradient based on only one example, SGD uses a minibatch of multiple examples to compute the average gradient. This results in a better generalization of the knowledge, since more parallel information are taken into consideration. Addressing the problem of the right choice of the learning rate, algorithms with adaptive learning rates have been developed. The most important in the field of deep learning are \textit{AdaGrad}, \textit{RMSProp} and \textit{Adam}. To present the algorithms in detail is beyond the scope of this thesis. \cite{goodfellow_deep_2016} gives a good overview of this topic. 
%show graphic with nice optimization algorithm and stepsize and so on... (implement maybe later, but no musthave!)

\begin{figure}[h]
	\centering
	\includegraphics[width=1\textwidth]{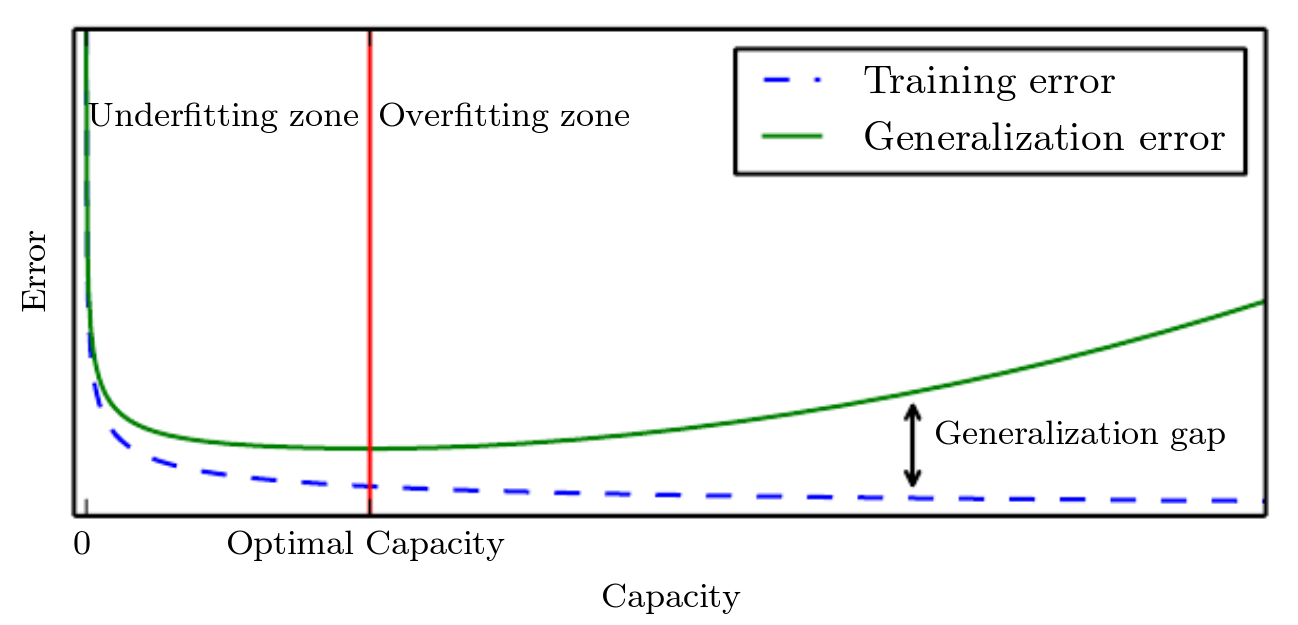}
	\caption[Underfitting and overfitting regarding model capacity] {Underfitting and overfitting regarding model capacity \cite{goodfellow_deep_2016}.} \label{bild.Loss_over_capacity}
\end{figure}

One of the main challenges when it comes to train neural networks is the phenomenon of \textit{underfitting} and \textit{overfitting}. Before training the model, the entire dataset is split into the \textit{training data} and \textit{test data}. The model is trained with the training data and afterwards the \textit{training loss} can be calculated. Furthermore, the model is tested on the test data it has never seen before during training. For the test session, the tests loss could be calculated as well. The ability of a model to make good predictions on data it has never observed before is called \textit{generalization}. The losses for training and testing over the \textit{model capacity} is shown in fig. \ref{bild.Loss_over_capacity} for a specific problem, which is not trivial but also not too complex. 
Simplified, the model capacity describes the complexity of the model regarding the number of hidden layers, the width and other structural properties. Models with a low capacity are not able to solve more complex problems. This leads to underfitting, where the training loss as well as the test loss is very high. The model cannot generalize knowledge, since it cannot even learn the information provided by the training data. Overfitting occurs if the model capacity is too high for the complexity of the given problem. The model has enough parameter to learn the whole training set, which results in a low training loss. The test loss is high though, because the model memorizes learned properties from the individual test data, which are not relevant for solving the problem. 
\begin{figure}[h]
	\centering
	\includegraphics[width=0.6\textwidth]{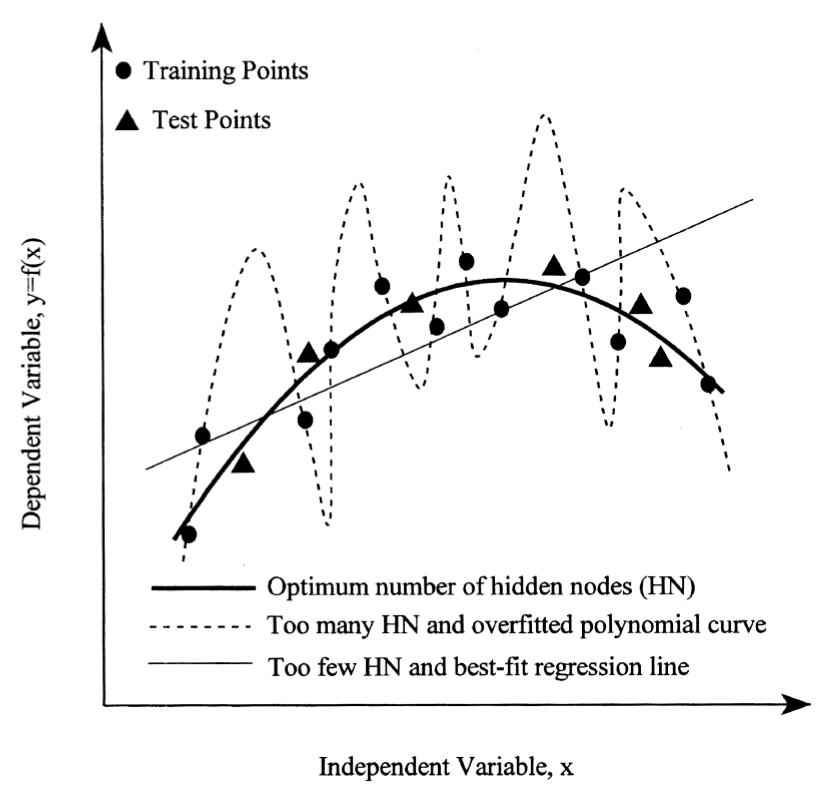}
	\caption[Visualization of the impact of underfitting and overfitting] {Visualization of the impact of underfitting and overfitting \cite{basheer_artificial_2000}.} \label{bild.under_overfitting}
\end{figure}
In this case, the model cannot generalize the knowledge either. This phenomenon is presented in detail in fig. \ref{bild.under_overfitting}.
For easier understanding of the figure, the model can be interpreted as a polynomial function. If the model complexity is too low (in this case polynomial of order one), the curve cannot fit the training points and also fails for an accurate prediction of the test points (both training and test loss are high). This is the classic case of underfitting. When the model complexity is increased to a polynomial function of second order, it fits quiet well for the training data but still has a rest loss, which is not zero. It also has a low loss for the test points, which is in this case the optimal solution. If the order of the polynomial increases even more (e.g order nine), the curve fits trough all the training points and the training loss is zero. But when it comes to predict the test points, the graph does not fit anymore and the test loss is very high. Then the model is overfitted. 

It is crucial to find the optimal model capacity to avoid underfitting or overfitting. The necessary complexity of the model highly depends on the individual problem to be solved. So there is no other way than using the trial-and-error method and maybe having experience from other applications. In addition to that, it is also possible (and recommended) to use regularization techniques during the training. Regularization has the aim to avoid overfitting and reduce the test loss on cost of an increasing train loss. In this section, only two (for this thesis relevant) regularization methods are briefly introduced. The first is \textit{parameter norm penalties}, in particular the $L^2$ \textit{parameter regularization}. 
\begin{figure}[h]
	\centering
	\includegraphics[width=0.6\textwidth]{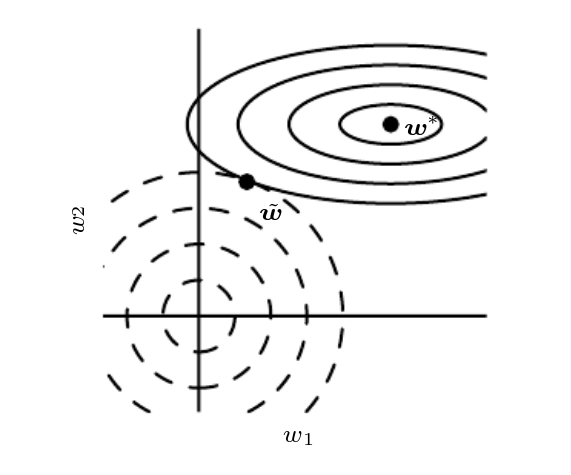}
	\caption[Visulaization of the $L^2$ regularization] {Visualization of the $L^2$ regularization \cite{goodfellow_deep_2016}. The point $\boldsymbol{w}^*$ represents the optimal parameter $w_1$ and $w_2$ to fit the training data. The dotted circles represent the increasing value of the penalty term with increasing distance to the origin. The optimum can be found in the point $\tilde{\boldsymbol{w}}$.} \label{bild.L2_regularization}
\end{figure}
Generally, for parameter regularization a penalty term is added to the loss functions which contains the weights $\boldsymbol{W}$ of each hidden layer. In case of the $L^2$ regularization the penalty term is $\Omega = \frac{1}{2} \lVert \boldsymbol{W} \rVert_2^2$. This additional term reduces high weights and overfitting to a particular training set. This method is also called \textit{weight decay}. 
In fig. \ref{bild.L2_regularization} the influence of the parameter regularization is presented.
Shown are the weights $\boldsymbol{w}^*$ which represents the parameter which would optimal fit the training dataset. The ellipses around this point visualize the increasing loss function without the penalty term for the parameter $w_1$ and $w_2$. The dotted circles represent the increasing value of the penalty term based on $w_1$ and $w_2$. The optimum between both the ellipses and dotted circles can be found in the point $\tilde{\boldsymbol{w}}$. In this way the $L^2$ regularization avoids the overfitting of the parameters to the point $\boldsymbol{w}^*$.
Another common regularization technique is \textit{dropout}. When dropout is applied during training, some weights are randomly set to zero. This creates a subnetwork of the original neural network which is then trained. In the next training step other weights are randomly chosen to be set to zero. This technique reduces on the one hand the capacity of the model which results in a need for higher number of nodes which requires more computational effort. On the other hand it reduces overfitting, since the structure of the neural network changes slightly in every training step. 

%\begin{equation}
%	\forall \; simulation, \forall \; timestep < t
%\end{equation}

The here briefly presented methods are fundamental for creating and training a deep learning model. However, deep learning is a large topic with many different applications and methods. A very good overview and introduction gives \cite{goodfellow_deep_2016}. In this thesis the focus in the following sections is on attention mechanisms and sequence to sequence problems in deep learning. Since in this approach the model uses data which have spatial correlations (lanes are connected to each other), it is beneficial to use attention mechanisms to let the model focus on relevant information.

\subsection{Graph Attention Networks} \label{sec:gat_layer}
The here presented \textit{graph attention networks} are designed to perform on geometric graphs which consist of \textit{nodes} and \textit{edges}. Each node is connected with other nodes by edges. Examples for graphs are social networks, road networks or 3D meshes. The structure does not necessarily have to be a grid-like structure. 
For solving tasks where it is possible to model correlations between data in a graph structure, it is often helpful to consider not only a single node, but its neighboring nodes. If the algorithm uses deep learning attention mechanisms for that, this process is called \textit{graph attention}. In this section a new method for learning attention for a graph, based on \cite{velickovic_graph_2017}, is presented. Described is a single \textit{graph attentional layer (GAT)}, which can be used to build larger graph attention networks. The input for this layer is a matrix which consists of every node of the graph including its features $\boldsymbol{h}_i$

\begin{equation}
\boldsymbol{H} = \{\boldsymbol{h}_1,\boldsymbol{h}_2,...,\boldsymbol{h}_N\},\boldsymbol{h}_i \in \R^F, \boldsymbol{H} \in \R^{N \times F}.
\end{equation}

where $N$ is the number of nodes and $F$ is the number of features. The second input for the GAT layer is the information about the connections of the nodes in the graph. It is represented by the \textit{adjacency matrix}

\begin{equation}
\boldsymbol{A} \in \{0, 1\}^{N \times N}.
\end{equation}

$\boldsymbol{A}$ is a binary matrix with the size number of nodes by number of nodes. If a node has a connection to another node, the value at this specific place in the matrix is one, otherwise it is zero. Before computing the output features for every node, it is necessary to compute \textit{attention coefficients} $e_{ij}$ for every node $i$ and its corresponding neighbors $j$ in the \textit{neighborhood} $\mathcal{N}_i$. The neighborhood $\forall \boldsymbol{h}_i: \mathcal{N}_i$ can be extracted by using the adjacency matrix. The single attention coefficients are calculated by

\begin{equation}
e_{ij} = \sigma (\boldsymbol{a}^T [\boldsymbol{Wh}_i \lVert \boldsymbol{Wh}_j])
\end{equation}

with the \textit{weight matrix} $\boldsymbol{W} \in \R^{F^{\prime} \times F}$, where $F^{\prime}$ is the number of the new output features, and a \textit{weight vector} $\boldsymbol{a} \in \R^{2F^{\prime}}$. Furthermore, the $\lVert$ symbol represents the concatenation of the results of both matrix multiplications $\boldsymbol{Wh}_i$ for the node itself and $\boldsymbol{Wh}_j$ for the neighboring node. Afterwards the attention coefficients are normalized by a \textit{softmax} function, which allows a comparison between the coefficients:

\begin{equation}
\alpha_{ij} = \frac{exp(e_{ij})}{\sum_{k \in \mathcal{N}_i} exp(e_{ik})}.
\end{equation}

After the normalization, it is possible to calculate the new hidden state of the node $i$ by executing one learnable linear transformation and weighting it with the normalized attention coefficient for every node in the neighborhood of the node (including the node itself):

\begin{equation}
\boldsymbol{h}^{\prime}_i = \sigma \Big( \sum_{j \in \mathcal{N}_i} \alpha_{ij} \boldsymbol{Wh}_j  \Big) ; \quad \boldsymbol{h}^{\prime}_i \in \R^{F^{\prime}}
\end{equation}

Generally, the normalized attention coefficient $\alpha_{ij}$ decides, how much the former features from node $j$ contributes to the new hidden state $\boldsymbol{h}_i^{\prime}$ of node $i$. 

\begin{figure}[h]
	\centering
	\includegraphics[width=0.65\textwidth]{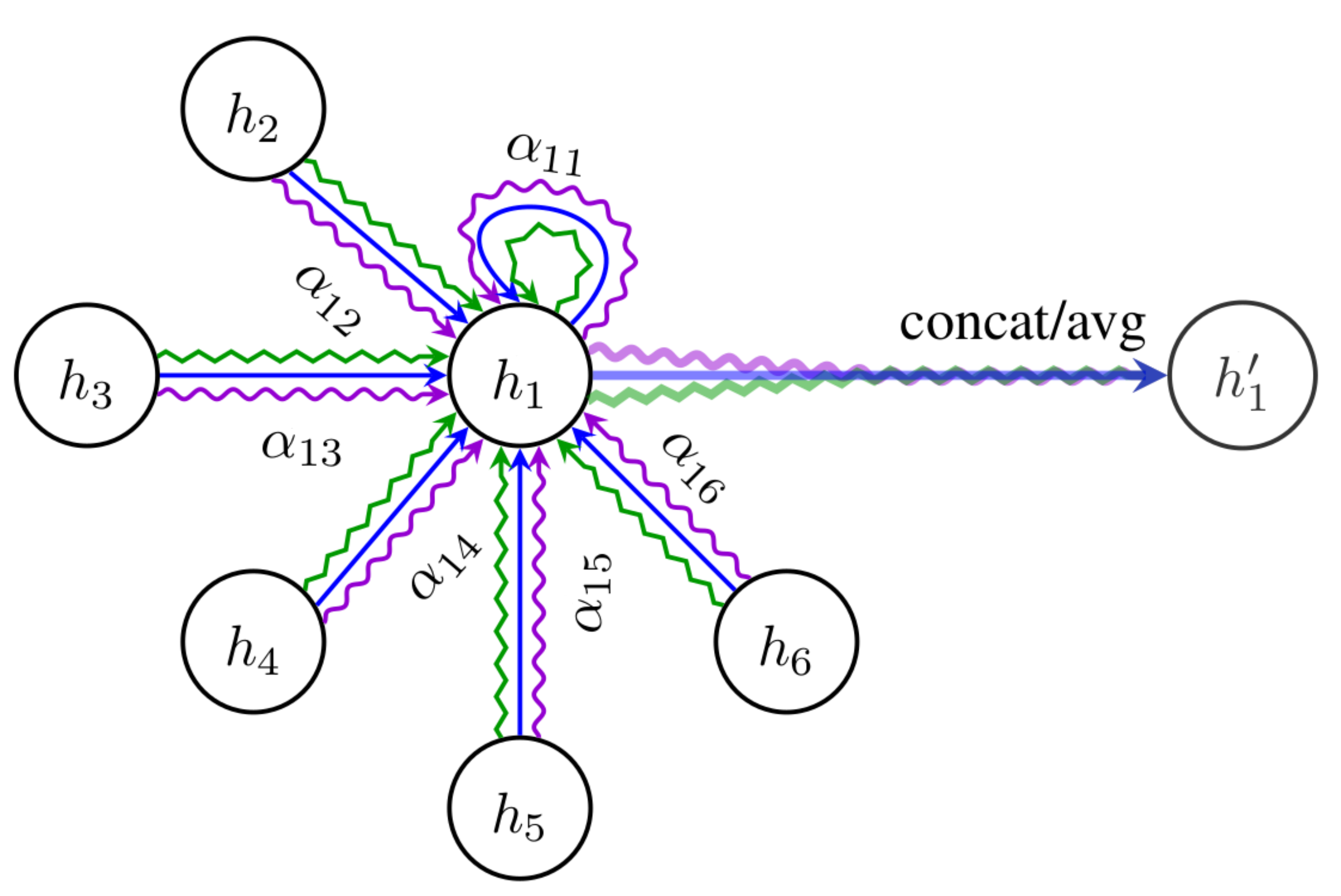}
	\caption[Visualization of graph attention mechanism with tree attention heads] {Visualization of graph attention mechanism with three attention heads \cite{velickovic_graph_2017}.} \label{bild.annt_mechanism}
\end{figure}

Furthermore, \cite{velickovic_graph_2017} notes that it can be helpful to execute the calculation of attention coefficients and new hidden states multiple times independently from each other. By that the attention learning process can be more stabilized. This is then called "\textit{multi-head attention}" \cite{velickovic_graph_2017}. After the execution of $K$ different independent estimations with the attention mechanism, one way to process the results is to concatenate them:

%formula concatenating
\begin{equation}
\boldsymbol{h}^{\prime}_i = \overset{K}{\underset {k=1}{\lVert}} \;
\sigma \Big( \sum_{j \in \mathcal{N}_i} \alpha_{ij}^k \boldsymbol{W}^k\boldsymbol{h}_j  \Big); \quad \boldsymbol{h}^{\prime}_i \in \R^{F^{\prime}}
\end{equation}

where $k$ stands for each single attention head. Also each independent attention mechanism has its own attention coefficients $\alpha_{ij}^k$ and weight matrix $\boldsymbol{W}^k$. When using concatenation of the single results, the final number of features will be $KF^{\prime}$ instead of $F^{\prime}$. Another option to process the single results from each attention head is to average over $K$

%formula averaging
\begin{equation}
\boldsymbol{h}^{\prime}_i = 
\sigma \Big( \frac{1}{K} \sum_{k=1}^{K} \sum_{j \in \mathcal{N}_i} \alpha_{ij}^k \boldsymbol{W}^k\boldsymbol{h}_j  \Big); \quad \boldsymbol{h}^{\prime}_i \in \R^{KF^{\prime}}
\end{equation}

where all the single results are summed-up and divided by the number of attention heads. In this case, the number of output features per node is $F^{\prime}$.
In fig. \ref{bild.annt_mechanism} a visualization of the attention mechanism with three attention heads is presented. In this picture $\boldsymbol{h}_1$ are the old features of node one, while $\boldsymbol{h}_2$ to $\boldsymbol{h}_6$ are the neighboring nodes. Each of them contributes, scaled by the attention coefficients $\boldsymbol{\alpha}_{1j}$, to the new hidden state $\boldsymbol{h}_1^{\prime}$. The different parallel attention heads are visualized by the colored arrows. 
Finally, the output of the GAT layer is a matrix which contains all the features for each node in the graph:

\begin{equation}
\boldsymbol{H}^{\prime} = \{\boldsymbol{h}_1^{\prime},\boldsymbol{h}_2^{\prime},...,\boldsymbol{h}_N^{\prime}\},\boldsymbol{h}_i^{\prime} \in \R^{F^{\prime}}, \boldsymbol{H}^{\prime} \in \R^{N \times F^{\prime}}
\end{equation}

Based on the presented single GAT layer, more complex networks can be build by using more than one graph attention layer in a network. 
%Specifically for solving the problem of queue length estimation, the GAT layer can help to take traffic data from other connected lanes into consideration. This can help to make the estimation more accurate. 

\subsection{Recurrent Neural Networks (RNN)} \label{sec:RNN}

All networks that have been presented in the previous chapters were feedforward neural networks. The information from the input propagates through the network and generates an output. Every new sample as input for the deep learning model is independent from the previous or following one. Also the order of the input-examples is mostly not very relevant. 
This becomes different when the input changes to sequences. 

\begin{figure}[h]
	\centering
	\includegraphics[width=1\textwidth]{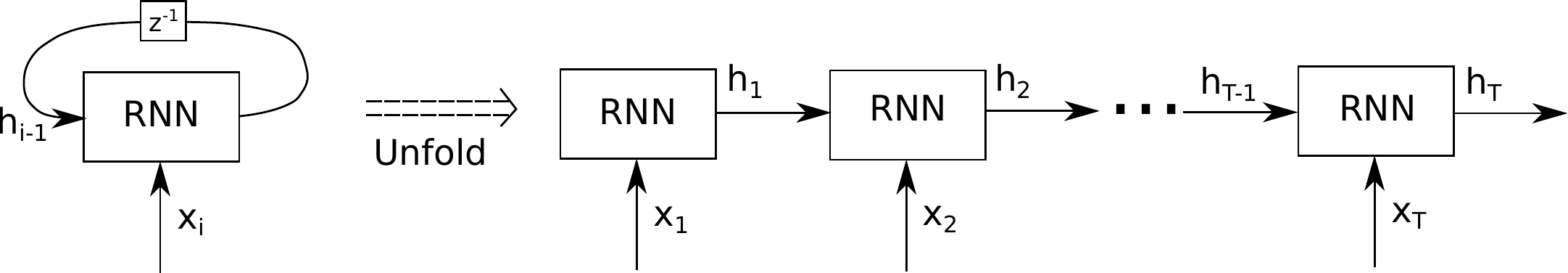}
	\caption[Unfold RNN] {Circuit diagram of the RNN (right) and its unfolded visualization (left) after  \cite{goodfellow_deep_2016}.} \label{bild.RNN_unfold}
\end{figure}

Sequences can be for instance sentences (sequence of words) and the task is to translate this sequence into another language. So the input is a sequence and the output is a sequence as well. The models which are used to solve this task are called \textit{sequence-to-sequence (seq2seq) models} \cite{neubig_neural_2017}. Another example are time sequences, like in the approach which is made in this thesis. The input is a time sequence of loop-detector data and the output is the queue length in form of a time sequence as well. In the case of sequences, the single steps of the sequence are not independent from each other anymore but correlated to each other. For example in the sentence "\textit{He} loves \textit{his} red bike", the third word "his" is correlated to the first word "He". It matches regarding the gender. So a deep learning model structure is necessary which reads in every single step, but does not treat them independently, but as a sequence. Therefore, \textit{recurrent neural networks (RNN)} were developed, which are able to process sequential data. These family of networks have an internal feedback loop, where the calculated hidden state $\boldsymbol{h}_{i-1}$ from the previous step is fed back as an input of the model to calculate the new hidden state $\boldsymbol{h}_i$ for the current step. This allows the network to take also information from previous steps into consideration. In fig. \ref{bild.RNN_unfold} the general topology of a RNN is presented. 

On the left site, the circuit diagram of the RNN is presented, where for the time step $i$ the RNN has as input $\boldsymbol{x}_i$ as well as the previous hidden state $\boldsymbol{h}_{i-1}$. In this case the RNN has no other output than its hidden state, but it is also possible to add another layer to compute an output. This is presented later in section \ref{sec:enc-dec}. On the right side, the unfold visualization with every time step is presented. It is important to note, that the recurrent neural network is for every time step the same with the same weights \cite{goodfellow_deep_2016}. 
The general equation to calculate the new hidden state is

\begin{equation}
\boldsymbol{h}_i = \sigma(\boldsymbol{W}_{xh} \boldsymbol{x}_i + \boldsymbol{W}_{hh}\boldsymbol{h}_{i-1}+\boldsymbol{b}_h).
\end{equation}

The only difference between a normal feedforward neural network (see eq. \ref{eq:normal_NN}) and a recurrent neural network is the term $\boldsymbol{W}_{hh}\boldsymbol{h}_{i-1}$. The first hidden state $\boldsymbol{h}_0$ has to be initialized, which is usually a vector of zeros. The reason why recurrent neural networks can deal with long distance dependencies is, that all relevant information of previous time steps are stored in the hidden state $\boldsymbol{h}_{i-1}$ \cite{neubig_neural_2017}. 

\begin{figure}[h]
	\centering
	\includegraphics[width=1\textwidth]{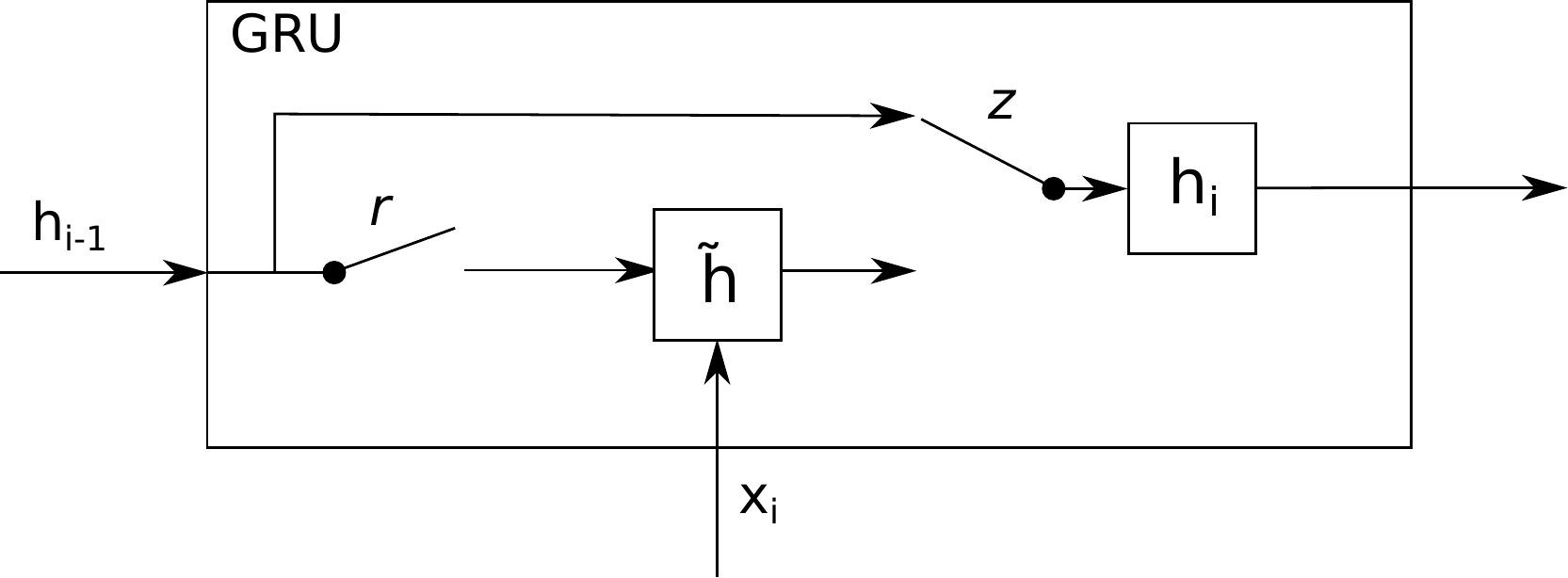}
	\caption[Visualization of a gated recurrent unit (GRU)] {Visualization of a gated recurrent unit (GRU) with the reset gate $\boldsymbol{r}$, update gate $\boldsymbol{z}$ and proposal hidden state $\boldsymbol{\tilde{h}}$ after  \cite{cho_learning_2014}. } \label{bild.GRU}
\end{figure}

The presented structure above is very simple and a straight forward development from a feedforward neural network. However, it has limitations due to the \textit{vanishing gradient} or \textit{exploding gradient} problem. The procedure to train a RNN is the same than for a normal neural network, which includes the forward propagation based on the input and following the back propagation using the gradients to optimize the weights. But in case of a RNN, the back propagation includes all previous time steps of the input sequence. If $\frac{d\boldsymbol{h}_{i-1}}{d\boldsymbol{h}_i}$ is not exactly one, the gradient between the hidden state at time step $i$ and the loss function $\frac{dl}{d\boldsymbol{h}_i}$ becomes either very small or very large. That means for a very long sequence, it is possible that the gradient at the beginning of the sequence is too small to have an effect to update the weights. 
The most popular solution for solving the vanishing and exploding gradient problem was introduced by \cite{hochreiter_long_1997} in 1997 and called \textit{long-short-term-memory (LSTM)}. It uses parallel to the hidden state $\boldsymbol{h}_i$ a \textit{memory cell} $\boldsymbol{c}$ which long and short term dependencies can be stored. Since $\frac{d\boldsymbol{c}_{i-1}}{d\boldsymbol{c}_i}$ is exactly one, no vanishing or exploding of the gradients occur. The LSTM can control which knowledge to keep in the memory cell and which knowledge to delete by using so called \textit{gates}. It uses \textit{input} and \textit{output} \textit{gates}, which both consist of an affine transformation with a following activation function. So the LSTM can learn, how to manage the information and which knowledge to keep over time and which not. LSTM had great success in solving seq2seq problems during the past years \cite{neubig_neural_2017}. 
However, in this thesis another variation of a recurrent neural network, which is called \textit{gated recurrent unit (GRU)}, will be used. It was first presented in \cite{cho_learning_2014} in 2014 and can be described as a simplification of an LSTM. It has no memory cell $\boldsymbol{c}$ but only uses the hidden state $\boldsymbol{h}$ for storing long term dependencies. It also has only two gates which are the \textit{update gate} $\boldsymbol{z}$ and the \textit{reset gate} $\boldsymbol{r}$. Although it is a simplification, the performance regarding sequence processing can be similar to an LSTM \cite{chung_empirical_2014}. The main structure is presented in fig. \ref{bild.GRU}. 

The hidden state from the previous time step $\boldsymbol{h}_{i-1}$ as well as the current time step from the input sequence $\boldsymbol{x}_i$ are fed into the GRU. The first step is to calculate the proposed new hidden state

% calculate new proposal state
\begin{equation}
\boldsymbol{\tilde{h}}_i = tanh(\boldsymbol{Wx}_i+\boldsymbol{U}[\boldsymbol{r}_i \circ \boldsymbol{h}_{i-1}]+\boldsymbol{b})
\end{equation}

by using $\boldsymbol{h}_{t-1}$ and $\boldsymbol{x}_i$. The input $\boldsymbol{x}_i$ is multiplied with the weight matrix $\boldsymbol{W}$. Furthermore, $\boldsymbol{U}$ weights the element-wise multiplication ($\circ$) between the reset gate $\boldsymbol{r}_i$ and $\boldsymbol{h}_{i-1}$. The reset gate controls, which information of the previous hidden state contributes to the proposal state. The reset gate can be calculated as

%reset gate r_i
\begin{equation}
\boldsymbol{r}_i = \sigma (\boldsymbol{W}_r \boldsymbol{x}_i + \boldsymbol{U}_r \boldsymbol{h}_{i-1}+ \boldsymbol{b}_r).
\end{equation}

The calculation of the new hidden state is

%update new hidden state
\begin{equation}
\boldsymbol{h}_i = (1-\boldsymbol{z}_i)\circ \boldsymbol{h}_{i-1}+\boldsymbol{z}_i \circ \boldsymbol{\tilde{h}}_i; \quad \boldsymbol{h}_0 = \boldsymbol{0}
\end{equation}

where the update gate determines the ratio between the contribution of the previous and the proposal hidden state. The update gate can be calculated as

% update gate z_i
\begin{equation}
\boldsymbol{z}_i = \sigma (\boldsymbol{W}_z \boldsymbol{x}_i + \boldsymbol{U}_z \boldsymbol{h}_{i-1} + \boldsymbol{b}_z)
\end{equation}

with the weight matrices $\boldsymbol{W}_z$ and $\boldsymbol{U}_z$. Due to this structure, the GRU is able to solve tasks with long time dependencies in a sequence and still having a quiet simple model complexity \cite{cho_learning_2014}\cite{bahdanau_neural_2014}. Gated recurrent units are the main component for the processing sequences with an encoder-decoder structure, which is presented in the following section.

\subsection{RNN Encoder-Decoder} \label{sec:enc-dec}

\begin{figure}[h]
	\centering
	\includegraphics[width=1\textwidth]{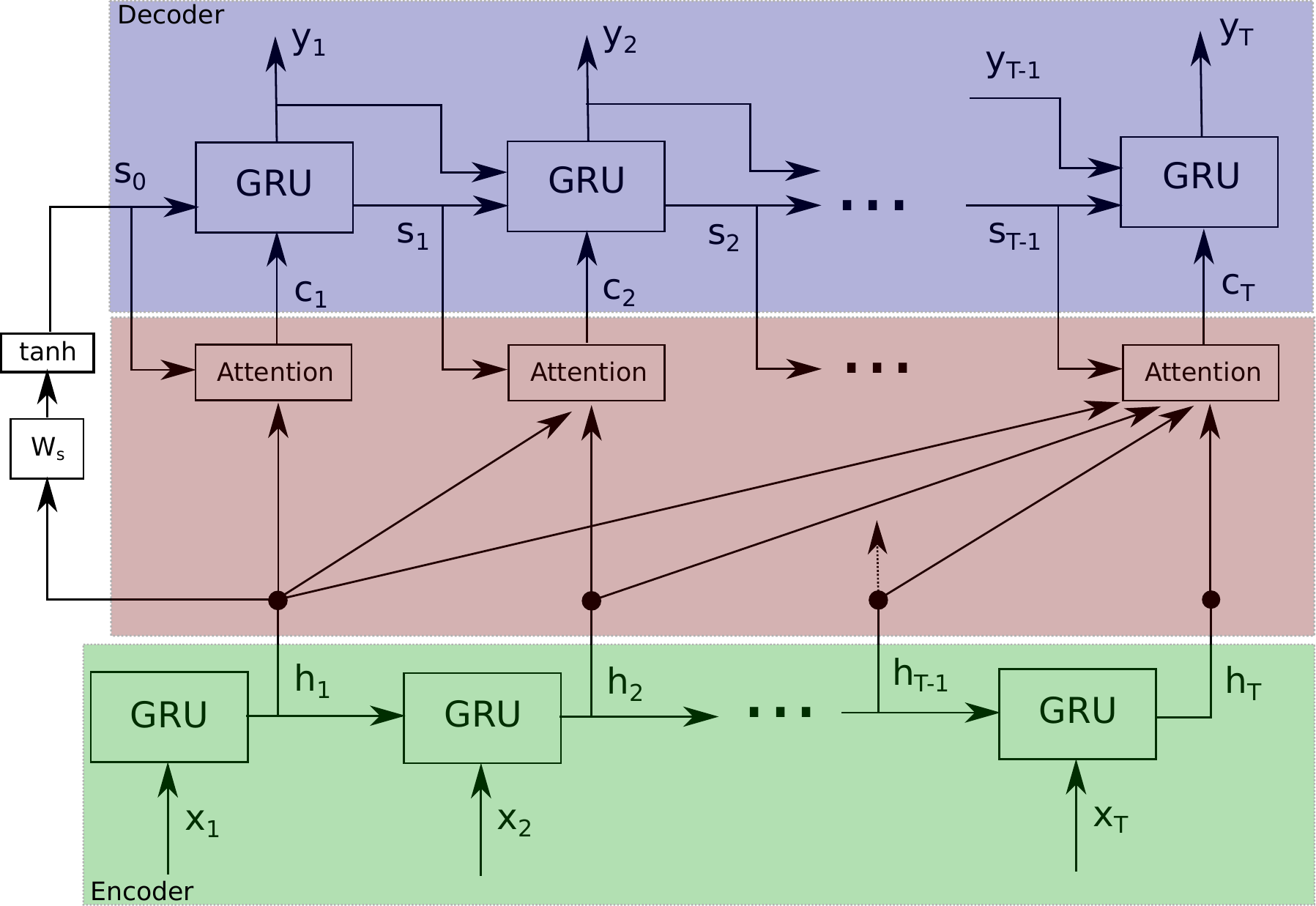}
	\caption[Main structure of the RNN Encoder-Attention Decoder] {Main structure of the RNN Encoder-Attention Decoder in unfolded visualization. The encoder is presented in green, the attention mechanism in red and the decoder in blue \cite{cho_learning_2014}. } \label{bild.RNN_enc_dec}
\end{figure}

The best approach to solve sequence-to-sequence models, is to use an \textit{RNN Encoder-Decoder} structure. Generally, two recurrent neural networks are sequentially used. The first RNN is the encoder, which gets a sequence as input and encodes it into hidden states. The decoder maps the information in the hidden states back into the output sequence. This algorithm is presented in \cite{cho_learning_2014}. In this thesis both RNN, the encoder and decoder, are GRU. In addition to that, in this thesis an \textit{attention decoder} based on \cite{bahdanau_neural_2014} is used. The attention mechanism is supposed to learn temporal attention within the time sequence. The main structure of the Encoder-Attention Decoder is presented in fig. \ref{bild.RNN_enc_dec}.

The encoder takes the input sequence $\boldsymbol{x}_1...\boldsymbol{x}_T$ and encodes it into hidden states for every time step. This process is equal to the GRU presented in the previous section \ref{sec:RNN}. The hidden state $\boldsymbol{h}_i$ has information about the current and all the previous inputs. The initial state $\boldsymbol{h}_0$ is chosen to a vector of zeros. 
The attention mechanism takes all the hidden states from the previous and current time step $\boldsymbol{h}_1 ... \boldsymbol{h}_i$ as well as the previous hidden state from the decoder $\boldsymbol{s}_{i-1}$ and calculates the \textit{context vector} $\boldsymbol{c}_i$ for every time step as

%context vector
\begin{equation}
\boldsymbol{c}_i = \sum_{j=1}^{i}\alpha_{ij}\boldsymbol{h}_j
\end{equation}

with the normalized attention coefficients $\alpha_{ij}$ for every previous and the current time step $1...i$ regarding the current time step $i$. In the convex combination $\alpha_{ij}$ decides, how much the hidden state from each time step contributes to the calculation of the new output by the decoder. The original source \cite{bahdanau_neural_2014} takes all hidden states from all time steps (past and future) into consideration, since the original application is in the field of language translation, where the full sequence is already known in advance. In this case, the attention mechanism takes only the time steps from the past into consideration, since it is a time sequence and the future time steps are unknown. 
The normalized attention coefficients are calculated as 

%normalization of attention coefficients
\begin{equation}
\alpha_{ij} = \frac{exp(e_{ij})}{\sum_{k=1}^{T} exp(e_{ik})}
\end{equation}
with the attention coefficients
%attention coefficients
\begin{equation}
e_{ij} = \boldsymbol{v}_a^T tanh(\boldsymbol{W}_a \boldsymbol{s}_{i-1} + \boldsymbol{U}_a \boldsymbol{h}_j).
\end{equation}

The previous hidden state from the decoder $\boldsymbol{s}_{i-1}$ is multiplied by the weight matrix $\boldsymbol{W}_a$ and the the hidden state at time step $j$ from the encoder is multiplied by $\boldsymbol{U}_a$. After applying a $tanh$ function as activation function, the vector $\boldsymbol{v}_a$ scales the term down to a scalar which results in the attention coefficient $e_{ij}$. Note that this attention mechanism is similar to the spatial attention for the GAT layer in section \ref{sec:gat_layer}. 

Now the sequnce is encoded into hidden states and the attention mechanism selects the relevant hidden states from the past time steps. The decoder, which consists also of an GRU, takes the context vector $\boldsymbol{c}_i$ in each time step and calculates, based on the previous hidden state $\boldsymbol{s}_{i-1}$ and the previous output $\boldsymbol{y}_{i-1}$, the new hidden state $\boldsymbol{s}_{i}$ and the current output $\boldsymbol{y}_{i}$. Generally, the decoder has the same structure as the encoder. The new hidden state is calculated as

\begin{equation}
\boldsymbol{s}_i = (1-\boldsymbol{z}_i) \circ \boldsymbol{s}_{i-1} + \boldsymbol{z}_i \circ \boldsymbol{\tilde{s}}_i
\end{equation}
where the proposal hidden state, the update gate and the reset gate are calculated as
\begin{equation}
\begin{split}
\boldsymbol{\tilde{s}}_i =& tanh(\boldsymbol{W}^{\prime} \boldsymbol{y}_{i-1} + \boldsymbol{U}^{\prime} [\boldsymbol{r}_i \circ \boldsymbol{s}_{i-1}] + \boldsymbol{C}^{\prime} \boldsymbol{c}_i + \boldsymbol{b}^{\prime}) \\
\boldsymbol{z}_i = &\sigma(\boldsymbol{W}_z^{\prime} \boldsymbol{y}_{i-1} + \boldsymbol{U}_z^{\prime} \boldsymbol{s}_{i-1} + \boldsymbol{C}_z^{\prime} \boldsymbol{c}_i + \boldsymbol{b}_z^{\prime}) \\
\boldsymbol{r}_i = &\sigma(\boldsymbol{W}_r^{\prime} \boldsymbol{y}_{i-1} + \boldsymbol{U}_r^{\prime} \boldsymbol{s}_{i-1} + \boldsymbol{C}_r^{\prime} \boldsymbol{c}_i + \boldsymbol{b}_r^{\prime}).
\end{split}
\end{equation}

The notation $(\cdot)^{\prime}$ at the weight matrices means, that these are different to the weight matrices for the encoder in section \ref{sec:RNN}. Furthermore, the context vector $\boldsymbol{c}$ is taken into consideration and weighted by the matrices $\boldsymbol{C}^{\prime}$, $\boldsymbol{C}_z^{\prime}$ and $\boldsymbol{C}_r^{\prime}$. The initial input vector $\boldsymbol{y}_0$ is set to a zero-vector. The initial hidden state for the decoder is determined as $\boldsymbol{s}_0 = tanh(\boldsymbol{W}_s^{\prime} \boldsymbol{h}_1)$, where $\boldsymbol{W}_s^{\prime}$ is a trainable weight matrix. Except of this modification the calculation of the new hidden state is similar to the presented GRU in section \ref{sec:RNN}. To calculate the output for the specific time step $i$, the decoder GRU is extended with a single output layer:

\begin{equation}
\boldsymbol{y}_i = \sigma(\boldsymbol{W}_o^{\prime} \boldsymbol{y}_{i-1} + \boldsymbol{U}_o^{\prime} \boldsymbol{s}_{i-1} + \boldsymbol{C}_o^{\prime} \boldsymbol{c}_i + \boldsymbol{b}_o).
\end{equation}

This layer computes, based on the previous output $\boldsymbol{y}_{i-1}$, hidden state $\boldsymbol{s}_{i-1}$ and current context vector $\boldsymbol{c}$ the current output for time step $i$ which is part of the output sequence \cite{bahdanau_neural_2014}.

In this chapter the basics of feedforward and recurrent neural networks as well as the RNN encoder-decoder algorithm is presented. Based on this methods, it is possible to create an approach for solving the task of queue length estimation in road networks. Also the presented attention mechanisms for spatial correlation (see GAT layer in section \ref{sec:gat_layer}) and temporal correlations in the sequence (this section) help to be potential superior to classic approaches like the method after \cite{liu_real-time_2009} presented in chapter \ref{sec.liu_method}.
%----------- Ende Abschnitt neu eingefügt ---------

\chapter{Scientific Goals} \label{sec.scientific_goals}

Based on the theoretical background of both the conventional and the deep learning approach presented in chapter \ref{sec.stateofart}, it is possible to formulate the scientific goals for this work. 
The main goal is to develop and present a geometric deep learning algorithm which is superior to the conventional state of the art method of \cite{liu_real-time_2009}, regarding the performance of queue length estimation in front of signalized intersections. It is to verify if the usage of traffic data of multiple lanes and the consideration of spatial as well as temporal correlations between measurements of each lane is beneficial for the estimation performance.  It is crucial, that the geometric deep learning approach relies only on loop detector and traffic light data, as well as the state of the art method. Otherwise, no direct comparison between both methods would be possible. 

To reach this main goal, some secondary objectives have to be fulfilled. It is fundamental to find a reliable metric for validation of the performance. Generally, there are two options: The \textit{mean average percentage error (MAPE)} and the \textit{mean average error (MAE)}. Both measure the performance of the estimation method compared to the ground-truth data, but the MAPE provides the percentage error while the MAE provides the absolute error. 

Another secondary goal is to investigate which of the two conventional models (basic and expansion model) presented by \cite{liu_real-time_2009} and introduced in chapter \ref{sec.basic_model} and \ref{sec.expasion_model} has the better performance based on the second-by-second simulated loop detector data which are generated by SUMO (introduced in chapter \ref{sec.simulation_model}). The model with the better performance is then chosen for the future investigation together with the deep learning approach. 

Once the best conventional model is chosen, it is important to investigate the limitations of this model by applying it to realistic traffic scenarios. Furthermore, it is important to understand why these limitations occur. Only by analyzing the limitations of the conventional method, the deep learning model can be applied to the same traffic scenarios and the superiority of the geometric deep learning approach can be validated. 

During the work with artificial neural networks it is very important to find methods to restrict typical problems like under- and overfitting. Therefore, it is crucial to choose the right regularization methods with the right hyperparameters. Furthermore, it must be ensured that the deep learning model generalizes the knowledge properly during training. 
Another secondary goal is to investigate, how the trained deep learning model reacts, if the test data have different (for the deep learning model unknown) properties than the training data. The focus in this thesis lies on the arrival rate of vehicles (veh/sec) of the randomized traffic. Essentially, it is important to investigate how the deep learning model performs for an arrival rate that was not part of the training but is now used for testing. 

Finally, it is worth to investigate if the estimation result of the conventional method is beneficial as input for the deep learning model. To answer this question it is necessary to train the deep learning model one time with the estimation results of the conventional method and one time without the results. If the performance is better for the model which uses the conventional estimation results as input, it is shown that usage of prior knowledge is beneficial for the generalization of knowledge. Furthermore, it would give an additional justification for implementing the conventional state of the art method for this thesis. Besides the opportunity to compare both approaches, it would be also beneficial to combine both.

% describe main goal which is estimate the traffic more accurate than the conventional liu method by using information about spatial and temporal correllations presented in state of the art chapter

% -> which is the best metric to measure the performance of the models

% -> figure out which model works better for second by second data created by SUMO (basic or expansion model)

% -> investigate what the limitations of the liu-method are -> lane changing and stop-and go

% -> investigate how the deep learning model deals with characteristical problems in the field of Deep Learning like to less training data. How good does the dl model generalize the knowledge? How does the model react on data it has never seen before during training -> robustness for different arrival rates

% is it a benefit to combine both the conventional and the deep learning method

\chapter{Traffic Simulation Model} \label{sec.simulation_model}

In this chapter the traffic simulation software SUMO is introduced. Based on SUMO, the used road-networks for the traffic simulations are presented. 

\section{Traffic Simulation Software SUMO}\label{sec.SUMO}

\begin{figure}[H]
	\centering
	\includegraphics[width=1\textwidth]{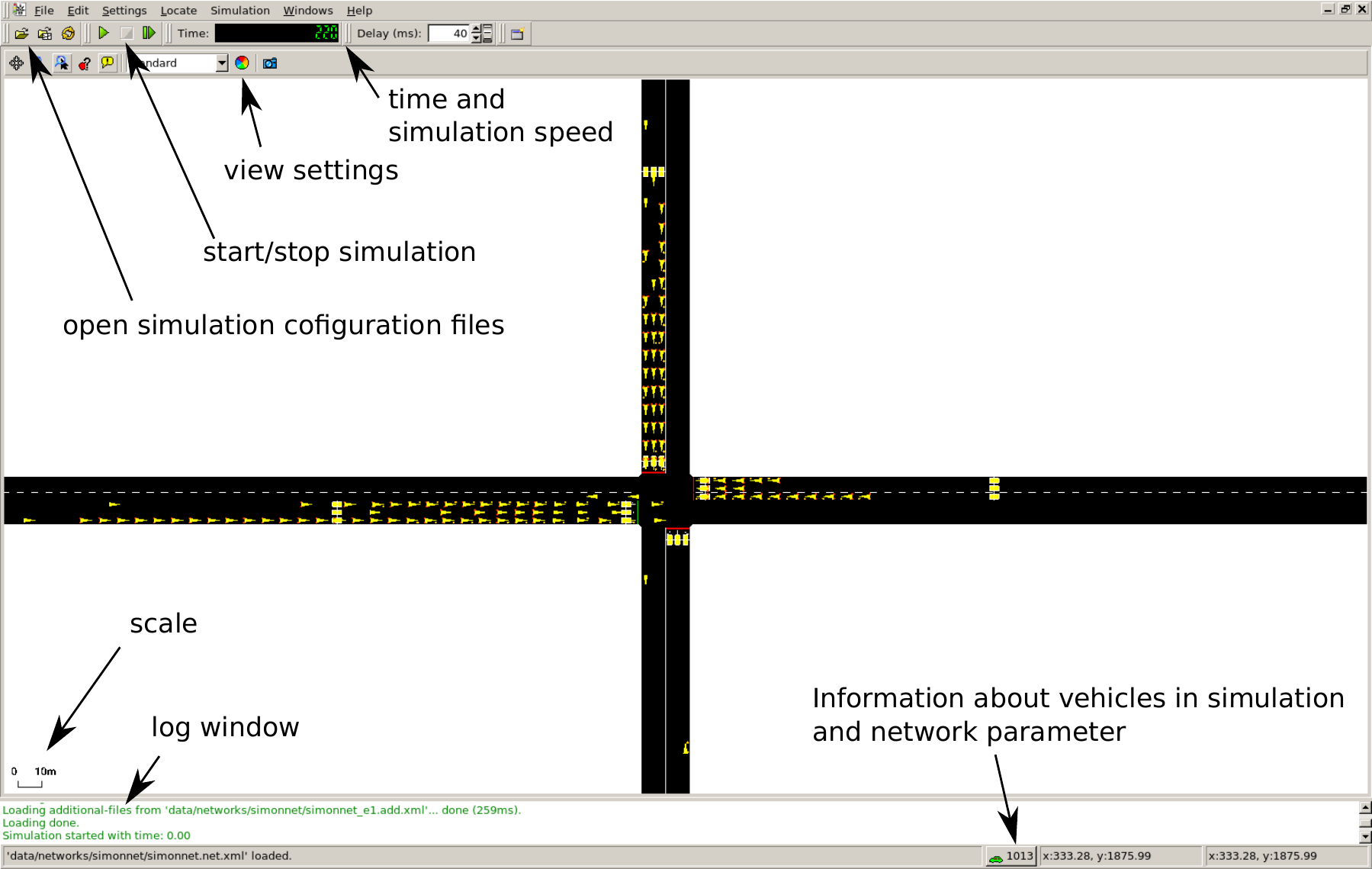}
	\caption[Presentation of the SUMO graphical user interface]{Presentation of the SUMO graphical user interface.}\label{bild.SUMO_GUI}
\end{figure}

For implementation and validation of the presented queue length estimation method, a reliable and variable data source is necessary. Therefore, in this work the open source traffic simulation software SUMO (Simulation of Urban MObility), developed by the German Aerospace Center (DLR) in 2001 \cite{krajzewicz_recent_2012-1}, is used. In SUMO it is possible to create large road networks, generate traffic, embed several sensor types and traffic light cycles.
Generally, SUMO is a microscopic simulator: Every vehicle is modeled explicitly with its individual route through the road network. The simulation can be either run deterministic, but also randomness can be applied. SUMO offers a variety of features that make a realistic simulation possible. Different traffic participants can be simulated, for instance cars, trucks, pedestrians and bicycles. Road networks with multiple lanes and lane changing can be implemented, as well as customized right-of-way rules and traffic light signals. 
Generally, SUMO is a command line application on Linux or Windows with no visualization, but it can be used with many other applications that come together in a package. One of them is the SUMO-GUI, which is a graphical user interface of the simulation software. The GUI is presented in Fig. \ref{bild.SUMO_GUI}. 

\begin{figure}[h]
	\centering
	\includegraphics[width=1\textwidth]{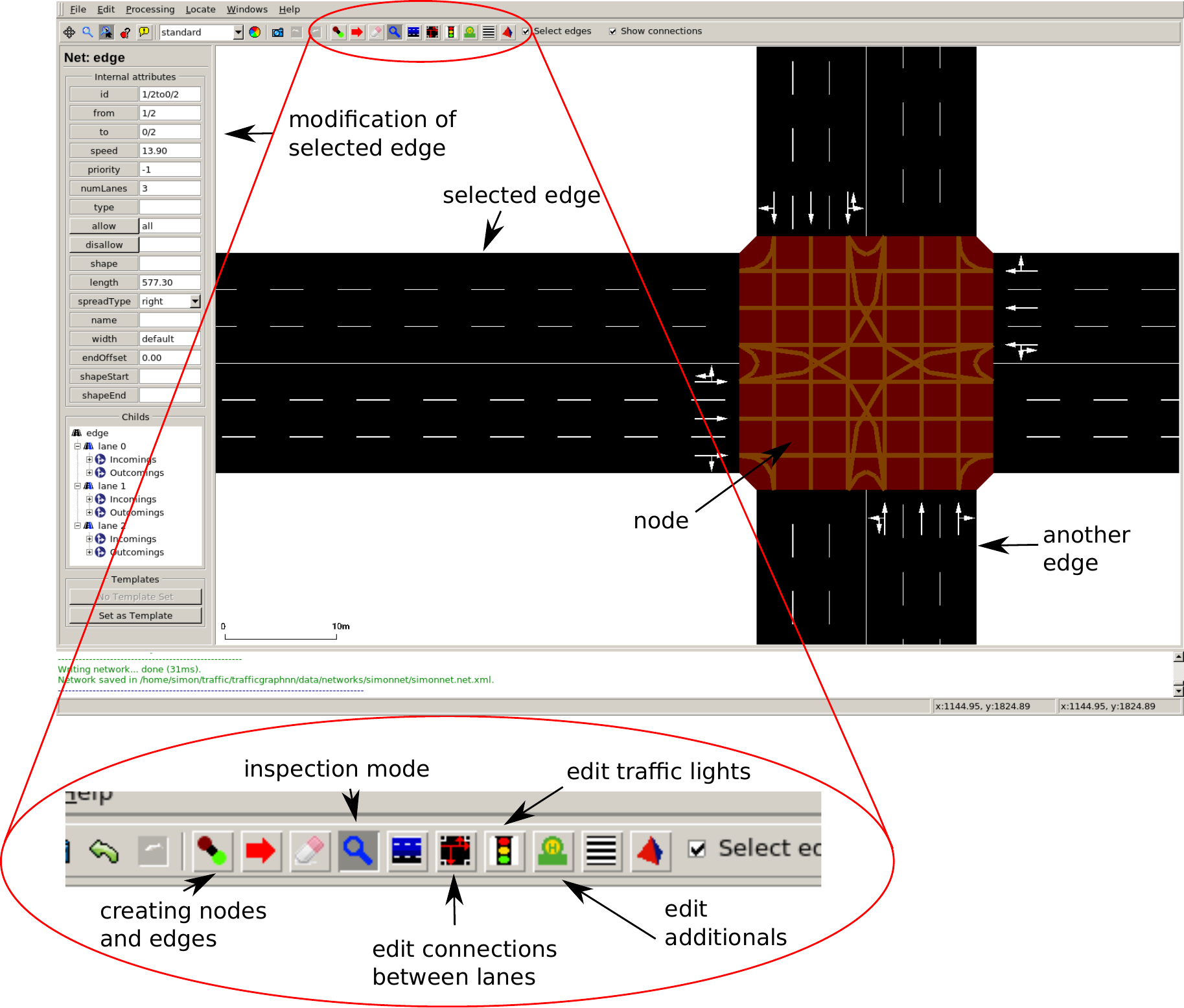}
	\caption[Presentation of the NETEDIT interface]{Presentation of the NETEDIT interface.}\label{bild.NETEDIT}
\end{figure}

For running the simulation, an existing road network is necessary. This can be created with the tool NETEDIT, which comes in the same package as SUMO. It is a visual network editor which can be used to design networks from the scratch but also to modify already existing road networks. The interface of NETEDIT is shown in Fig. \ref{bild.NETEDIT}.

In SUMO, road networks exists of nodes and edges. Nodes represent junctions in reality which are connected with each other by edges (streets in reality). In NETEDIT, nodes can be created at specific positions and graphically connected by edges. Ones the edges are created, the attributes of the edge (like length, name and number of lanes) can be modified. Furthermore, connections between lanes can be edited. It is possible to determine, which lanes are spatial connected between each other within a junction, so that vehicles can move through the junction. The traffic light phases can be modified, as well. That determines the temporal connection between two lanes at a junction. Whole traffic light cycles with several phases for each lane can be created. Finally, additionals can be added to the road networks. That are in this case e1 detectors (inductive loop detectors) and e2 detectors (ground truth detectors) which record the traffic behavior. The first test intersections in this work are created this way, but when it comes to lager road networks, it is much more efficient to use the command line application NETGENERATE. With this tool it is possible to configure and generate grid networks, spider networks, but also random networks by specifying the parameters in the command line. 

Once the network is created, it is necessary to determine start and destination for every single vehicle and create routes, respectively. Therefore, in this work the script randomTrips.py is called in the command line with arguments regarding the arrival rate of vehicles and the randomization. By that, randomTrips.py generates for each vehicle a random trip for the specified network. All the files (network, additionals, trips) are stored in XML files and can be used by the sumo application \cite{krajzewicz_recent_2012-1}. 

With the presented tools it is possible to create road networks with randomized traffic for testing and validation of the queue length estimation methods. These road networks are introduced in the following sections. 

\section{Single Test Intersection}

\begin{figure}[H]
	\centering
	\includegraphics[width=1\textwidth]{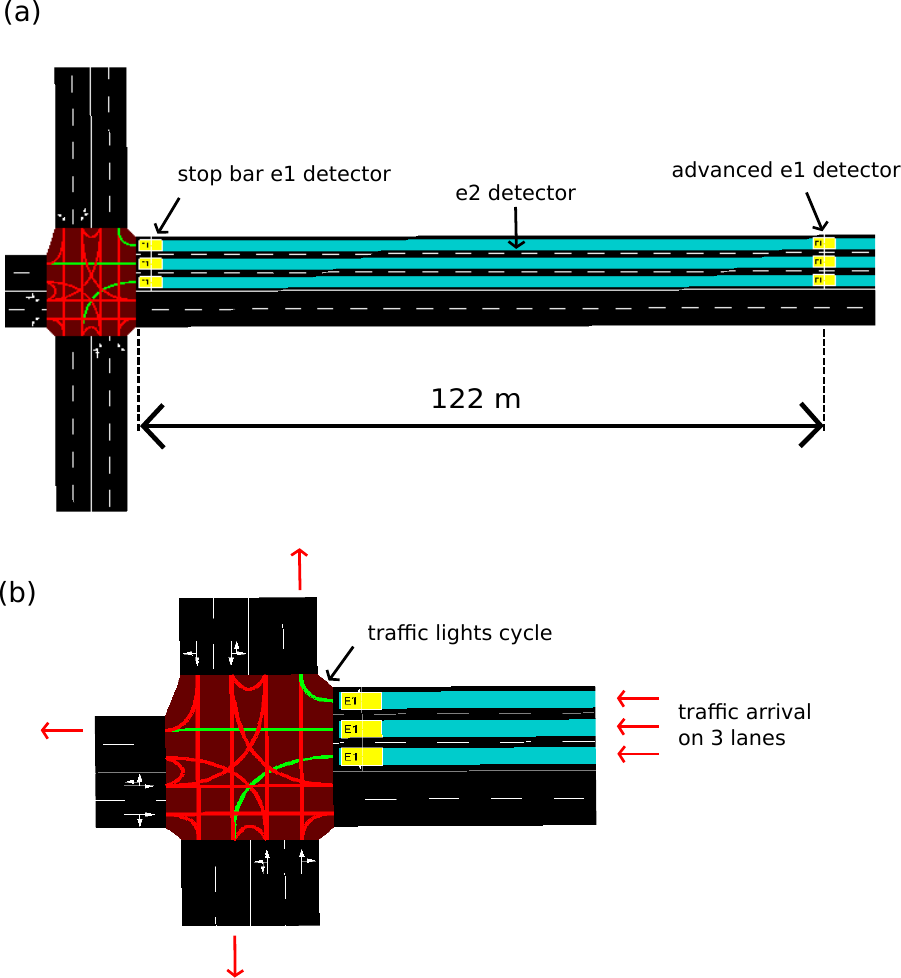}
	\caption[Test intersection in SUMO]{(a) Overview over intersection with position of stop bar and advanced e1 detector (both in yellow) as well as with e2 ground-truth detector (cyan); (b) Detailed view with marking of traffic directions and traffic light prioritizing (green and red connections between lanes). }\label{bild.one_intersection}
\end{figure}

In Fig. \ref{bild.one_intersection} (a) an overview of the set for testing and validation is presented. The queue length estimation is applied to three lanes, at which every lane has a different direction. The left lane is left turn only while the right lane is right turn only. All discharge lanes have green light, while every other lane has red traffic light (see Fig. \ref{bild.one_intersection} (b)). This is important, because otherwise the left turning vehicles would have to wait for the cars from other direction which would be prioritized. This could cause shockwaves in the queue that are not induced by traffic lights but by waiting for other vehicles. This disturbance shockwaves would be misinterpreted by the estimation algorithm and would lead to unreliable results. This issue will be discussed in section \ref{sec.liu_on_grid}.
Furthermore, in Fig. \ref{bild.one_intersection} (a) the stop bar e1 detector and advanced e1 detector is shown. In SUMO an e1 detector equates to single loop detector in the real world situation. These kind of detectors measure the detector occupancy over time and calculate \textit{speed} and \textit{number of passing vehicle} over a period of time based on detector reporting frequency. In SUMO the maximum reporting frequency of the detectors is 1 Hertz. Since the estimation method presented in \ref{bild.breakpoints} needs high resolution data to characterize the breakpoints, in this work the reporting frequency 1 Hz is chosen. For each lane stop bar detectors in front of the traffic lights are embedded. The advanced detectors are at a distance of 122 meters in front of the intersection. This was chosen because of the same distance in \cite{liu_real-time_2009}. The distance has to be chosen in consideration of the expected queue length and vehicle size. If the distance is too short, the shockwaves don't have enough the time and space to propagate trough the queue. If the distance is too big, no breakpoints can be detected. 

While the e1 detectors are data source for the queue estimation algorithm, e2 detectors provide ground-truth data. Ground-truth data reflect the real queue that occurs at the lane. The e2 detectors are extended over the whole lane and reporting with 1 Hz the \textit{number of started halts}. By summing-up the amount of halted vehicles over a whole cycle, the maximum queue length in vehicles can be calculated. This can be converted in meters by using the jam density factor $k_j$. In SUMO, e2 detectors also provide the value of \textit{maxJamLengthInMeters}, but this cannot be used as ground-truth data. This value has a different definition of queue length than \cite{liu_real-time_2009}, where maximum queue length is the distance from last halting vehicle to the stopbar during a cycle. In contrast to that, the e2 detector detects with value \textit{maxJamLengthInMeters} the length of halting vehicles within the one second reporting frequency. So for this definition the length can decrease when the front vehicles are discharging while at the rear cars stack on. This does not accord with the definition presented in the estimation algorithm. The results are slightly underestimating the real queue length. 
Not only e1 and e2 detector data can be exported by SUMO but also traffic light data with the \textit{start, end} and \textit{duration} of green light phases. Based on this data the period of time for each cycle can be determined. 

For creating vehicle routes for this network, the SUMO build-in file \textit{randomTrips.py} is used. This generator creates random routes for every lane in the network. For that matter the \textit{randomization}, \textit{start/end probabilities} and \textit{arrival rate} can be configured. Especially the arrival rate in combination with the traffic light cycle length should be adjusted, so that at the advanced e1 detector every breakpoint for every lane exists and is detectable. Because of the randomness this is not possible for every cycle. Randomization is important to give a proof of the accuracy of the algorithm, especially in situations where not all breakpoints can be detected.

\section{Grid-Road-Network} \label{sec.model_grid}

\begin{figure}[H]
	\centering
	\includegraphics[width=1\textwidth]{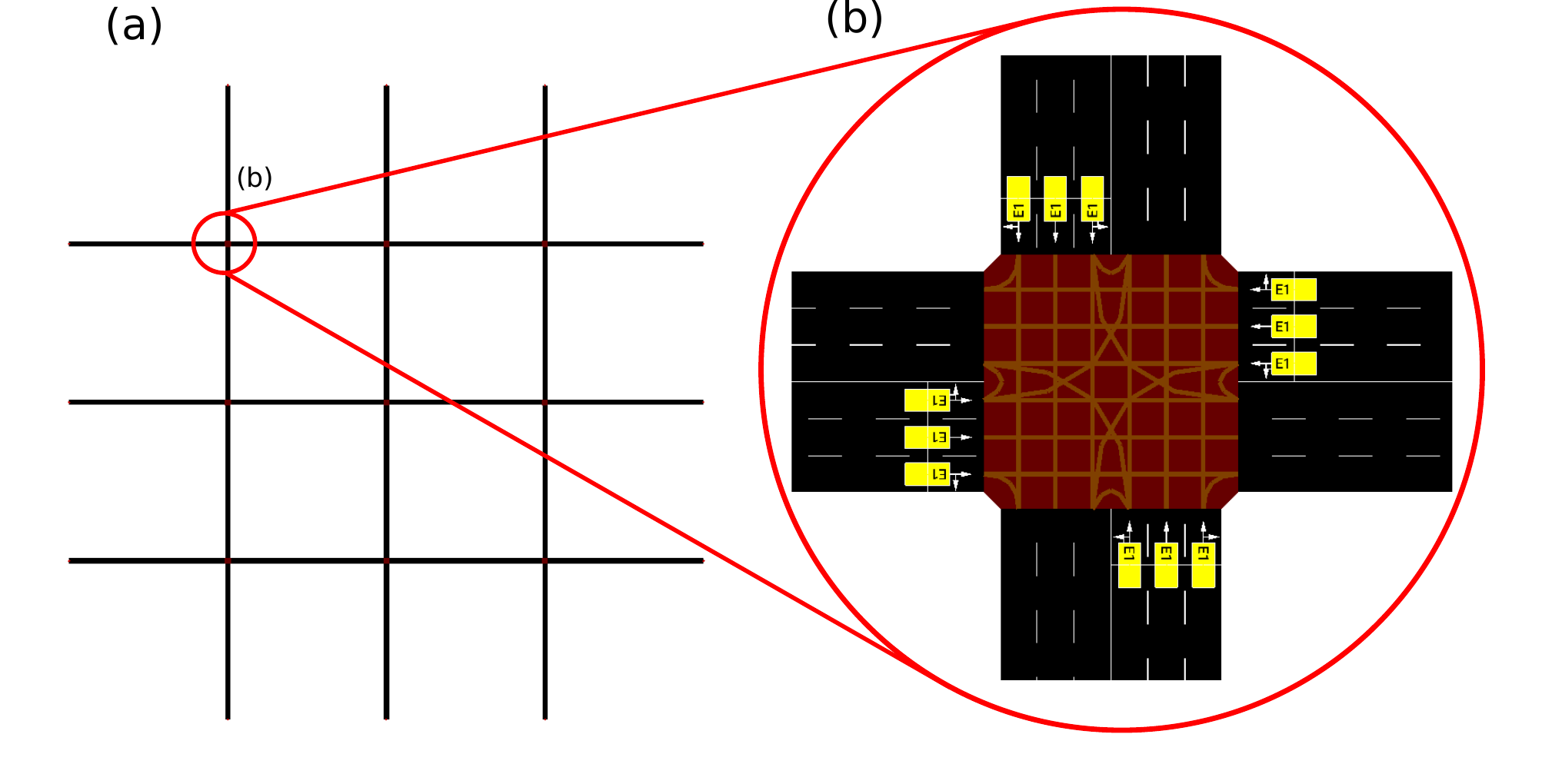}
	\caption[Topology of the test grid-road network]{(a) Overview of grid network with 3x3 intersections, a lane length of 600m and 3 lanes per direction; (b) Detailed view on intersection with connections for each lane and stop bar loop detectors.}\label{bild.grid_network}
\end{figure}
The single intersection model is useful to validate if the Liu-method is implemented and working correctly. To test queue length estimation methods under more realistic conditions, a grid-road-network is necessary. To generate these networks, a Python script in combination with NETEDIT is used. NETEDIT is an editor for road networks developed by the German Aerospace Center. It is open-source and comes together with SUMO. With the python script, number of intersections, street length and number of lanes can be adjusted.
In Fig. \ref{bild.grid_network} (a) an overview of the generated road network is shown. At each fringe intersection additional streets are attached. Only at the end of these lanes, vehicles can enter and leave the network. If this condition would not be met, vehicles were allowed to appear and disappear in the middle of the road network, which is not realistic and would lead to inadequate estimation results. In this work it is also not considered that vehicles park at the side of the road, which would have the same effect as appearing and disappearing. 

It is also necessary to make a simplification regarding the traffic lights. In this road network every direction has its own traffic light phase where only the lanes of a particular direction have green traffic lights. Every other lane has a red signal. This avoids shockwaves which are caused by waiting left turning vehicles. This simplification is used for testing the conventional method at the beginning of section \ref{sec.liu_on_grid}. For testing under even more realistic traffic conditions, the traffic lights are not simplified anymore, which is the case at the end of sec. \ref{sec.liu_on_grid} and in sec. \ref{sec.DL_implementation}. 
\chapter{Implementation of the Conventional Queue Length Estimation} \label{sec.Implementation_liu}

In this chapter the implementation of the conventional queue length estimation method (Liu-method) is presented. In the first section the Liu-method is implemented on a single test intersection to investigate the best conventional model working with the SUMO simulation data. Following, the Liu-method is implemented on a grid-road network and the limitations of this method are shown.  

\section{Implementation for a Single Intersection}

\begin{figure}[h]
	\centering
	\includegraphics[width=0.8\textwidth]{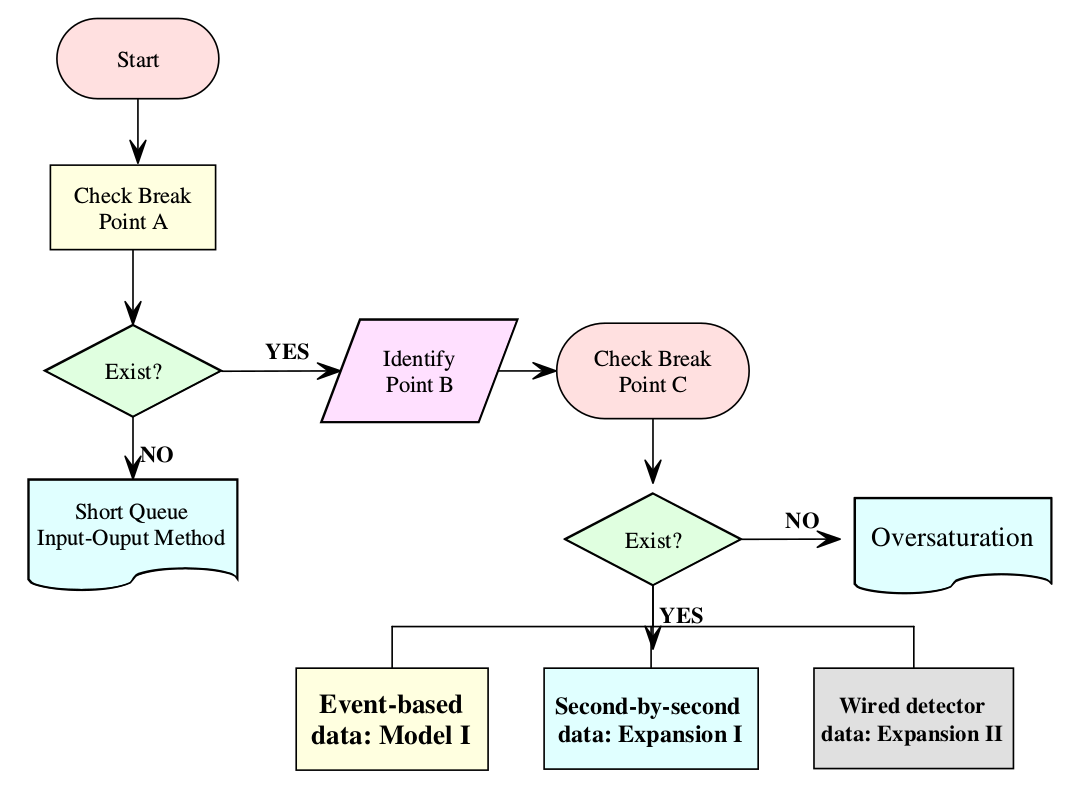}
	\caption[Flow chart for queue length estimation algorithm]{Flow chart for queue length estimation algorithm \cite{liu_real-time_2009}. }\label{bild.flow_chart}
\end{figure}

In realistic traffic behavior, sometimes not all three breakpoints can be detected in every cycle. Therefore, alternative strategies are necessary for these special cases. In Fig. \ref{bild.flow_chart} the main flow chart of the algorithm is presented. 
The first step is to check whether breakpoint A does exist or not. If A does not exist, since the queue length is too short and the advanced loop detector has no high occupancy, a simple \textit{input-output method} is used to estimate the queue length. Basically the estimation relies on counting the vehicle that leave or enter the section between stop bar and advanced detector during the cycle. The maximum queue length in vehicle for this cycle $n$ can be calculated as

\begin{equation}
N_{max}^n = max(N_{max}^{n-1}-N_{l}^{n-1}+N_{a}^{n,g}, 0) + N_{a}^{n, r}
\end{equation}  

where $N_{max}^{n-1}$ is the estimated maximum number of queued vehicles from the former cycle $n-1$, $N_{l}^{n-1}$ is the number of vehicles who are leaving during green phase of the $(n-1)$th cycle while $N_{a}^{n-1,g}$ is the number of vehicles who are passing the advanced detector from the start of green phase $T_r^{n-1}$ until end of the green light minus a time shift $T_g^{n-1*} = T_g^{n-1}-\frac{L_d}{v_0}$. This time shift covers the case that a vehicle passes the advanced detector while the traffic light is green, but has to stop because the light turns red after this moment. $N_{a}^{n, r}$ are the vehicles who pass the detector from $T_g^{n-1*}$ on until the end of red phase $T_r^n$. 
%used here in simulation seed = 40 and period = 0.7
\begin{figure}[h]
	\centering
	\includegraphics[width=1\textwidth]{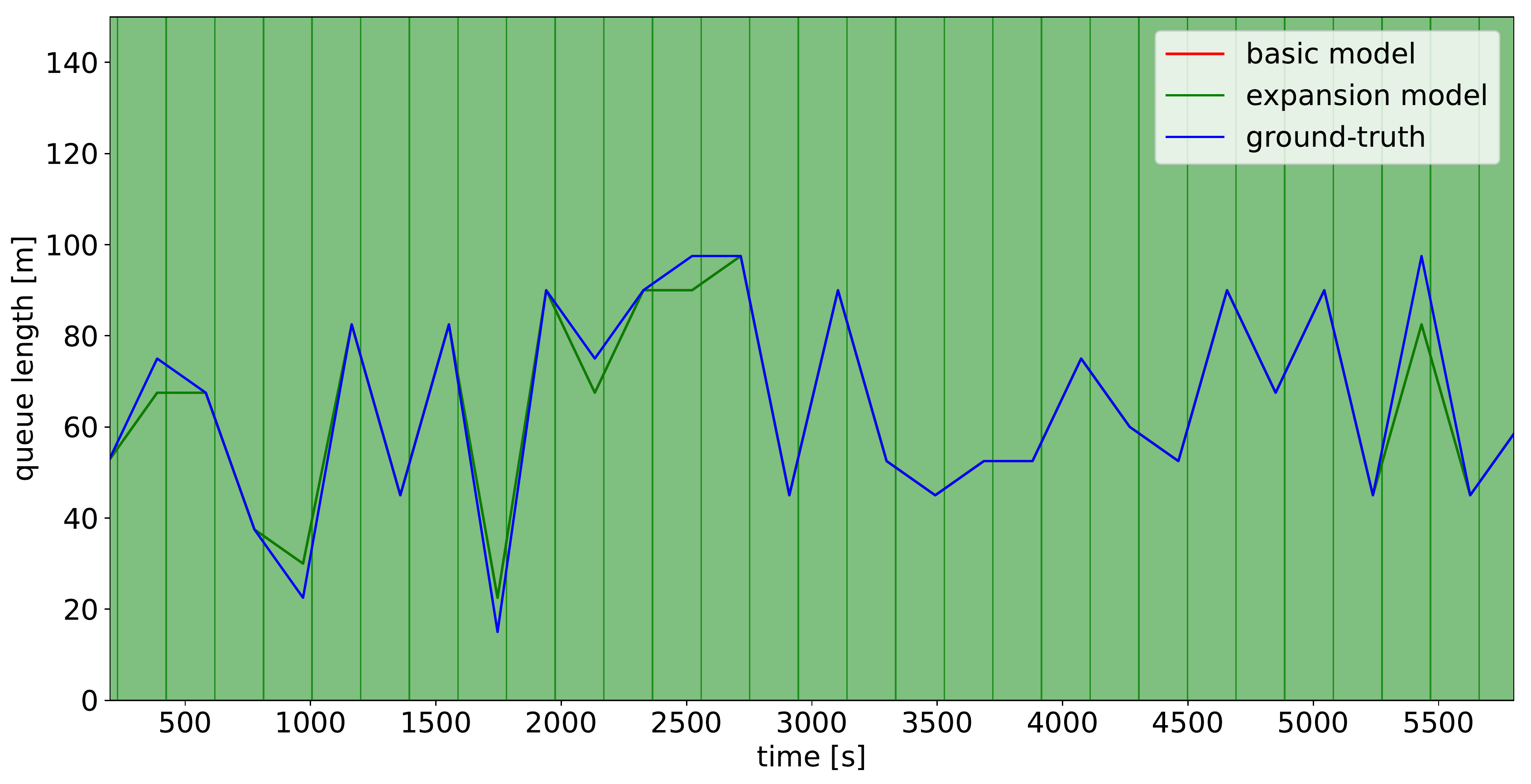}
	\caption[Results for simple input-output method]{Input-output method only for the left turn lane with a MAPE of 0.042. Green background indicates the usage of input-output method, yellow stands for basic and expansion model and red for oversaturation}\label{bild.plot_input_output}
\end{figure}
Basically, this calculation is divided in two parts: Vehicle input-output measurement during the green phase and a separated vehicle counting during red light. While the traffic light is on green, the queue length could become zero, so the maximum has to be taken. 
The results of an estimation where the input-output method is used only, are shown in Fig. \ref{bild.plot_input_output}. To get this results, the arrival rate parameter in the \textit{randomTrips.py} file was adjusted to 1.2 periods per generated vehicle.
The dark green vertical lines in the background of the figure signalize different cycles. Green background means, that in every cycle the input output method were used. In the following plots a yellow background stands for usage of the basic or expansion model while a red background is shown in case of oversaturation. 
In Fig. \ref{bild.plot_input_output} is presented, that the input output method works quiet reliable, only at some points an error of one vehicle in estimation occurs. This is caused by the unknown dynamic in the distance between stop bar and advanced detector. The \textit{MAPE (Mean Absolute Percentage Error)} is defined in \cite{liu_real-time_2009} as

\begin{equation}
MAPE = \frac{1}{m} \sum_{m} \bigm| \frac{Observation - Estimation}{Observation} \bigm| \times 100 \%.
\end{equation}

In this particular case the MAPE is 0.042 or 4.2\% which corresponds to the expectations. 

If breakpoint A exists, the algorithm has to find breakpoint B and check weather C exits. If breakpoint C does not exist, the case of oversaturation occurs. Thus an estimation is not possible, a valid solution is to assume the same queue length as in the time step before. 
If C exists, all breakpoints are detected and the basic and expansion model can be applied. In every such cycle the shockwave velocity $v_2$ and $v_3$ is estimated to make sure the estimation is robust against traffic arrival flow changes. To estimate the velocities from sensor data, the number of vehicles and individual speed during a specific period of time is recorded and eq. \ref{eq:space_mean_speed}, \ref{eq:flow} and \ref{eq:density} are applied to calculate the flow and density. To characterize the discharge flow $q_a$ and density $k_a$, the sensor data of the period from $T_r^n +5 seconds$ to $T_g^{n+1}$ are used. The five seconds are added to avoid dynamical influences and to record a static flow. To estimate the arrival flow $q_a$ and density $k_a$, sensor data of the period from $T_C^{n-1}$ until $T_A^n$ are processed. If $T_C^{n-1}$ does not exist, the red light start $T_g^n$ is used instead. Now eq. \ref{eq:v_2} and \ref{eq:v_3} can be applied. 
In this approach only the spatial and temporal information about the maximum queue length in each cycle is relevant. Therefore, it is not necessary to calculate minimum queue length and $v_1$. To apply the basic model only, eq. \ref{eq:L_max_basic} and \ref{eq:T_max_basic} are used and so for expansion model eq. \ref{eq:L_max_expansion} and \ref{eq:T_max_expansion}. 

\begin{figure}[H]
	\centering
	\includegraphics[width=1\textwidth]{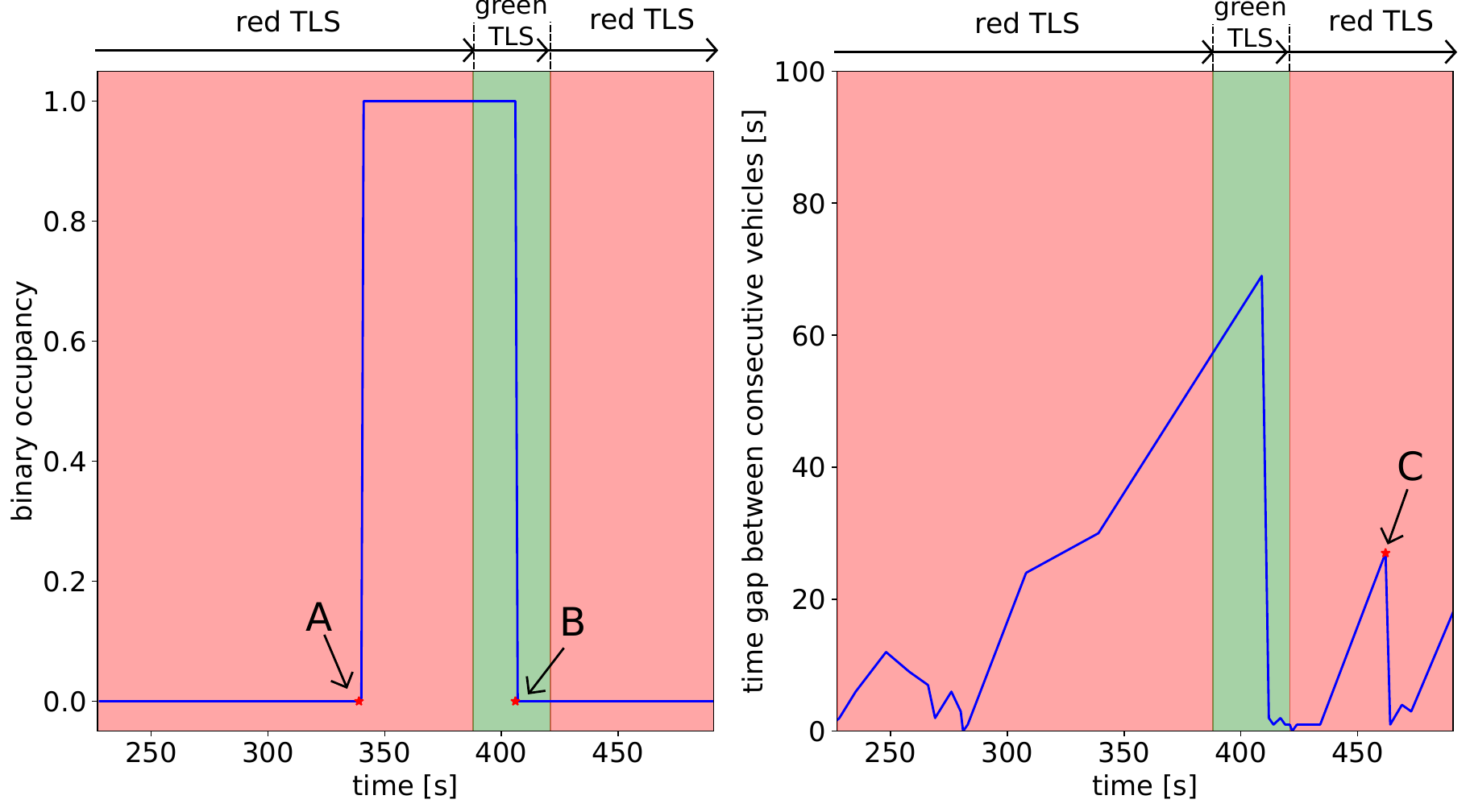}
	\caption[Characterization of breakpoints in simulation loop-detector data]{Characterization of breakpoints for one cycle based on simulation data in SUMO. The binary occupancy of the detector is shown on the left, while the time gap between consecutive vehicles is presented on the right. The red and green background colors identify the red and green traffic light signals (TLS).}\label{bild.breakpoints_simulation}
\end{figure}
\begin{figure}[H]
	\centering
	\includegraphics[width=1\textwidth]{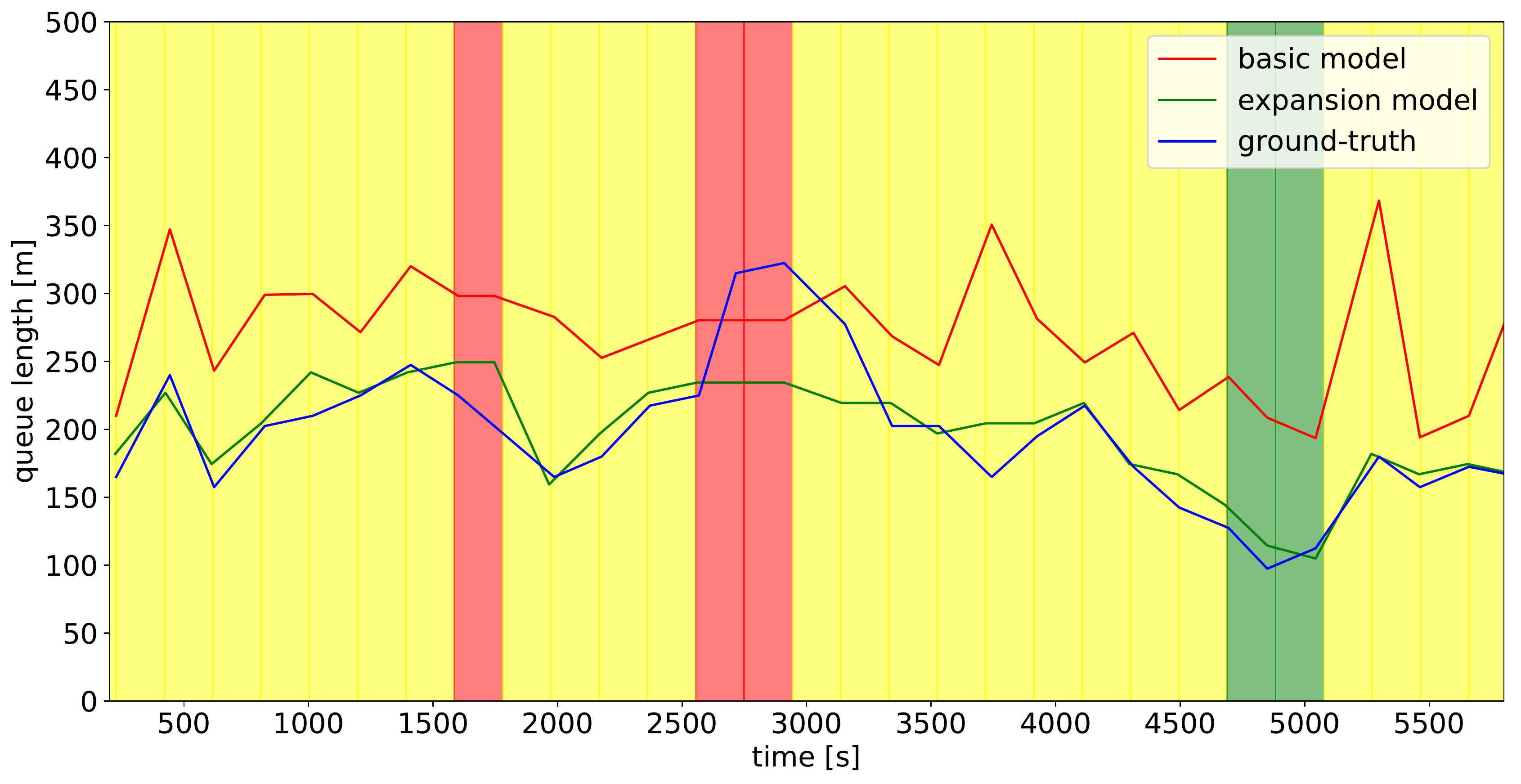}
	\caption[Estimation resuts for basic and expansion model with breakpoint C]{Estimation results for the basic and expansion model for the right turn lane. Green background indicates the usage of input-output method, yellow stands for basic and expansion model and red for oversaturation.}\label{bild.simulation_results_liu_1}
\end{figure}
In Fig. \ref{bild.breakpoints_simulation} is the characterization of breakpoints for a specific cycle shown. Similar to Fig. \ref{bild.breakpoints}, occupancy and time gap between consecutive vehicles are presented. But instead of using occupancy time as in \cite{liu_real-time_2009}, the binary occupancy is used, which value is one when occupancy is at $100\%$ for longer than one second and zero when not.
The breakpoint A is at the beginning of $100\%$ occupancy phase while breakpoint B is at the end. Breakpoint C is only detectable in the time gap data. Based on \cite{liu_real-time_2009} it is defined as the point where the next big time gap between the vehicles after breakpoint B occurs. Based on this points, the queue length estimation can be calculated and presented for the basic and expansion model in Fig. \ref{bild.simulation_results_liu_1}.
The expansion model (MAPE = 9.4\%) is much more accurate than the basic model (MAPE = 46.2\%), since the basic model overestimates queue length in every cycle, the expansion model has mostly the same shape than the ground-truth data. Only for the cycles where oversaturation occurs, the estimation becomes inaccurate, since constant queue length is assumed. The reason for overestimation of the basic model is because the characterization of breakpoint C is not reasonable for the randomized arrival traffic generated by SUMO. 

\begin{figure}[H]
	\centering
	\includegraphics[width=1\textwidth]{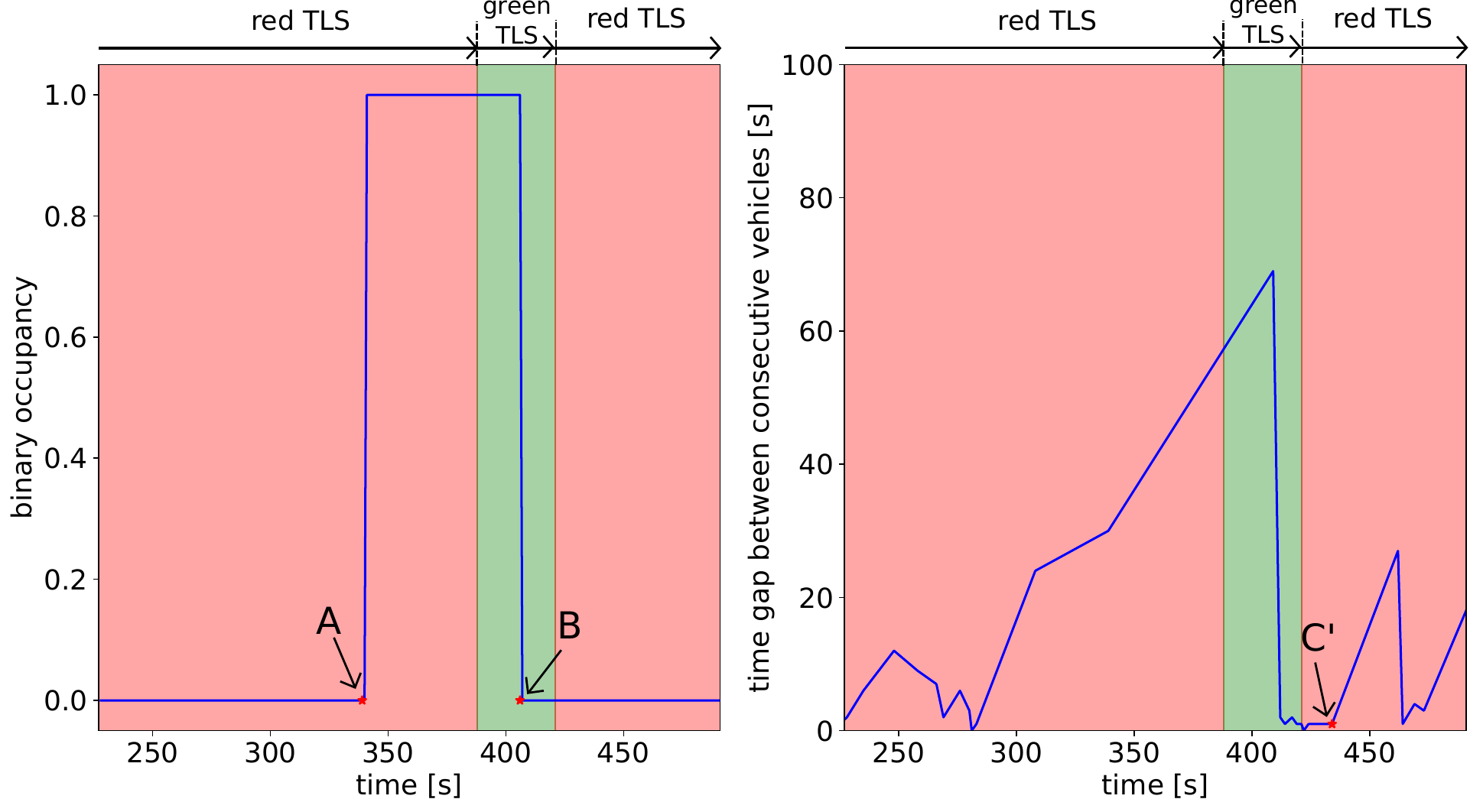}
	\caption[Alternative detection of breakpoint C]{Alternative characterization of breakpoint C'. The red and green background colors signalize the  red and green traffic light signals (TLS).}\label{bild.breakpoints_alternative}
\end{figure}

Breakpoint C is dependent on the next arriving vehicle after the end of the original queue passed the detector. Since this time gap is randomized by SUMO, it is not reliable information. Instead of choosing the maximum time gap between vehicles (original breakpoint C), the point of time when the last vehicle of the original queue passes the detector is chosen (new breakpoint C').
Thus the new breakpoint C' depends only on the last queued vehicle and not on the time gap of randomized traffic, the estimation results of the basic model becomes much more reliable. In Fig. \ref{bild.breakpoints_alternative} the alternative breakpoint C' is shown.
Thus the last vehicle of the queue passing detector is chosen as breakpoint C', the point of time becomes earlier and has no random influences anymore. An early C' effects a smaller length estimation and the overestimation is compensated.

\begin{figure}[H]
	\centering
	\includegraphics[width=1\textwidth]{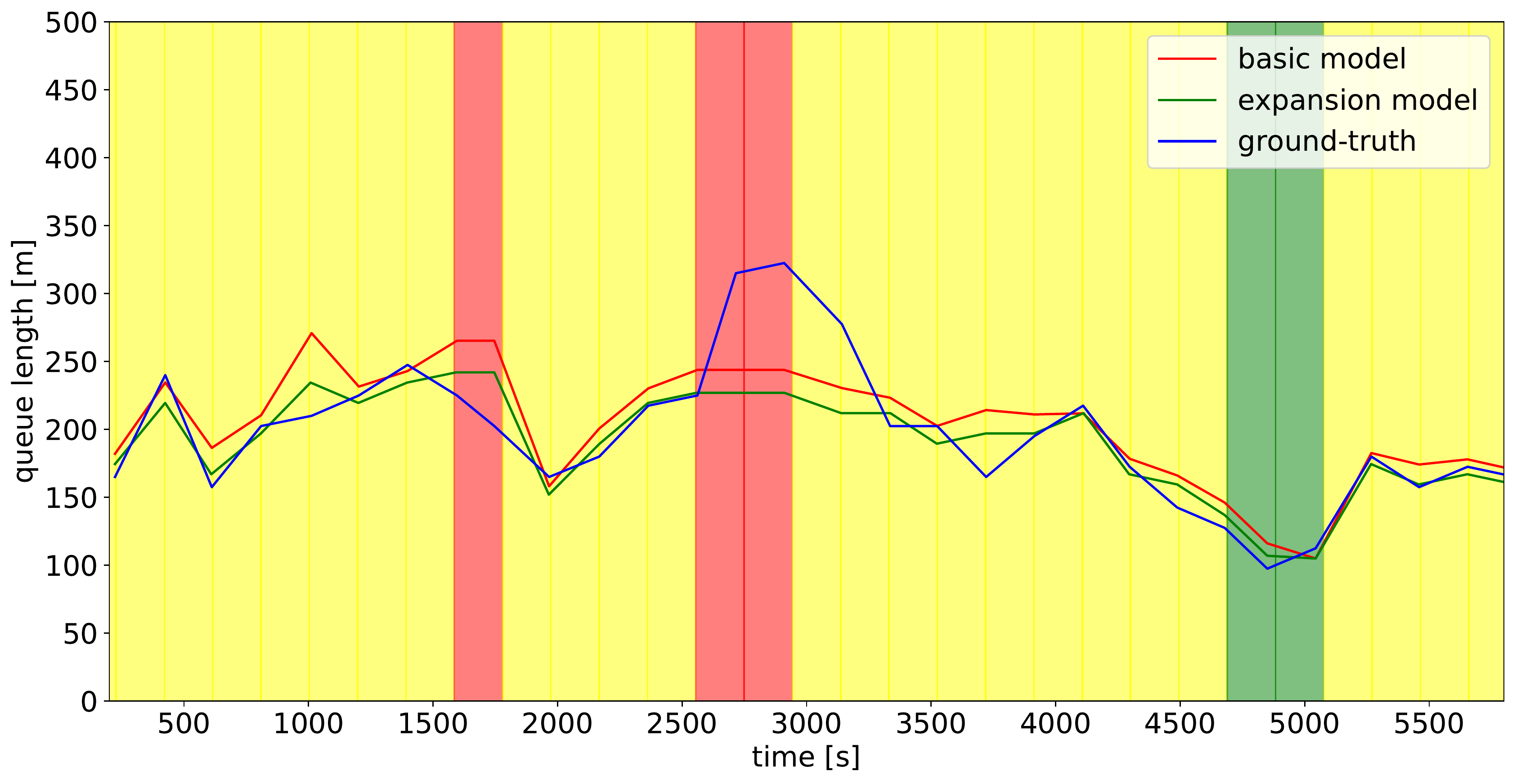}
	\caption[Estimation resuts for basic and expansion model with breakpoint C']{Estimation results for the basic and expansion model for the right turn lane based on the same simulation than for Fig. \ref{bild.simulation_results_liu_1}, but with usage of breakpoint C' instead of C. Green background indicates the usage of input-output method, yellow stands for basic and expansion model and red for oversaturation.}\label{bild.simulation_results_liu_1_alternative}
\end{figure}

The results of this modification are shown in Fig. \ref{bild.simulation_results_liu_1_alternative}. The accuracy of the basic model has improved. The basic model has for this particular example a MAPE of 11.4\% while the MAPE of the expansion model is at 8.3\%. The estimation results are now satisfying. Possible reasons for the remaining error are slightly wrong estimations of the shockwave velocities $v_2$ and $v_3$ due to not covered dynamics of the vehicles like acceleration processes. Another reason could be an imprecise characterization of the breakpoint C due to the limited resolution of one second. Breakpoint C has a high influence on the estimated maximum queue length.
However, the expansion model is much more robust compared to the basic model, since it relies on fewer assumptions.

\begin{figure}[H]
	\centering
	\includegraphics[width=1\textwidth]{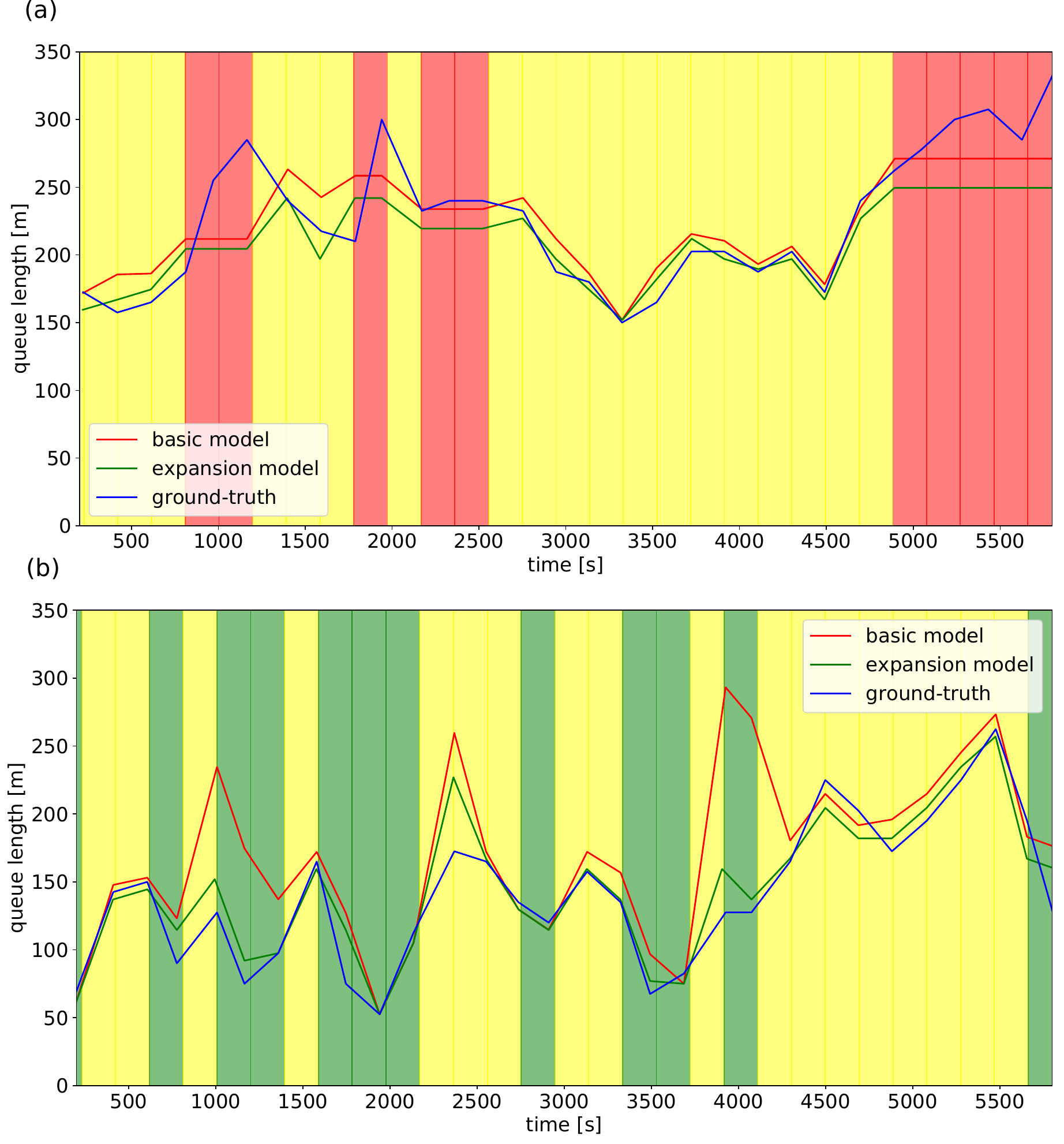}
	\caption[Estimation results for straight on and left turn lane]{Estimation results for the basic and expansion model for the straight on (a) and left turn (b) lane. Green background indicates the usage of input-output method, yellow stands for basic and expansion model and red for oversaturation; (a) MAPE basic model 8.7\%, MAPE expasion model 9.2\% ; (b) MAPE basic model 29.4\%, MAPE expansion model 11.2\% }\label{bild.simulation_results_liu_1_lane_1_2}
\end{figure}

 The expansion model uses only shockwave velocity $v_2$, which can be reliably estimated by eq. \ref{eq:v_2_alterative}, and the number of vehicles that are passing the detector between start of green traffic light and breakpoint C. So neither uncertainties in the estimation of $v_3$ nor $T_C$ influence the result. This significantly improves the robustness, especially since traffic arrival in SUMO is randomized. Therefore, as reference for the following neural network approach, only the expansion model is used. 
%\raggedbottom
In Fig. \ref{bild.simulation_results_liu_1_lane_1_2} the estimation results for the straight on and left turn lane are presented. In Fig. \ref{bild.simulation_results_liu_1_lane_1_2} (a) is shown, that in case of oversaturation the model cannot predict any queue length. In Fig. \ref{bild.simulation_results_liu_1_lane_1_2} the used method switches very often between input-output and basic/expansion model estimation. Especially around this critical queue length, breakpoint C is not precisely detectable and thus overestimation occurs. Immense estimation errors could lead to problems if a subsequent traffic control relies on this information. 

Generally, it is possible to estimate the queue length with the presented methods based on \cite{liu_real-time_2009} and the results are satisfactory . But it has to be taken into account, that on the resolution loop detector data must be available and this method works only for a specific range of the queue length. As a basic requirement the rear end of the queue has to pass the detector during the cycle, yielding the detectable breakpoints. If the queue is too short, a simple input-output method is applied. If the queue length exceeds the detector during the whole cycle, an estimation is not possible anymore. Hence, this method works only for a specific set of traffic arrival parameter, where arriving and leaving cars are in equilibrium. 

\section{Implementation for a Grid-Road-Network} \label{sec.liu_on_grid}

\begin{figure}[h]
	\centering
	\includegraphics[width=1\textwidth]{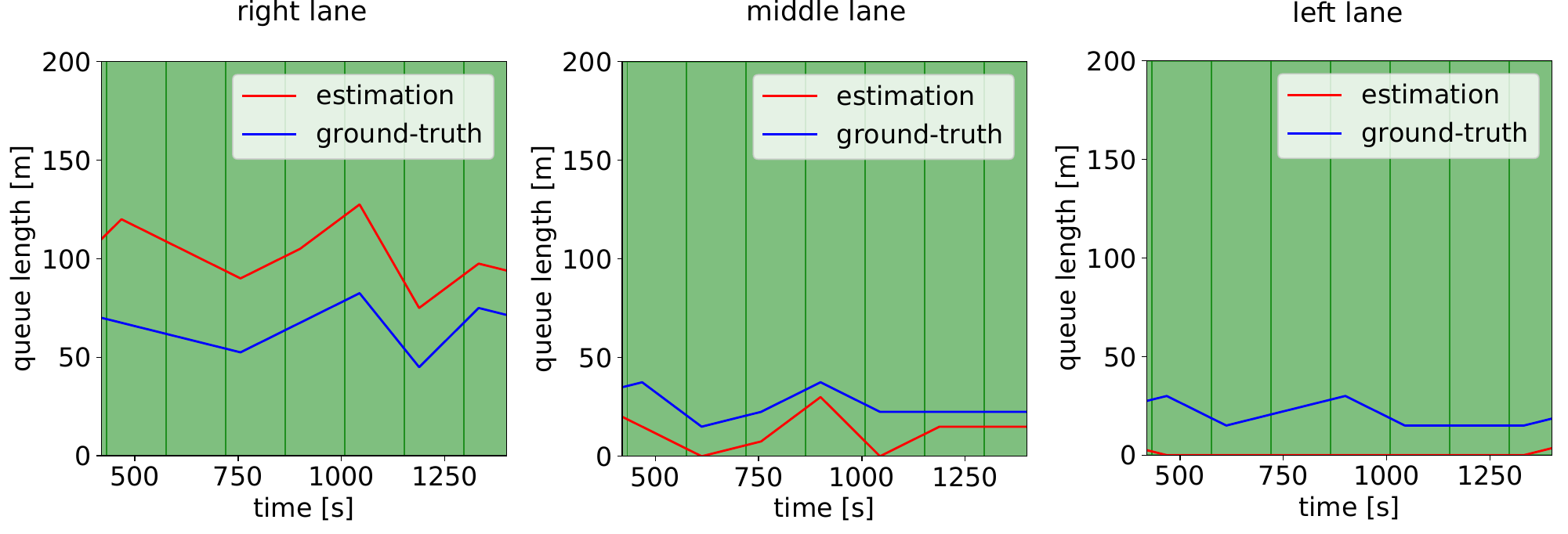}
	\caption[Essect of lane changing]{Estimation results with input-output method for right, middle and left lane. Green background indicates the usage of input-output method, yellow stands for basic and expansion model and red for oversaturation.}\label{bild.grid_network_wrong_estimation}
\end{figure}

Based on the successful implementation of the expansion model for a single intersection, the aim is to implement an adequate queue length estimation for every lane in every traffic light cycle for a grid network with randomized traffic arrival. 
%Abschnitt entfernt: Jetzt in Kapitel Traffic Simulation Model
A issue which occurs with the structure of a grid-road-network is lane changing of cars downstream of the advanced loop detectors. In Fig \ref{bild.grid_network_wrong_estimation} the estimation results for every lane in a specific direction is shown.

\begin{figure}[h]
	\centering
	\includegraphics[width=1\textwidth]{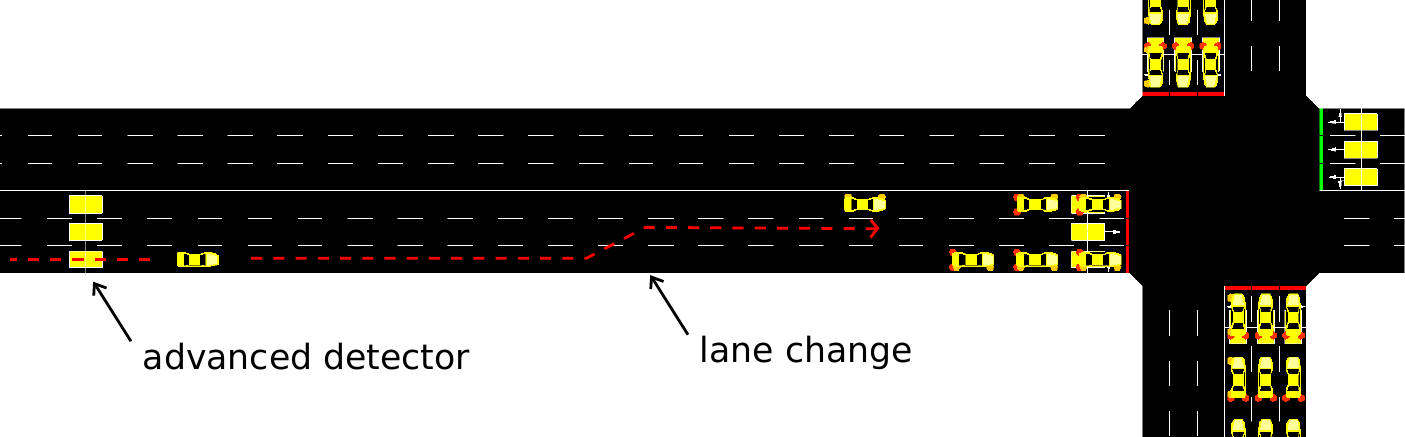}
	\caption[Visualization of lane change behind advance detector]{Visualization of lane change in SUMO after the advanced detector }\label{bild.lane_change_simulation}
\end{figure}

Note that the queue length for the right lane is overestimated while the middle and left lane is underestimated. This occurs because of the late lane change of cars behind the advanced loop detector. The input-output method cannot handle this case. Once a vehicle passed the advanced detector on a specific lane, the estimation algorithm assumes that the vehicle will stack on the queue at the same lane. In the particular example in Fig \ref{bild.grid_network_wrong_estimation} most cars arrive on the right lane, pass the advanced detector and change the lane then to the middle or left lane. Due to this driving behavior an overestimation for the right lane occurs. Because less vehicles were detected by the advanced detectors on the middle and left lane, underestimation occurs. This correlation is visualized in Fig \ref{bild.lane_change_simulation}.

The lane change behavior can be adjusted by parameters in SUMO like \textit{lcStrategic}, which describes the eagerness for performing strategic driving behavior and lane changing. Another parameter is \textit{lcKeepRight}, which is the eagerness to keep right on the street. But in this work the parameters were adjusted in a way that lane changes after the advanced detector are still possible. Due to this, the simulation is more realistic.

An alternative way to improve the estimation results is to apply the expansion model from chapter \ref{sec.expasion_model} not only to the advanced detector, but also to the stop bar detector. In other words, the input-output method is replaced by the expansion model method. In general the accuracy of the expansion model is not as adequate as the input-output method because it relies on more assumptions. But in case of lane changing the expansion model is superior because it counts only vehicle that pass the stop bar detector. So the results are more independent from lane changes. 

\begin{figure}[h]
	\centering
	\includegraphics[width=1\textwidth]{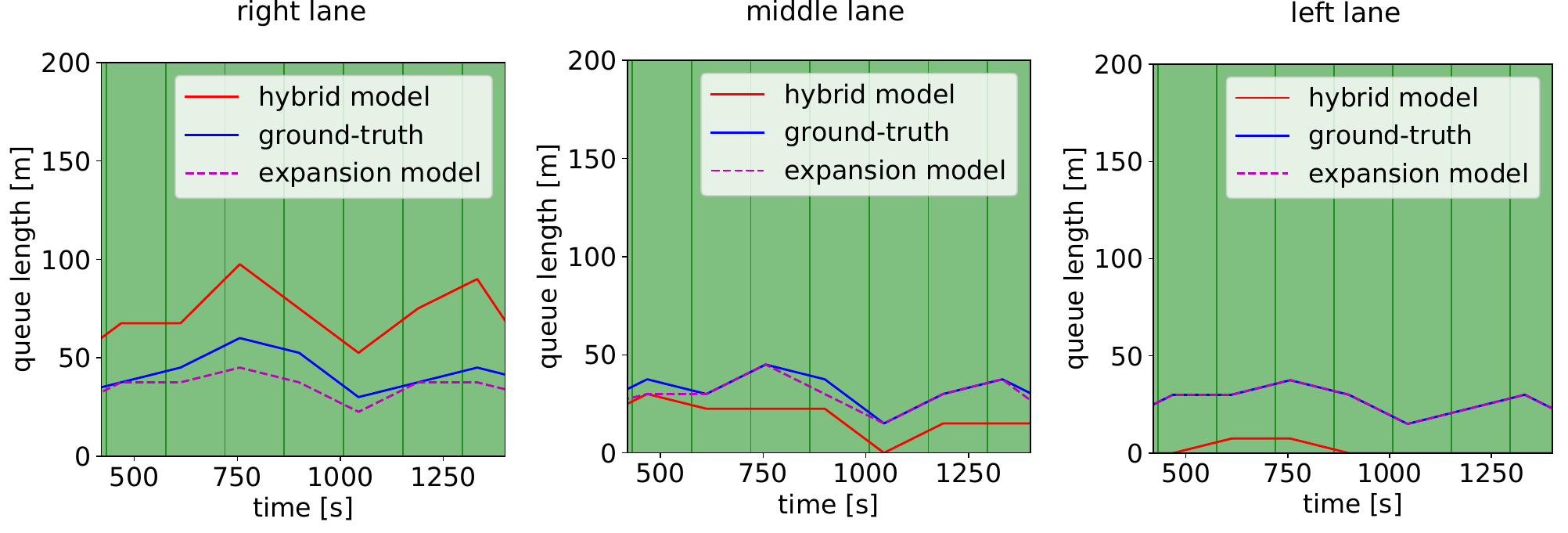}
	\caption[Comparison between results of hybrid and pure expansion model]{Estimation results of expansion model for short queues in comparison with input-output method for right, middle and left lane. Green background indicates the usage of input-output method, yellow stands for basic and expansion model and red for oversaturation.}\label{bild.grid_network_pure_liu}
\end{figure}

\begin{figure}[h]
	\centering
	\includegraphics[width=1\textwidth]{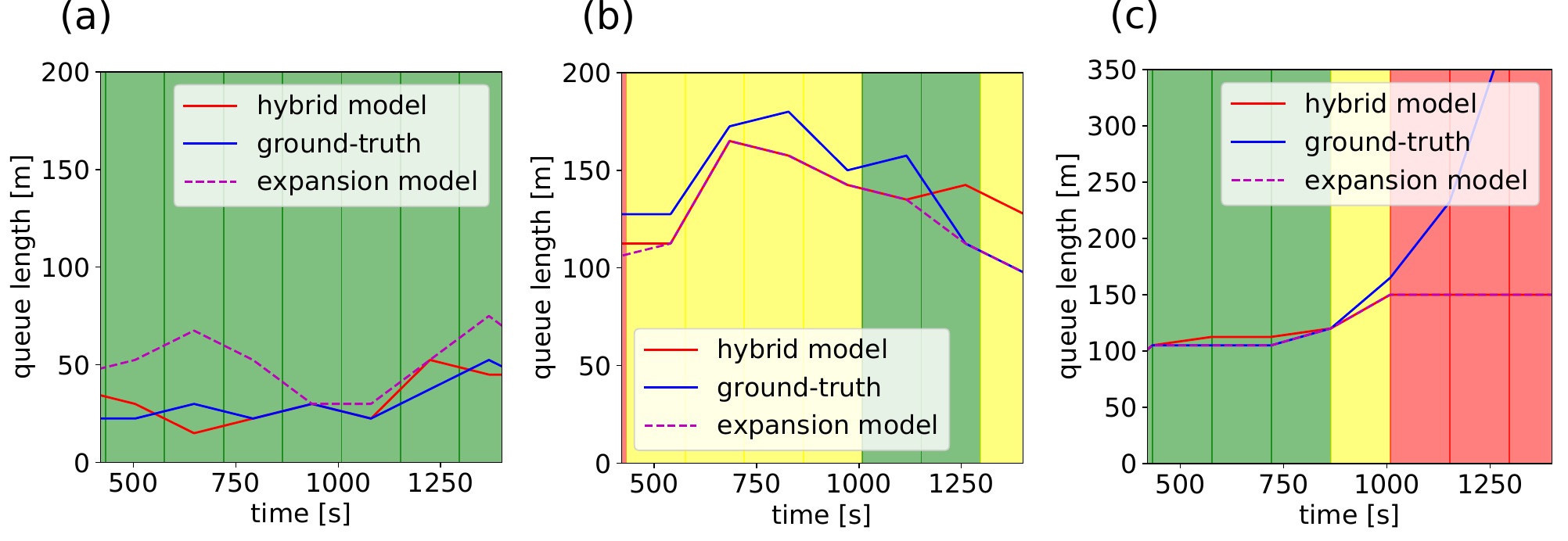}
	\caption[Estimation results for grid-road network]{Estimation results of expansion model: (a) Overestimation by expansion model; (b) Estimation for long queues are for both methods equal, because the rely on the basic model; (c) In case of oversaturation an estimation becomes impossible. Green background indicates the usage of input-output method, yellow stands for basic and expansion model and red for oversaturation.}\label{bild.grid_network_results_example}
\end{figure}

In Fig. \ref{bild.grid_network_pure_liu} the same lanes as in Fig \ref{bild.grid_network_wrong_estimation} are shown. The \textit{hybrid model} means the usage of the input-output method for short queues and the expansion model for longer queues. The \textit{expansion model} in Fig. \ref{bild.grid_network_pure_liu} stands for usage of the expansion model both for short and long queues. It is obvious that the accuracy of the expansion model for short queue prediction is better than the hybrid model. However, the expansion model has some issues too, as discussed in chapter \ref{sec.expasion_model}. The expansion model is in some situations overestimating, due to the assumption that every vehicle passes the detector before breakpoint C was part of the queue. This is not always the case. A characteristic overestimation is presented in Fig. \ref{bild.grid_network_results_example} (a).

In this particular case the stop bar detector counts more vehicles than vehicles were part of the former queue. Unfortunately, it is not possible to find an adequate breakpoint C for this case. Fig. \ref{bild.grid_network_results_example} (b) shows an estimation of a longer queue. As expected the estimations for both hybrid and expansion model are equal for long queues (yellow background), because they use the same equations. Only for the short queue estimations differences occur. At some lanes of the network oversaturation occurs (see Fig. \ref{bild.grid_network_results_example} (c)). For this case an estimation is not possible and a constant queue length is assumed. 
It is expectable that some lanes of the network have a higher demand than other lanes due to their location. Basically both methods (hybrid model and expansion model) generate reliable estimations for every lane as long no oversaturation occurs. In Tab. \ref{tab:MAPE_grid_net} the MAPE for the hybrid and expansion model for different arrival rates are presented.

% \usepackage{booktabs}
\begin{table}[]
	\centering
	\begin{tabular}{@{}cc|c|cc@{}}
		& \multicolumn{2}{c|}{\textbf{hybrid model}} & \multicolumn{2}{c}{\textbf{expansion model}}     \\ \midrule
		\multicolumn{1}{c|}{\textbf{veh/sec}} & MAPE {[}\%{]}         & SD {[}\%{]}        & \multicolumn{1}{c|}{MAPE {[}\%{]}} & SD {[}\%{]} \\ \midrule
		\multicolumn{1}{c|}{0.3}              & 39.7                  & 22.4               & \multicolumn{1}{c|}{22.9}          & 18.1        \\
		\multicolumn{1}{c|}{0.4}              & 36.1                  & 22.6               & \multicolumn{1}{c|}{18.4}          & 13.6        \\
		\multicolumn{1}{c|}{0.5}              & 34.1                  & 25.3               & \multicolumn{1}{c|}{14.4}          & 12.4        \\
		\multicolumn{1}{c|}{0.6}              & 33.5                  & 21.5               & \multicolumn{1}{c|}{15.55}         & 16.7       
	\end{tabular}
	\caption[MAPE and standard deviation for certain arrival rates]{MAPE and Standard Deviation (SD) for certain arrival rates. The number of estimations (number of lanes) is 120.}\label{tab:MAPE_grid_net}
\end{table}

\begin{figure}[h]
	\centering
	\includegraphics[width=1\textwidth]{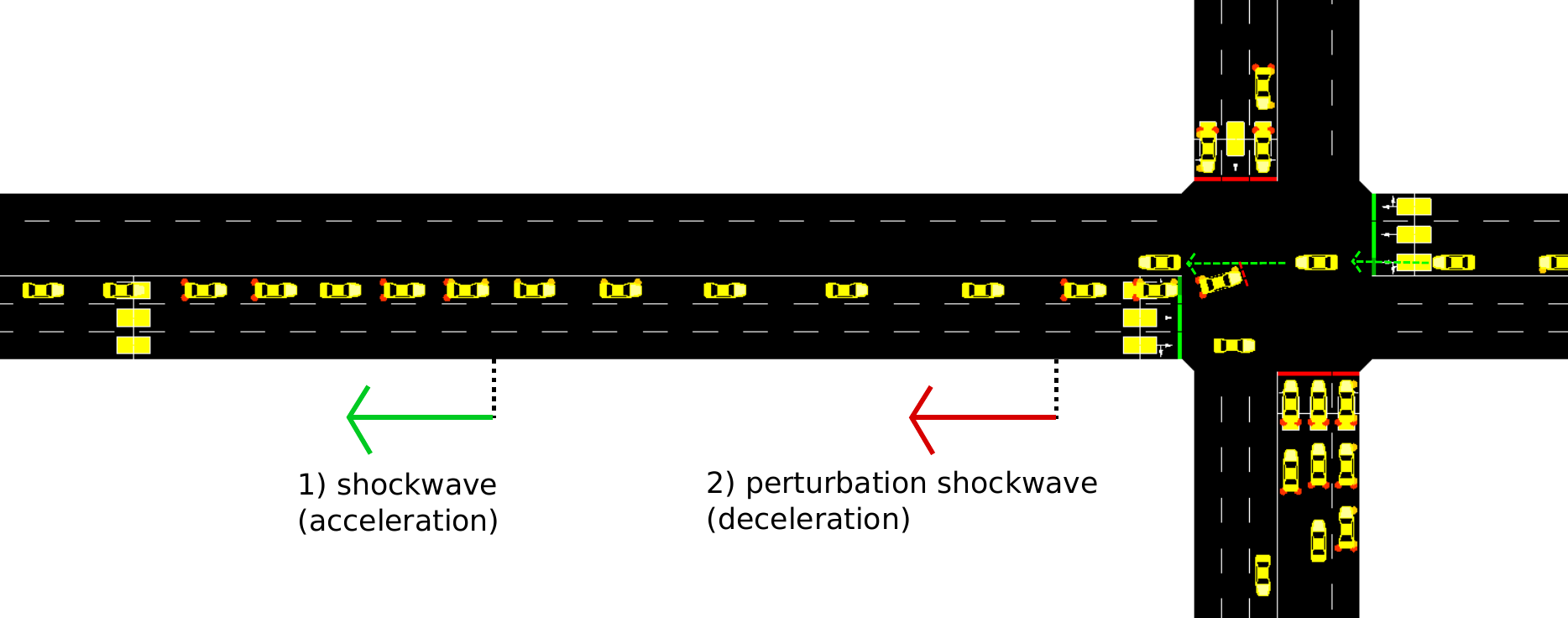}
	\caption[Visualization of perturbation shockwaves due to stop-and-go traffic]{Perturbation shockwaves caused by waiting vehicles on a left turn lane}\label{bild.perturbation_shockwaves}
\end{figure}

In every case the expansion model is better in accuracy than the hybrid model. This is due to the lane changes not covered by the hybrid model. This explains also the higher standard deviation for the hybrid model. For lanes where a lot of lane changes occur, the estimation becomes inaccurate and the standard deviation increases. For a very high arrival rate (0.3 veh/sec) the MAPE increases for both the hybrid and expansion model, since more oversaturation occurs.  Considering the issues with overestimation, lane changing and randomized traffic arrivals, the expansion model is satisfying regarding the accuracy. The hybrid model becomes inaccurate due to lane changing, but considering the issues it is still acceptable.
However, so far only intersections with prioritized traffic light signals were taken in consideration. Due to simplification of the traffic lights, every lane could discharge without any perturbation. Intersections in reality have more complicated traffic light circuits with parallel discharging processes to increase the efficiency. As a result, vehicles which want to make a left turn, have to wait for the vehicles that go straight from the other direction. This situation is visualized in Fig. \ref{bild.perturbation_shockwaves}. 

\begin{figure}[h]
	\centering
	\includegraphics[width=1\textwidth]{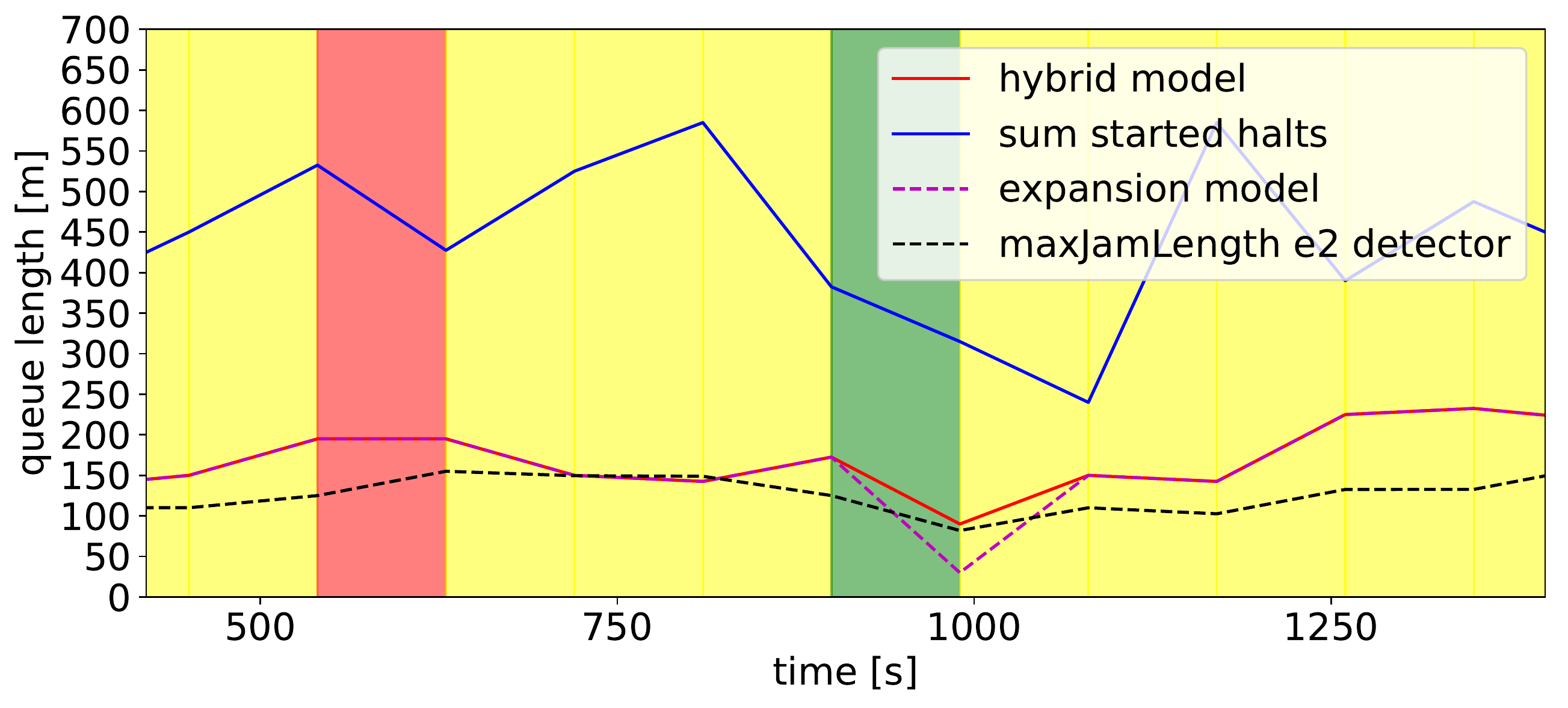}
	\caption[Overcounting of started halts as ground-truth data]{Comparison between overcounting the ground truth data by sum-up the stated halts and usage of the \textit{maxJamLength} of the e2 detector. Green background indicates the usage of input-output method, yellow stands for basic and expansion model and red for oversaturation.}\label{bild.started_halts_overcounting}
\end{figure}

The first vehicle on the left turn lane can move on the intersection as soon as the traffic light becomes green. That causes the fist shockwave which propagates upstream and vehicles starts to accelerate. However, the trough traffic from the other direction has right of way. Left turning vehicles have to stop again and a second shockwave propagates upstream where vehicles decelerate and halt. Due to this uncertainty in the discharging process, a successful application of the expansion model, which is based on the shockwave theory, becomes not reliable anymore. Furthermore, the counting of started halts over a traffic light cycle as ground-truth data is not a feasible method anymore, since vehicles stop and go again multiple times during a cycle. An alternative is to take the maximum of \textit{maxJamLengthInMeters} over an entire cycle provided by the e2 detectors in SUMO. As already discussed in section \ref{sec.SUMO} this method is slightly undercounting, because it only counts the vehicle that are stacked together. If vehicles are discharging at the beginning of the green phase, the calculated queue length is decreasing although vehicles stack on at the rear end of the queue. This undercounting has also an influence on the MAPE because the groung-truth data change slightly. As an demonstration of this effect the MAPE is calculated for a grid network with three lanes in each directions with simplified traffic light cycles. That means that the sum of started halts are reliable ground-truth data because the lanes are discharging without any disturbance, while the maximum of maxJamLengthInMeters is undercounting. The MAPE for using \textit{maxJamLengthInMeters} is for the hybrid model 45.9\% and for the pure basic model 30.4\%. In comparison to that the MAPE for the ground-truth data based on the started halts are for the hybrid model 37.7\% and for the pure basic model 23.1\%. Essentially it is better to use the started halts method when there are no disturbance influences. However, the started halts method becomes extreme unreliable when stop and go behavior of the vehicles occur. This is visualized in Fig. \ref{bild.started_halts_overcounting}. 

The calculated ground-truth queue length exceeds 500 meters which is not realistic for that traffic case and only caused by the disturbance shockwaves induced by waiting vehicles.  
However, the undercounting error of the maxJamLengthInMeters is less of an issue as the overcounting error of the other method in case of realistic traffic lights. Due to this, in the further work the maxJamLengthInMeters as ground-truth data for training the neural networks and reference is used. The resulting MAPE for an estimation with the maxJamLengthInMeters as ground-truth data, realistic traffic lights and the same network than described above is for the hybrid model 0.595 and for the pure basic model 0.730.
Due to disturbances in form of lane changing, stop and go traffic but also the undercounting error of the used ground-truth data, the results of the estimation become not accurate anymore. The basic model presented in section \ref{sec.expasion_model} cannot handle these disturbances. This is the motivation to investigate another approach for realistic road networks by using Geometric Deep Learning which might be superior due to learning spatial correlations over a whole road network. 

%
%Due to this disturbances the accuracy of estimation decreases. In Tab. [...] the results are for both estimation methods presented. Additionally, the results for left turn and any other lanes are shown separately to emphasize the influence of disturbance on left turn lanes.
%
%\begin{table}[h!]
%	\begin{center}
%		\caption{MAPE for certain arrival rates with complex traffic light circuits}
%		\label{tab:MAPE_real_net}
%		\begin{tabular}{c|c|c|c|c|c|c} % <-- Alignments: 1st column left, 2nd middle and 3rd right, with vertical lines in between
%			\multicolumn{1}{c    }{} & \multicolumn{3}{ |c|  }{\textbf{hybrid model}} & \multicolumn{3}{c}{\textbf{expansion model}}\\
%			\hline
%			sec/veh & left & other & total& left &other & total\\
%			\hline
%			0.3 & 0.258 & 0.222 & 0.258 & 0.222 & 0.258 & 0.222\\
%			0.4 & 0.193 & 0.173 & 0.258 & 0.222 & 0.258 & 0.222\\
%			0.5 & 0.184 & 0.149 & 0.258 & 0.222 & 0.258 & 0.222\\
%			0.6 & 0.176 & 0.167 & 0.258 & 0.222 & 0.258 & 0.222
%		\end{tabular}
%	\end{center}
%\end{table}
%
%
% They are used on one hand for reference with the neural network approach. On the other hand they could be data source as ground-truth data for training a neural network under a realistic scenario.
%
%
%
%
%mention, that in average the MAPE of left turn lanes is worse than straight on and right turn. Show comparison from left lane to other lanes. 
%
%Formulate motivation, that because of lane changes, left turn waiting the estimation results become very bad -> for realistic road networks is another approach necessary -> geometric deep learning!
%
%

\chapter{Queue Length Estimation with Geometric Deep Learning} \label{sec.geometric_deep_learning}

The theoretical background for implementing the deep learning model was presented in section \ref{sec:basics_NN}. In section \ref{sec.DL_implementation}, the focus is on the implementation of these methods to create a deep learning model in the programming language Python. Furthermore, the procedure for generating the dataset, which is used for training, validation and testing, is presented. 
In addition to that, section \ref{sec.DL_results} presents the results of the geometric deep learning model for queue length estimation and compares the performance to the conventional Liu-method. 

\section{Implementation} \label{sec.DL_implementation}

There has been a lot of progress in the recent years regarding software tools, which are developed to simplify the implementation of powerful deep learning applications. The most common software library in Python, which is used in the field of deep learning, is \textit{Tensorflow} \cite{martin_abadi_tensorflow:_2015}. It is an open source software library which is developed to perform numerical computation on a wide range of platforms (e.g. CPUs and GPUs on desktop computers but also on server clusters). It was originally developed by the Google Brain Team and became open source in 2015. Another important Python software library in this field is \textit{Keras} \cite{chollet_keras_2015}. The goal of Keras is to provide a high-level API (application program interface), which simplifies the implementation of machine learning and deep learning solutions even more. With Keras it is possible to prototype solutions very fast and generate early results. The focus of Keras is on user friendliness, modularity and easy extensibility by custom layers. Keras works on top of other, more fundamental, deep learning libraries like Tensorflow or Theano. In this thesis, the high-level API Keras is used with Tensorflow as backend. That means, all the code written with Keras is converted to Tensorflow code, which runs then as Python code on the CPU or GPU. 

\begin{figure}[H]
	\centering
	\includegraphics[width=1\textwidth]{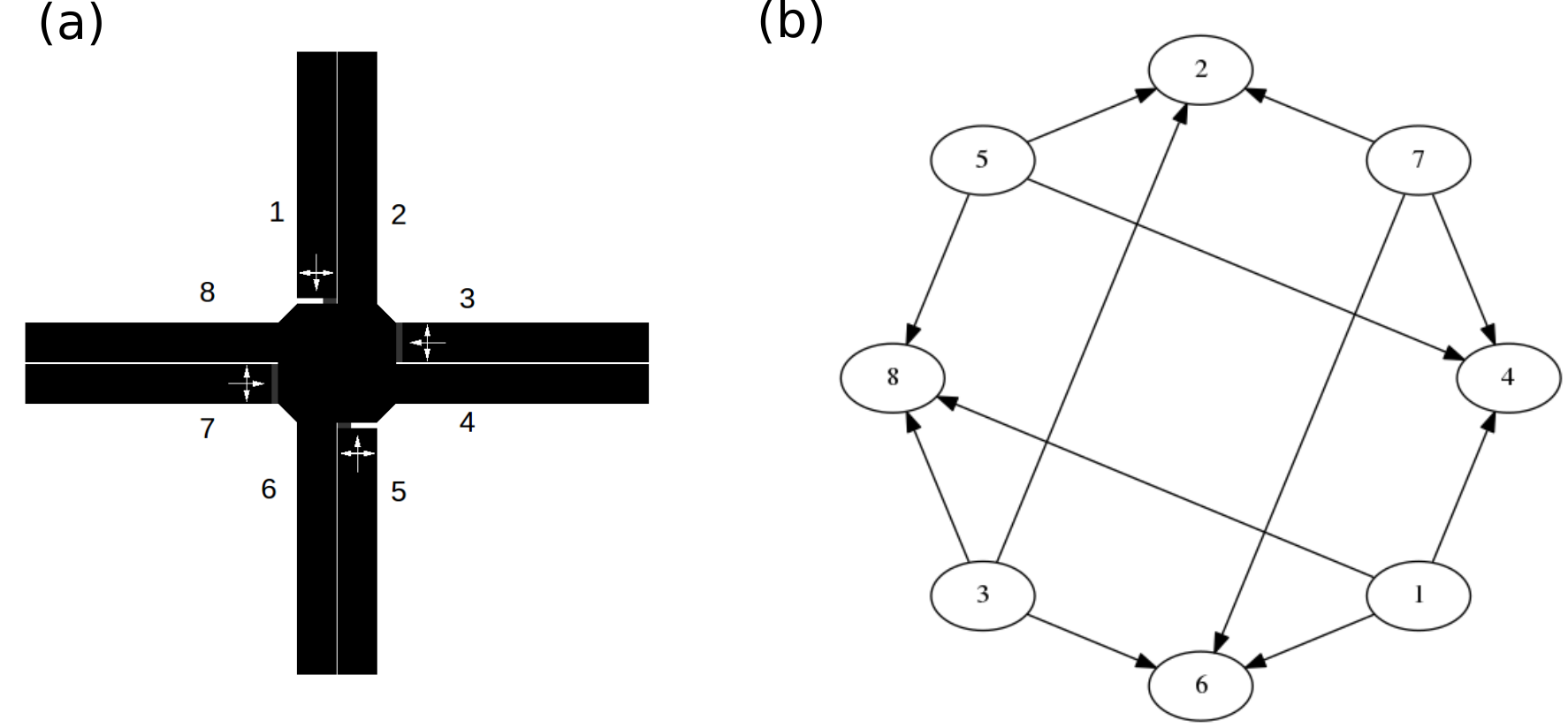}
	\caption[Abstracting graph with nodes and edges from rod network] {Abstracting graph with nodes and edges from road network; (a) represents the classic road network in SUMO; (b) visualizes the graph representation with nodes and edges \cite{wright_geometric_2018}. } \label{bild.NetworkX}
\end{figure}

Another important part of the implementation is to create a graph, which is an abstract representation of the road network topology. The knowledge about the graph is crucial for the graph attention layer used in the deep learning model. Fortunately, SUMO is able to export the structure of the road network as a graph using the open source Python library \textit{NetworkX} \cite{hagberg_exploring_2008}. In NetworkX it is possible to have an abstract representation of the road network as a NetworkX-graph which consists of nodes and edges. The nodes are the lanes, since they are the objects of interests with information about traffic. The lanes are connected by intersections, so in this case the intersections are represented by edges. This method of abstraction is visualized in fig. \ref{bild.NetworkX}.

%create figure which is similar to the poster from Matt

On the right side (a), the classic road network with only one lane for each direction (to keep it simple) is presented. Each lane represents a node with a certain number from 1 to 8. These lanes are connected by the intersection. The figure (b) visualizes the abstracted graph representation for this example intersection. The ellipses represent the nodes (lanes with each number) which are connected by edges (straight lines). For instance cars that arrive on lane 1 in figure (a): They can leave the intersection either on lane 8 (right turn), lane 6 (straight) or lane 4 (left turn). Looking at figure (b), the node 1 is connected by arrows to node 8, 6 and 4. In this particular case it is a \textit{directed graph}, since the edges have a direction (e.g. cars can only change from lane 1 to 8 but not vice versa). If the edges do not have a certain direction, it is called an \textit{undirected graph}. The library NetworkX provides a function to abstract the adjacency matrix from the graph. In case of the example in fig. \ref{bild.NetworkX}, the adjacency matrix would be
\begin{equation}
A =  \begin{pmatrix} 
0 & 0 & 0 & 1 & 0 & 1 & 0 & 1 \\
0 & 0 & 0 & 0 & 0 & 0 & 0 & 0 \\
0 & 1 & 0 & 0 & 0 & 1 & 0 & 1 \\
0 & 0 & 0 & 0 & 0 & 0 & 0 & 0 \\
0 & 1 & 0 & 1 & 0 & 0 & 0 & 1 \\
0 & 0 & 0 & 0 & 0 & 0 & 0 & 0 \\
0 & 1 & 0 & 1 & 0 & 1 & 0 & 0 \\
0 & 0 & 0 & 0 & 0 & 0 & 0 & 0 
\end{pmatrix}.
\end{equation}

For the usage in the graph attention layer, it is necessary to add the presented adjacency matrix to an identity matrix of the same size, so that the main diagonal of the matrix are ones. Just by that, each lane is able to consider also its own features as input rather than just its neighbors. Of course for larger road networks, like in this thesis, the road networks have more lanes and the adjacency matrix $A$ becomes larger. Especially for this thesis, $A$ includes the upstream and downstream connections of each lane, but not the neighboring lanes which are eligible for lane changing. So this is a good point to start, since the complexity of the adjacency matrix is not too high. The calculated adjacency matrix is then an input for the GAT layer and does not change over time. This is how the spatial structure of the road network is provided to the deep learning model.

The other very important input for the deep learning model are the features. In this particular approach, eight input features for each lane for every time step are used. The first six features come from the e1-detectors in SUMO, which include the \textit{number of contributed vehicles}, \textit{occupancy} and the \textit{speed} of the vehicles for the stop bar and the advanced detector. The e1 detectors provide the data for every second. The seventh input is the information about the traffic light signal at the end of the lane. It is a second-by-second binary signal which is one, if the traffic light is green and zero otherwise. The eighth feature is the estimation results of the method after \cite{liu_real-time_2009} presented in chapter \ref{sec.liu_method}. Although the estimation of this method is not accurate on larger road networks due to lane changing and stop an go traffic, it can help to guide the estimation in the right direction. The information provided by this method is just the maximum queue length for an entire traffic light cycle. Therefore, it is necessary to process a linear interpolation between the maximum length, to create second-by-second data as well.
Once, the second-by-second data for the whole simulation for every lane are collected, the features have to be gathered in a design matrix $\boldsymbol{X}$. The design matrix is assembled to the shape $[simulations \times timesteps \times lanes \times features]$. Generally, this is an array of four dimensions, where the features for each time step are collected for each lane and put together in a two dimensional matrix. In the deep learning field a four dimensional vector is also often called a \textit{tensor}. Furthermore, this is done for every time step of the simulation, which yields the third dimension. Finally, data from multiple simulations can be fed into the network, which results then in the fourth dimension. 
The same procedure can be executed for building the $Y$ array, which has the same structure except of replacing the features with the targets. In this approach the number of targets is two. The first target is the maximum queue length measured by the e2-detector in SUMO. The algorithm gives us the ground-truth data of the maximum queue length for each traffic light cycle. To get second-by-second data, a linear interpolation has to be made between the data points. The second target is the number of vehicles that are on the entire lane, and it is not relevant if they are part of the queue or not. These data can be provided by the e2-detector as well by extracting the value \textit{nVehSeen} for each second. Finally, the structure of $\boldsymbol{Y}$ is $[simulations \times timesteps \times lanes \times targets]$

\begin{figure}[h]
	\centering
	\includegraphics[width=0.6\textwidth]{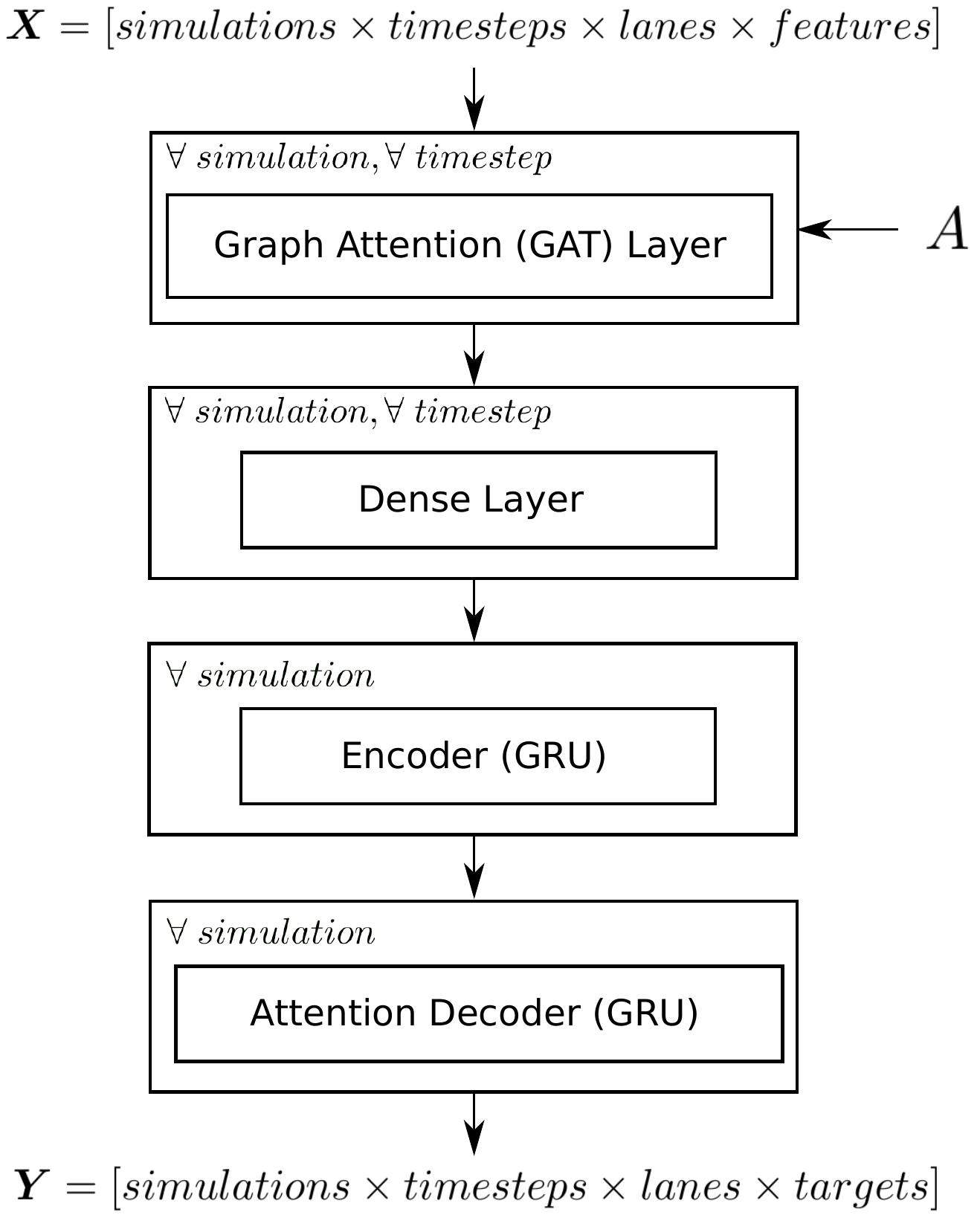}
	\caption[Topology of the deep learning model] {Topology of the deep learning model. } \label{bild.DL_model}
\end{figure}

After building $\boldsymbol{X}$ and $\boldsymbol{Y}$, the second-by-second data can be averaged over a certain time interval. In this thesis the time interval is chosen to be 10 seconds. By that, the size of the input data is reduced and the information density increased. $\boldsymbol{X}$ and $\boldsymbol{Y}$ can  now be used to train the model. The used model structure is shown in fig. \ref{bild.DL_model}.
The design matrix $\boldsymbol{X}$ is fed into the first GAT layer, which is supposed to learn spatial attention between the lanes. As a second input, the GAT layer receives the adjacency matrix $\boldsymbol{A}$. In fig. \ref{bild.DL_model} only one GAT layer for simplicity is shown. In the final approach two GAT layer are stacked together behind each other. The number of output features for both GAT layer is chosen to be 128, which was determined iteratively by trial and error. Generally, it is recommended to choose the number of features as a power of two to optimize storage usage. Since the GAT layer only take a matrix of the dimension $lanes \times features$ as an input to process the learning algorithm presented in \ref{sec:gat_layer}, wrapper-code is necessary, which feeds in the matrix for all time steps and for all simulations [$\forall simulations, \forall timesteps$]. The following Dense layer are fully connected layer which are presented in sec. \ref{sec:basics_NN}. The dense layer with the width of 128 are supposed to rearrange the learned features from the GAT layer in a way it is necessary for the following encoder-decoder structure. For this layer, the same wrapper-code is necessary, since fully connected layer only take a matrix as input. Following, the encoder takes the whole data for each simulation as an input, including the time steps. So for the encoder, the wrapper code is not necessary anymore, since the encoder handles the time steps sequentially (see sec. \ref{sec:enc-dec}). The inputs per time step are encoded to hidden states, which are used by the following attention decoder to decode the hidden states back into the output time sequence. The temporal attention mechanism is included in the decoder. Both, the encoder and attention decoder, have 128 units per hidden state. Generally, in this approach two different attention mechanism are used. In the GAT layer spatial attention is applied, while the decoder uses temporal attention to estimate the output sequence. 
Furthermore, $L^2$ regularization and dropout (see sec. \ref{sec:basics_NN}) after each GAT and dense layer are applied to avoid overfitting. During training, the loss function for the optimization algorithm is chosen as the mean squared error (MSE) between the prediction and the ground-truth data.

\begin{figure}[h]
	\centering
	\includegraphics[width=0.8\textwidth]{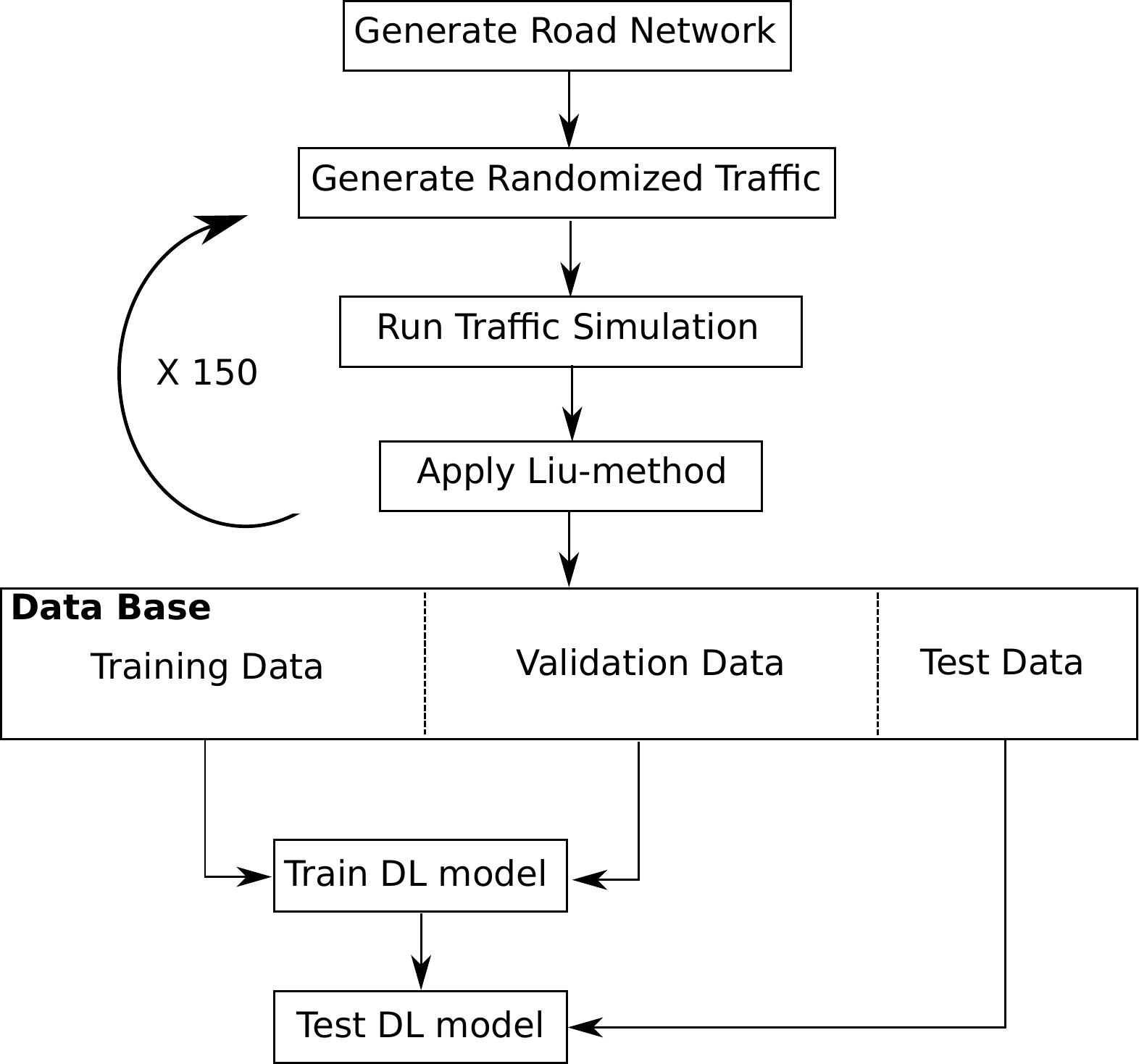}
	\caption[Software workflow for training the DL model] {Software workflow for training the DL model.} \label{bild.software_workflow}
\end{figure}

The entire procedure for training and testing of the deep learning model is presented in fig. \ref{bild.software_workflow} and starts with running several traffic simulations (e.g. 150) with all the same settings in SUMO on the same generated road-network and collecting for each simulation the detector data. The single SUMO simulations are all different from each other although they have the same settings, since the traffic in SUMO is randomized. The used road network in SUMO is exactly the same than in sec. \ref{sec.model_grid}. Based on the detector data the queue length estimation after chapter \ref{sec.liu_method} is executed for each simulation. Afterwards the collection of simulations are divided into training, validation as well as the test data and the $\boldsymbol{X}$ and $\boldsymbol{Y}$ matrices are build. During training, the gradient updates for the parameter are calculated based on the training data. The validation data are also used during training to calculate the validation loss and monitoring the success of the training regarding under- or overfitting. After the training is finished, the model is tested on the test data, which it has never seen before during training, to evaluate the performance. The results of this test data are presented in the following chapter. 

\section{Results} \label{sec.DL_results}

In this chapter the results of the introduced deep learning model are presented. The first part is about the practical usage of the mean average error (MAPE), and which problems result based on that. Furthermore, an alternative metric for evaluation, the \textit{mean average error} is introduced. Afterwards, the results from selected lanes are presented and the performance of the deep learning model is compared to the method of \cite{liu_real-time_2009}. In addition to that, the overall performance for the entire network for multiple simulations is presented. Afterwards, the difference of the estimation accuracy is discussed, in case the results from the method presented in \ref{sec.liu_method} (here in future called \textit{Liu-method}) is not an input feature for the deep learning model. Finally, the performance of the model is validated in case traffic light cycles or traffic arrival flow changes for the test data, compared to the training data. 

\begin{figure}[h]
	\centering
	\includegraphics[width=1\textwidth]{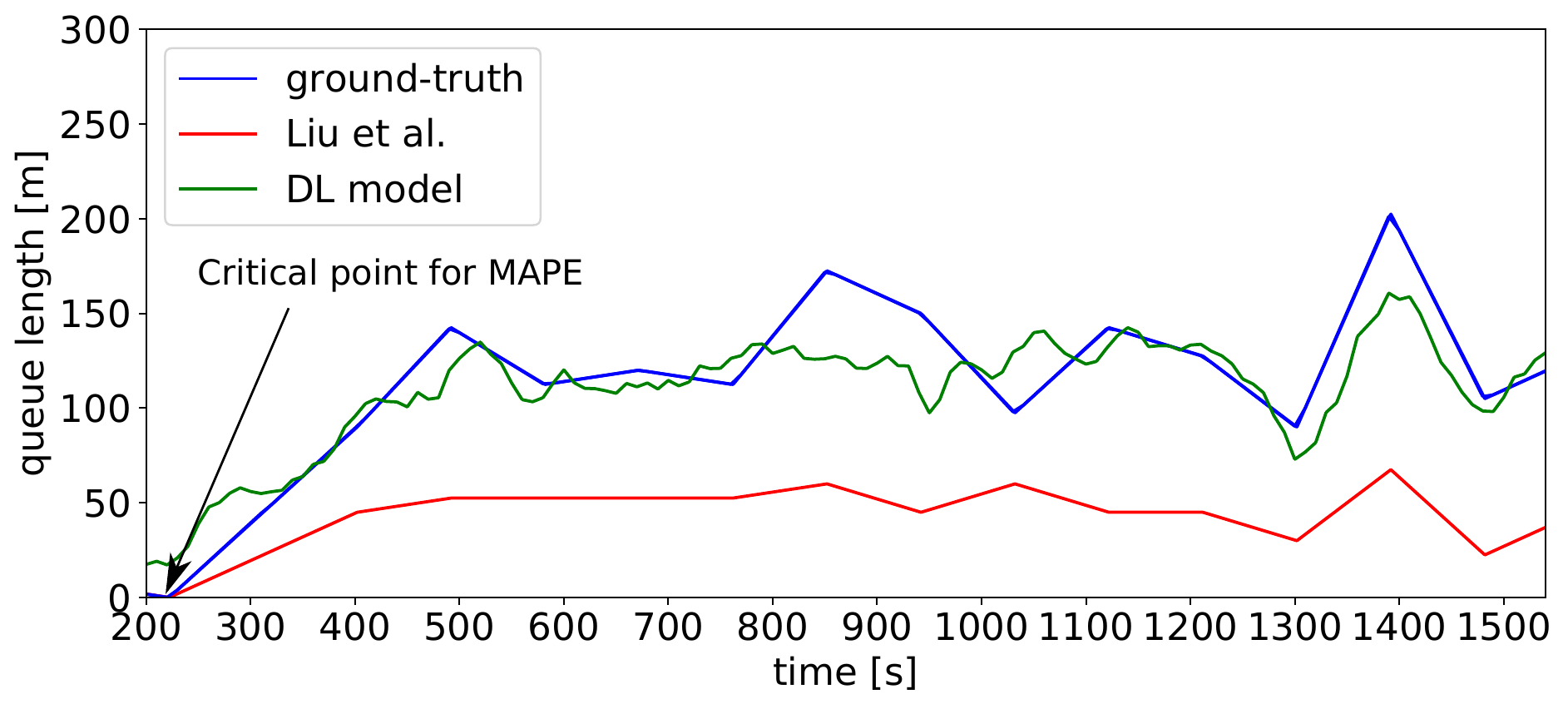}
	\caption[Example for not reliable MAPE] {
		Example for not reliable MAPE: \\
		\begin{tabular}{c|c|c|} 
			\cline{2-3}
			& \multicolumn{2}{c|}{\textbf{Queue estimation:}} \\
			& \textbf{DL model}     & \textbf{Liu-method}     \\ \hline
			\multicolumn{1}{|c|}{MAE {[}m{]}}           & 15                  & 67.1          \\ \hline
			\multicolumn{1}{|c|}{MAPE {[}\%{]}} & 190.9                  & 53.4           \\ \hline
		\end{tabular}
	} \label{bild.MAPE_problem}
\end{figure}

Before the final results are presented, it is important to emphasize the limitations of the MAPE for evaluation of the performance. Due to the fact that the MAPE measures the percentage error, the metric is not reliable anymore, when the ground-truth value goes close to zero, while the estimation has a larger value. This correlation is presented in fig. \ref{bild.MAPE_problem}, which is a real prediction from the trained DL model. The MAPE of the DL model (190.9\%) is worse than the MAPE of the Liu-method (53.4\%), although the DL model has a much better performance considering the impression of the graph in fig. \ref{bild.MAPE_problem}. The reason for that is located at the beginning of the time axis after 200 seconds. The ground-truth value is close to zero, while the estimated value of the DL model is at around 20 meters. Considering the the formula of the MAPE which is

\begin{equation}
MAPE = \frac{1}{m} \sum_{m} \bigm| \frac{Observation - Estimation}{Observation} \bigm| \times 100 \%,
\end{equation}

the phenomenon can be explained. When the observation is close to zero but the estimation is not, the result of the fraction becomes very high and does not represent the real performance of the estimation anymore. An alternative is the \textit{mean average error (MAE)}

\begin{equation}
MAE = \frac{1}{m} \sum_{m} \bigm| Observation - Estimation \bigm|.
\end{equation}

The MAE does not have the problem of the exploding value in some cases, and is still interpretable, since it has the same dimension as the estimation (in this case meters or number of vehicle). Regarding fig. \ref{bild.MAPE_problem}, the MAE for the DL model is 15.0 meters while for the Liu-method 67.1 meters. In the following both metrics (MAPE and MAE) are presented, but in some cases the MAPE does not give reliable results.

\subsection{General Results}

\begin{figure}[H]
	\centering
	\includegraphics[width=1\textwidth]{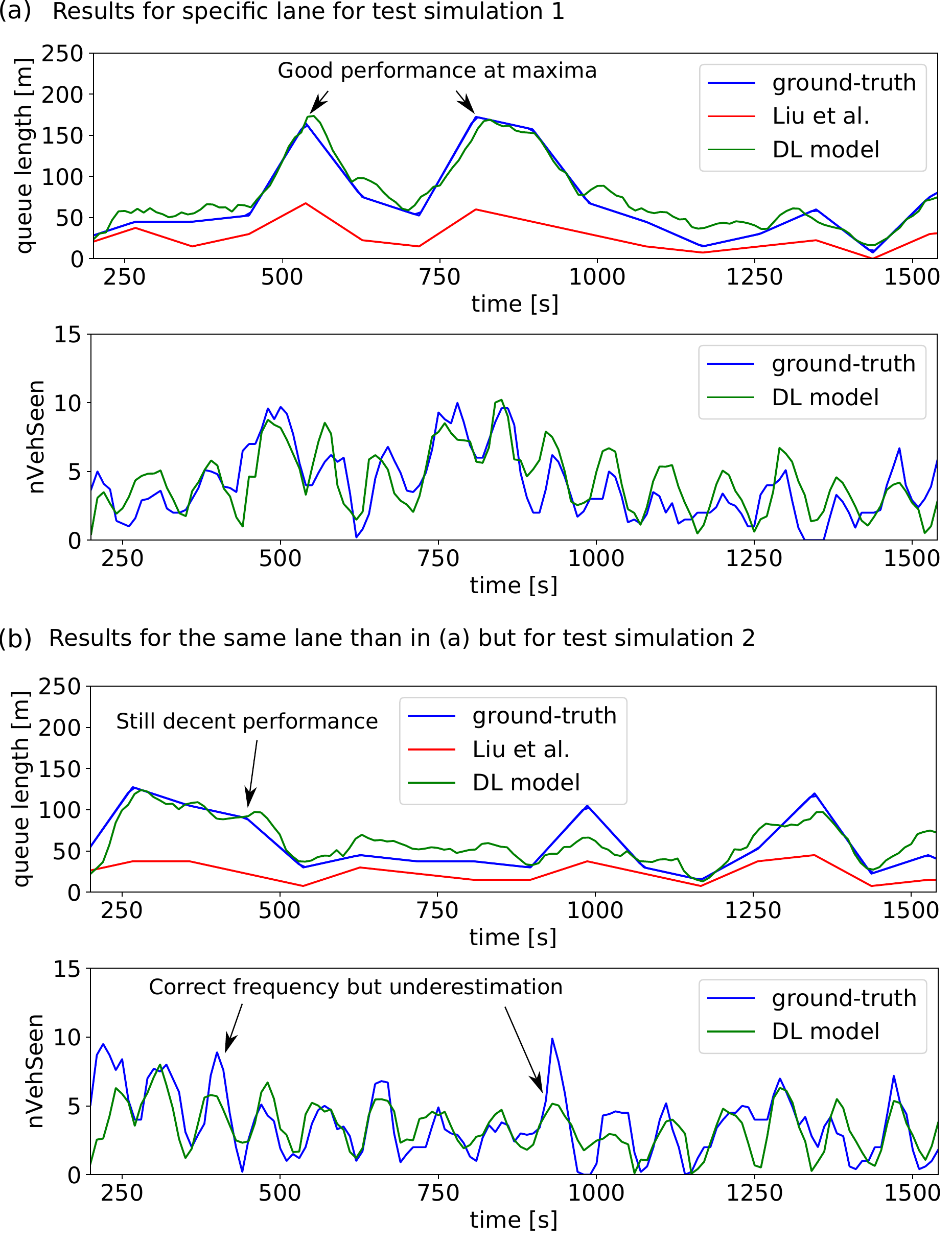}
	\caption[Comparison of results for the same lane from different test simulations] {Comparison of results for the same lane from different test simulations (DL model is the same). \\
		\begin{tabular}{c|c|c|c|}
			\cline{2-4}
			& \multicolumn{2}{c|}{\textbf{Queue estimation:}} & \textbf{nVehSeen:} \\
			& \textbf{DL model}     & \textbf{Liu-method}     & \textbf{DL model}  \\ \hline
			\multicolumn{1}{|c|}{MAE (a)}           & 10.5 m                & 43.2 m                  & 1.9 veh            \\ \hline
			\multicolumn{1}{|c|}{MAPE (a) {[}\%{]}} & 24.7                  & 59.3                    & 61124881.6         \\ \hline
			\multicolumn{1}{|c|}{MAE (b)}           & 13.2 m                & 31.8 m                  & 1.3 veh            \\ \hline
			\multicolumn{1}{|c|}{MAPE (b) {[}\%{]}} & 27.2                  & 50.4                    & 50774859.8         \\ \hline
		\end{tabular}} \label{bild.compare_simus}
\end{figure}

In fig. \ref{bild.compare_simus} the results for another lane in the network are presented. The traffic is always randomized and the arrival rate for train and test data is 2.5 veh/sec. Figure \ref{bild.compare_simus} (a) shows the estimated queue length from the DL model and Liu-method compared to the ground-truth data in the upper graph. The performance of the deep learning model is very accurate, it follows the ground-truth data very well. Especially the two maximums at 500 and 800 seconds are good estimated.
 
This is also represented by the MAPE (24.7\%) and MAE (10.5 m). The Liu-method underestimates, which is caused by the effect of stop-and-go traffic, discussed in sec. \ref{sec.liu_on_grid}, since the presented lane is a left-turn lane. 
The breakpoint C is detected too early, which results in continuous underestimation.
The MAPE for the Liu-method is 59.3\% and the MAE 43.2 meter, which attest the better performance of the DL model. The Liu-method has the same issue for almost every left-turn lane in the road-network.
In the lower graph of (a) the estimation for the number of vehicles on the entire lane are shown. Due to the fact, that second-by-second data are used which are averaged to 10 second intervals, the graph seems to have a higher resolution than the upper graph for the queue length, where only one value per traffic light cycle is linear interpolated. Generally, the ground frequency depends on the traffic light phases. The neural network is able to detect the frequency and tries to estimate the peaks, which is not everywhere accurate. With a MAE of 1.9 vehicle (the MAPE is not reliable in this case), the performance is acceptable, but not perfect yet. 
In fig. \ref{bild.compare_simus} (b) results for exact the same lane in the same road network are shown. They are predicted by the same DL model, just for another test simulation.

\begin{figure}[H]
	\centering
	\includegraphics[width=1\textwidth]{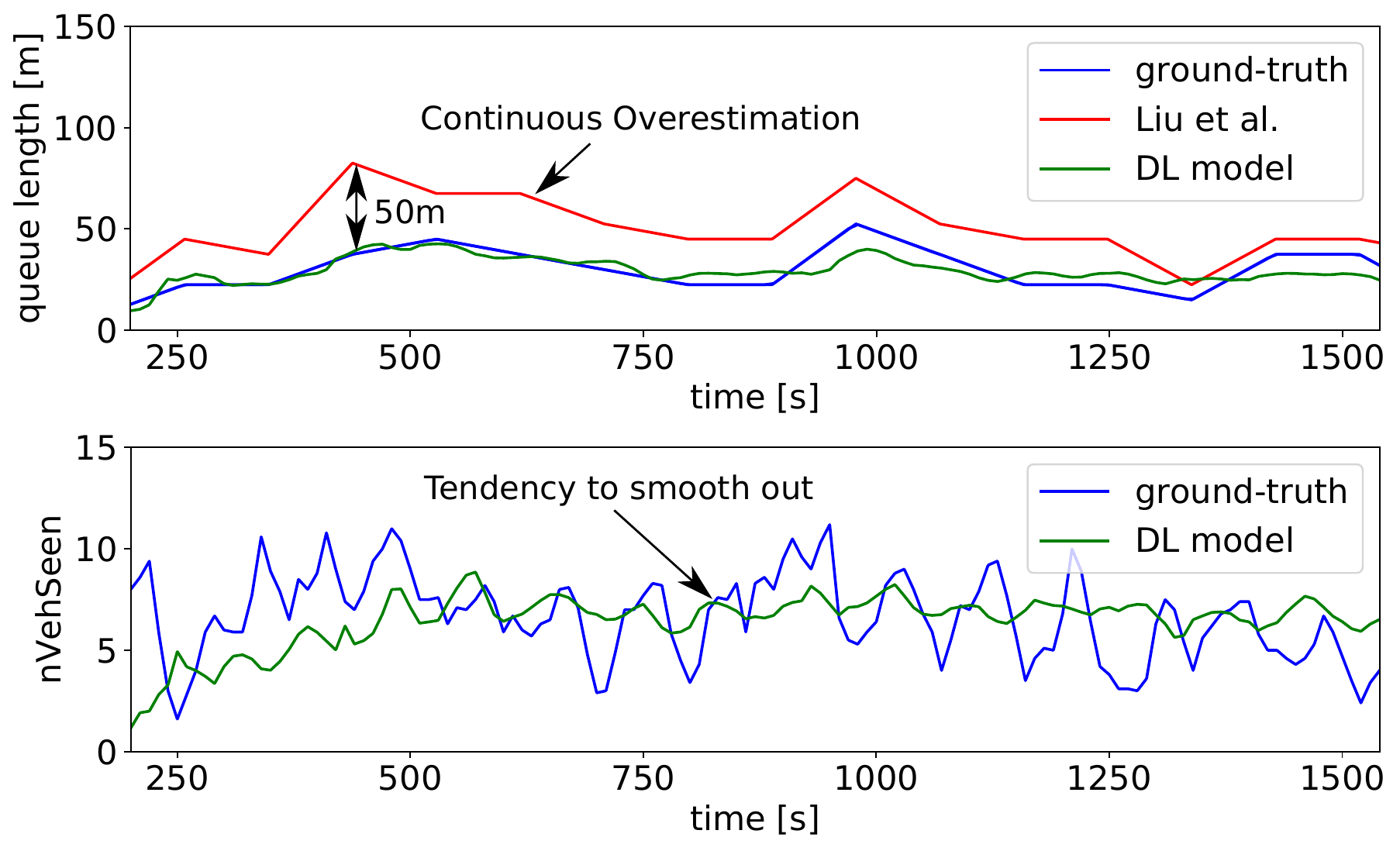}
	\caption[Estimation results for a right-turn lane] {Estimation results for a right-turn lane. \\
		\begin{tabular}{c|c|c|c|}
			\cline{2-4}
			& \multicolumn{2}{c|}{\textbf{Queue estimation:}} & \textbf{nVehSeen:} \\
			& \textbf{DL model}     & \textbf{Liu-method}     & \textbf{DL model}  \\ \hline
			\multicolumn{1}{|c|}{MAE}           & 4.8 m                 & 19.8 m                  & 1.9 veh            \\ \hline
			\multicolumn{1}{|c|}{MAPE {[}\%{]}} & 16.5                  & 77.3                    & 33.9               \\ \hline
		\end{tabular}} \label{bild.right_turn_exmpl}
	\end{figure}
This should emphasize two things: First, the simulation data generated by SUMO, which are used for training, validation and testing, are very different from each other. 	
The goal is to have a high diversity in the data, so that the deep learning model is able to generalize knowledge and does not overfit.  
Second, although the DL model performs very well in (a), it has still a satisfying performance in (b) for a completely different ground-truth, which is also caused by the high diversity in the training data. The MAPE for the queue length in (b) is 27.2\% and the MAE is 13.2 meters. The Liu-method also suffers here from the stop-and-go traffic and underestimates.

In fig. \ref{bild.right_turn_exmpl} an example for a right-turn lane is shown. In this case, the DL model has satisfying performance with a MAPE of 16.5\% (MAE = 4.8 m), although it starts to "smooth out" after 1000 seconds. This effect could be caused by to many time steps that are taken into consideration for the decoding. The Liu-method is overestimating, since the lane is a right-turn lane where a lot of lane changes occur. Vehicles are changing after passing the advanced detector to other lanes, which cannot be handled by the Liu-method (discussed in sec. \ref{sec.liu_on_grid}). This is also shown by the MAPE (77.3\%) and MAE (19.8 m). For the estimation of nVehSeen, the curve is again smoothing and not very accurate. The MAPE (33.9\%) and MAE (1.9 veh) is also not completely satisfying. Also this could be caused by to many time steps, that are taken into consideration for the attention decoding. In addition to that underestimation until the first 500 seconds occurs. This could be a warm-up phase, since the deep learning model has to less time steps in the past to estimate properly. This phenomenon occurs only at the beginning of an estimation. Especially for estimating the second target (nVehSeen), the optimal model configuration is not found yet. However, note that the approach has a lot of potential so far. This is also emphasized by considering the averaged MAPE and MAE for the entire road network (not only for a single lane, as presented above). In table \ref{tab.MAE_entire_net} the MAEs for several test simulations are presented. The MAPEs are not reliable sources, since at least one lane has a MAPE which has a very high (and not reliable) value.

\begin{table}[h]
\centering
\begin{tabular}{c|c|c|c}
	& \multicolumn{2}{c|}{\textbf{MAE queue length {[}m{]}}} & \textbf{MAE nVehSeen {[}veh{]}} \\ \hline
	\textbf{Simulation \#} & \textbf{DL model}         & \textbf{Liu-method}        & \textbf{DL model}               \\ \hline
	1                      & 9.1                       & 18.1                       & 1.4                             \\
	2                      & 9.5                       & 17.8                       & 1.5                             \\
	3                      & 9.0                       & 18.6                       & 1.6                             \\
	4                      & 8.9                       & 18.1                       & 1.4                             \\
	5                      & 9.4                       & 18.5                       & 1.5                             \\ \hline
	\textbf{Average:}      & 9.2                       & 18.2                       & 1.5                            
\end{tabular}
\caption[Estimation results (MAE) of an entire road-network] {Estimation results (MAE) of an entire road-network for five test simulations.} \label{tab.MAE_entire_net}
\end{table}

It turns out, that the performance of the deep learning model is very constant over different independent test simulations. In addition to that, the DL model has around twice better performance regarding the MAE than the Liu-method. For the number of vehicles on the entire lane (nVehSeen), the deep learning model has a MAPE around 1.5 vehicles, which is acceptable considering the number of vehicles that are on the entire lane, but there is no reference to other methods available.  

\subsection{Robustness for different Arrival Rates} \label{sec.DL_robust}

\begin{figure}[h]
\centering
\includegraphics[width=1\textwidth]{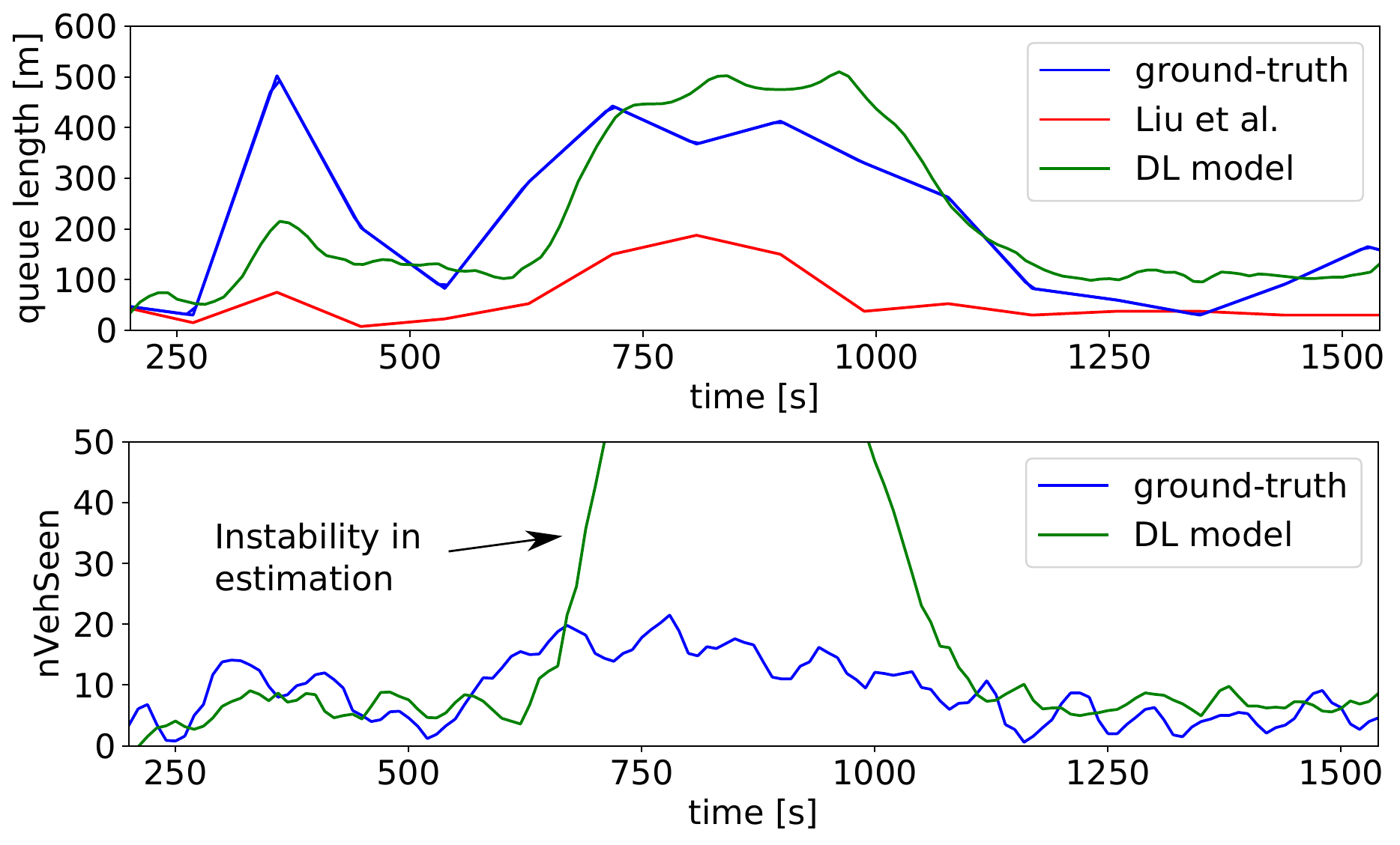}
\caption[Test of the DL model for different traffic arrival rate] {Test of the DL model for different traffic arrival rate of 3.33 veh/s instead of 2.5 veh/sec.\\
	\begin{tabular}{c|c|c|c|}
		\cline{2-4}
		& \multicolumn{2}{c|}{\textbf{Queue estimation:}} & \textbf{nVehSeen:} \\
		& \textbf{DL model}     & \textbf{Liu-method}     & \textbf{DL model}  \\ \hline
		\multicolumn{1}{|c|}{MAE}           & 75.6 m                & 149.6 m                 & 13.9 veh           \\ \hline
		\multicolumn{1}{|c|}{MAPE {[}\%{]}} & 45.6                  & 63.7                    & 148.5              \\ \hline
	\end{tabular}
} \label{bild.03_period}
\end{figure}

Although the performance, especially of the queue length estimation, is satisfying, it is important to note that the traffic parameter are the same for training and testing. For example the arrival rate of the traffic is the same, although the traffic routes for the single vehicles are randomized by SUMO. It would be interesting to analyze, how the DL model behaves when the input data have different arrival rates. This study is shown in fig. \ref{bild.03_period} for a selected left-turn lane.
The presented test data have a arrival rate of 3.33 veh/s while the DL model is only trained on data with 2.5 veh/s arrival rate. Regarding the estimation of the queue length, the DL model has some problems to estimate accurately the first maximum. But it still has with a MAE of 75.6 meters a better performance than the Liu-method with MAE = 149.6. The estimation of the nVehSeen is more critical. In the period of time between 650 and 1150 seconds, the estimated vehicles on the lane reach very high values. The DL model shows instable behavior. However, this behavior is expected, since it has never seen a nVehSeen of 20 during training before. So it has no knowledge about how to handle this situation. This is a common problem for any sort of statistical model which is used for prediction out of samples. The worse performance is also illustrated by the average MAEs for all lanes. The MAE for the DL model is 31.5 m, for the Liu-method 56.2 m and for the DL model estimating the nVehSeen the MAE is 5.9 vehicle. 

\begin{figure}[h]
\centering
\includegraphics[width=1\textwidth]{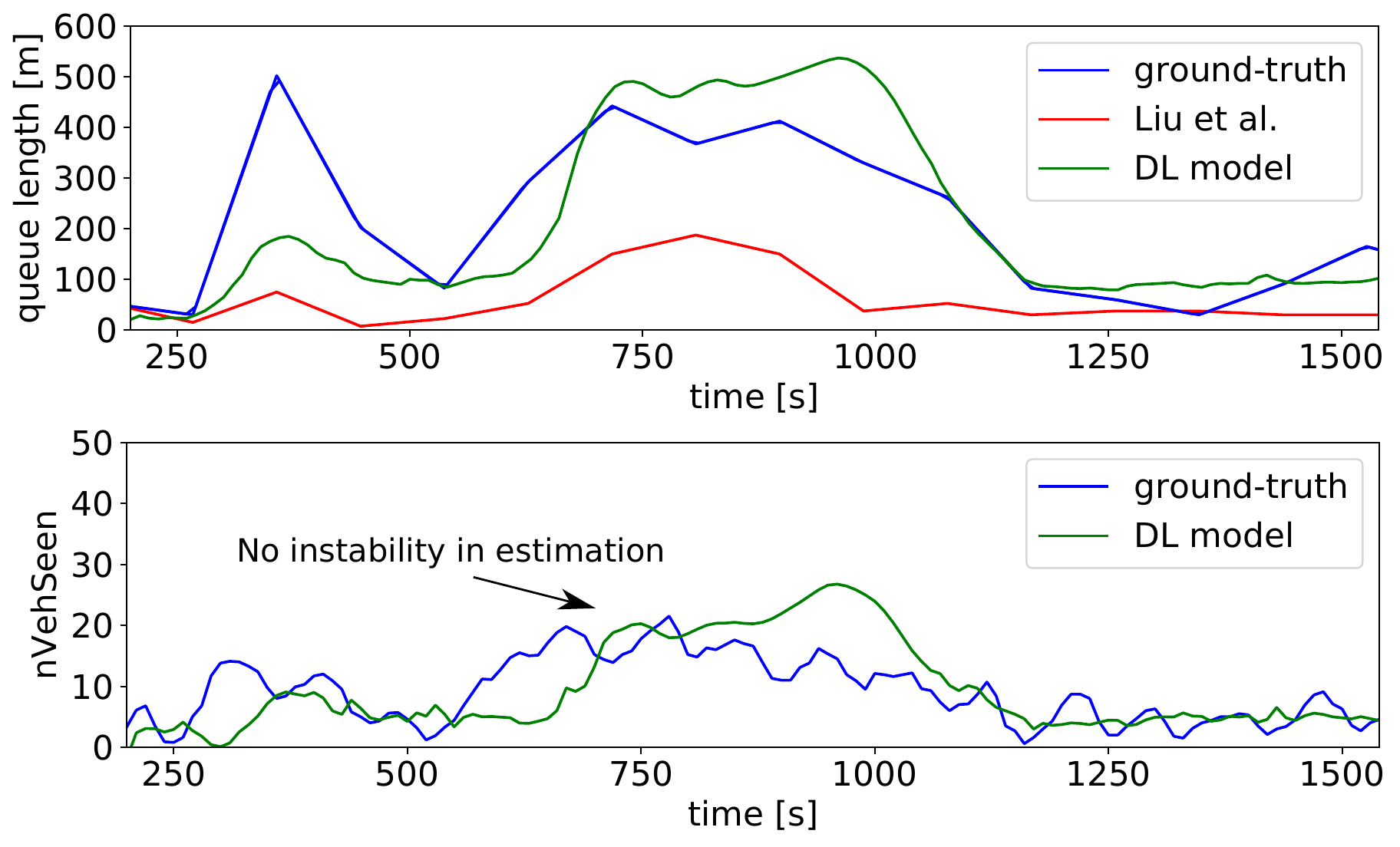}
\caption[Test of the DL model for traffic arrival rate of 3.33 veh/sec, when model is trained on multiple arrival rates] {Test of the DL model for traffic arrival rate of 3.33 veh/sec, when model is trained on multiple arrival rates.\\
	\begin{tabular}{c|c|c|c|}
		\cline{2-4}
		& \multicolumn{2}{c|}{\textbf{Queue estimation:}} & \textbf{nVehSeen:} \\
		& \textbf{DL model}     & \textbf{Liu-method}     & \textbf{DL model}  \\ \hline
		\multicolumn{1}{|c|}{MAE}           & 77.8 m                & 149.5 m                 & 4.4 veh           \\ \hline
		\multicolumn{1}{|c|}{MAPE {[}\%{]}} & 39.4                 & 63.7                    & 61.3              \\ \hline
	\end{tabular}\label{bild.03_period_new}
} 
\end{figure}

The performance can be improved, if the model is not only trained with simulation data based on a specific arrival rate, but with a high diversity of arrival rates. If the deep learning model has a higher diversity in the training data, the generalization of knowledge is better. In fig. \ref{bild.03_period_new} the results are shown for the same DL model but trained with 180 simulations with a randomized arrival rate between 2.22 veh/sec and 4 veh/sec for each simulation, rather than with 125 simulations of 2.5 veh/sec data as for fig. \ref{bild.03_period}. Since the data have a higher diversity, more simulations are necessary to allow similar performance. 

The performance of the queue length estimation in fig. \ref{bild.03_period_new} is not better compared to fig. \ref{bild.03_period} which has a MAE of (75.6 m). The reason is that 180 simulations are too less for training a model properly for all possible arrival rates. But it is still enough to stabilize the prediction for nVehSeen in the lower graph of fig. \ref{bild.03_period_new}. While in fig. \ref{bild.03_period} the nVehSeen becomes unstable and have a MAE of 13.9 veh, in fig. \ref{bild.03_period_new} the behavior is stable with a MAE of 4.4 veh. The model has already seen similar traffic situations during training and can handle the situation. The average MAEs for the entire network are compared in table \ref{tab.compare} for both different models.

\begin{table}[H] 
\centering
\begin{tabular}{|c|c|c|c|c|}
	\hline
	\multirow{2}{*}{\textbf{\begin{tabular}[c]{@{}c@{}}Arrival rate\\  for training\\ {[}veh/sec{]}\end{tabular}}} & \multirow{2}{*}{\textbf{\begin{tabular}[c]{@{}c@{}}Number of\\ training \\ simulations\end{tabular}}} & \multicolumn{2}{c|}{\textbf{\begin{tabular}[c]{@{}c@{}}Queue estimation\\ MAE\\ (MAD)\end{tabular}}}                   & \textbf{\begin{tabular}[c]{@{}c@{}}nVehSeen\\ MAE\\ (MAD)\end{tabular}} \\ \cline{3-5} 
	&                                                                                                       & \textbf{DL model}                                          & \textbf{Liu-method}                                       & \textbf{DL model}                                                       \\ \hline
	\textbf{2.5}                                                                                                   & \textbf{125}                                                                                          & \begin{tabular}[c]{@{}c@{}}31.5 m \\ (30.2 m)\end{tabular} & \begin{tabular}[c]{@{}c@{}}56.2 m\\ (66.4 m)\end{tabular} & \begin{tabular}[c]{@{}c@{}}5.9 veh\\ (6.8 veh)\end{tabular}             \\ \hline
	\textbf{2.22 - 4}                                                                                           & \textbf{180}                                                                                          & \begin{tabular}[c]{@{}c@{}}24.7 m\\ (21.1 m)\end{tabular}  & \begin{tabular}[c]{@{}c@{}}56.2 m\\ (66.4 m)\end{tabular} & \begin{tabular}[c]{@{}c@{}}3.8 veh\\ (3.2 veh)\end{tabular}             \\ \hline
\end{tabular}
\caption[MAEs and MADs for the entire road network predicted by the same model which was trained on datasets with different arrival rates] {MAEs and MADs for the entire road network predicted by the same model which was trained on datasets with different arrival rates.} \label{tab.compare}
\end{table}

The MAE for the queue estimation is better for the deep learning model trained on multiple arrival rates compared to the model trained on only one arrival rate. Even better is the improvement for the estimated nVehSeen. As additional metric the \textit{mean absolute deviation (MAD)} is used, which is 
\begin{equation}
MAD = \frac{1}{m} \sum_{i = 1}^{m} \mid MAE_i - \overline{MAE} \mid,
\end{equation}
where $m$ is the number of lanes and $\overline{MAE}$ is the average MAE for all lanes. The MAD is very high for both models due to the instable situations that occur for both. Still, the model trained on more arrival rates, has a lower MAD in all categories. 
Furthermore, the DL model does not become instable in certain situations anymore. While for the old model on 16 of 120 lanes instability occurs, the improved model has 5 lanes where the nVehSeen becomes instable and overestimation occurs. The improved model has not the optimal performance yet, since a higher amount of simulations are necessary to gain reliability.
However, this demonstration should emphasize how important it is to train on many different relevant data as possible to have a maximum of generalized knowledge for every traffic case.

% show how dl model performs on changed traffic light cycles(optional if time left)
%train DL model with different traffic light cycles (optinal)

%-> verify that using the results from liu are gaining the accuracy of the estimation -> come to the conclusion, that it was worth it to do the preliminary work with liu

\subsection{Influence of Liu-Method as Input} \label{sec.Liu_as_input}

\begin{figure}[H]
\centering
\includegraphics[width=1\textwidth]{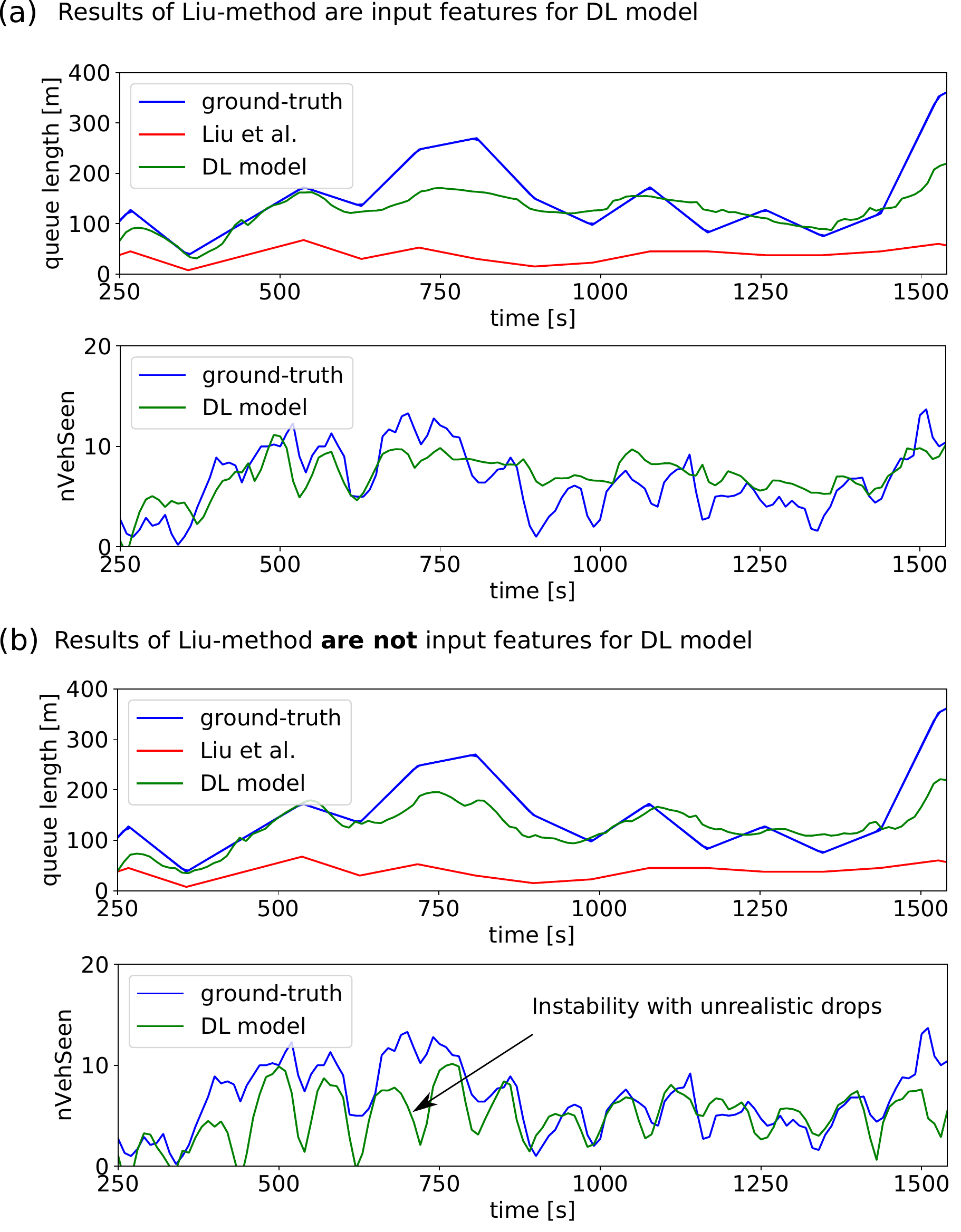}
\caption[Comparison of results for the same lane if results of Liu-method are input feature or not] {Comparison of results for the same lane if results of Liu-method are input feature or not. \\
	\begin{tabular}{c|c|c|c|}
		\cline{2-4}
		& \multicolumn{2}{c|}{\textbf{Queue estimation:}} & \textbf{nVehSeen:} \\
		& \textbf{DL model}     & \textbf{Liu-method}     & \textbf{DL model}  \\ \hline
		\multicolumn{1}{|c|}{MAE (a)}           & 31.1 m                & 121.7 m                  & 2.0 veh            \\ \hline
		\multicolumn{1}{|c|}{MAPE (a) {[}\%{]}} & 18.86                  & 68.2                    & 64.6         \\ \hline
		\multicolumn{1}{|c|}{MAE (b)}           & 32.9 m                & 121.8 m                  & 2.3 veh            \\ \hline
		\multicolumn{1}{|c|}{MAPE (b) {[}\%{]}} & 21.5                  & 68.2                    & 40.9         \\ \hline
	\end{tabular}} \label{bild.without_liu}
\end{figure}

Finally, it is also important to investigate, if the usage of the Liu-method as an input feature is a benefit. Therefore, the exactly same model was trained twice. 
For the first dataset the results of the Liu-method were included (8 input features) which is in the following called \textit{with-Liu model}, for the second dataset the Liu-method was not included (7 input features) also called as \textit{no-Liu model}.  
In fig. \ref{bild.without_liu} the results of both training sessions are presented for the same lane. In fig. \ref{bild.without_liu} (a) the estimated queue length and nVehSeen for the model with Liu-method as input are presented, while in (b) the results for the model without Liu-method are shown. For the comparison of the estimated queue length, the no-Liu model with a MAE of 32.9 m seems to have almost the same performance as the with-Liu model with a MAE of 31.1 m.
The comparison between the estimated nVehSeen is more interesting: While the MAEs are also not significant different (2.0 veh for with-Liu and 2.3 veh for no-Liu) the difference between the graphs are higher. The nVehSeen in (a) (with-Liu) follow the ground truth curve quiet well with a slightly smoothing character. For the nVehSeen in (b) (no-Liu), the behavior of the estimation becomes sometimes instable with many drops, at some points the estimation is even below zero. It is suspected, that this unrealistic behavior is caused by overfitting of the training data.

 Not only on this particular lane, but also on 15 other lanes the estimated value drops 25 times in total below zero. This does not happen for the DL model in trained with the Liu-method as input. The average MAEs as well as the MADs of both models are compared in table \ref{tab.MAE_wo_liu}. 

\begin{table}[h]
	\centering
	\begin{tabular}{|c|c|c|c|c|}
		\hline
		\multirow{2}{*}{\textbf{\begin{tabular}[c]{@{}c@{}}Usage\\ of\\ Liu-method\end{tabular}}} & \multirow{2}{*}{\textbf{\begin{tabular}[c]{@{}c@{}}Number of\\ training \\ simulations\end{tabular}}} & \multicolumn{2}{c|}{\textbf{\begin{tabular}[c]{@{}c@{}}Queue estimation\\ MAE\\ (MAD)\end{tabular}}}                 & \textbf{\begin{tabular}[c]{@{}c@{}}nVehSeen\\ MAE\\ (MAD)\end{tabular}} \\ \cline{3-5} 
		&                                                                                                       & \textbf{DL model}                                        & \textbf{Liu-method}                                       & \textbf{DL model}                                                       \\ \hline
		\textbf{w/ Liu}                                                                           & \textbf{125}                                                                                          & \begin{tabular}[c]{@{}c@{}}9.4 m \\ (5.3 m)\end{tabular} & \begin{tabular}[c]{@{}c@{}}18.1 m\\ (13.7 m)\end{tabular} & \begin{tabular}[c]{@{}c@{}}1.6 veh\\ (0.4 veh)\end{tabular}             \\ \hline
		\textbf{w/o Liu}                                                                          & \textbf{125}                                                                                          & \begin{tabular}[c]{@{}c@{}}9.0 m\\ (5.4 m)\end{tabular}  & \begin{tabular}[c]{@{}c@{}}18.1 m\\ (13.7 m)\end{tabular} & \begin{tabular}[c]{@{}c@{}}1.4 veh\\ (0.4 veh)\end{tabular}             \\ \hline
	\end{tabular}
	\caption[Estimation results (MAE) for model with and without Liu-method as input] {Estimation results (MAE) for model with and without Liu-method as input.} \label{tab.MAE_wo_liu}
\end{table}

Only considering the MAEs, the no-Liu model seems to be slightly better. Furthermore, there is no difference for the MADs between both models. However, this data do not represent the performance regarding the stability for the estimation of nVehSeen. The Liu-method used as input feature prevents the model from overfitting, which results in a more stable behavior of nVehSeen. In this case, prior knowledge about the traffic is used, to avoid overfitting and guide the model in the right direction. In this case, to add prior knowledge has similar effects than classic regularization techniques like dropout or $L^2$-Regularization. It would be also possible to avoid overfitting by using more training data, but in reality traffic data are rare and expensive. So it is beneficial for the learning process to have preprocessed data by the Liu-method as input, which saves some effort for data mining and computational cost for training. 

Summarized, the performance of the deep learning approach is about twice as good as the performance of the conventional Liu-method measured in MAE. Especially in complex traffic situations, in this case lane changing or stop-and-go traffic, the Liu-method is not working reliable anymore and the results are under- or overestimating with an percentage error of around 50\% to 70\%. 
The geometric deep learning model is able to handle these complex traffic situations and has overall a better performance which results in lower MAEs. For the estimation of nVehSeen (number of vehicle on the entire lane) there is a certain warm-up phase necessary. Furthermore, the graph has the tendency to smooth the curve out for a higher simulation time. Very important is to train the neural network with a lot of data with different traffic arrival rates. Otherwise, it is possible that instability occurs. In this chapter it was also shown, that the usage of the Liu-method as input for the DL model is beneficial for the stability of the estimation.
The good performance of the geometric deep learning model, especially in complex traffic situations, makes the estimation results eligible for usage by intelligent traffic control algorithms. Therefore it is necessary to calculate the traffic pressure based on the estimated queue length (see sec. \ref{sec.pressure}).
\chapter{Conclusion} \label{sec.conclusion}

In this chapter the results presented in section \ref{sec.DL_results} are discussed and evaluated. Following, further investigations are suggested. 

\section{Summary and Evaluation of the Results}

In this work the goal was to present an estimation method that is robust to complex traffic situations such as excess demand, lane changing or stop-and-go traffic, but still uses common inductive loop detectors only. As preliminary work, the method of Liu \cite{liu_real-time_2009} was implemented, which uses only second-by-second loop detector data from the last three traffic cycles to estimate the queue length in front of signalized intersections, based on the detection of shockwaves in the queue. After the implementation of both conventional models (basic and expansion model), presented in chapter \ref{sec.Implementation_liu}, the expansion model was chosen for further investigation with the deep learning approach, since it has better performance on second-by-second loop detector data. While the Liu-method has good performance under optimal test conditions at a single intersection, the performance suffers under more realistic traffic scenarios in a grid-road network, as shown in sec. \ref{sec.model_grid}. In particular, the effects of lane changing and stop-and-go traffic were studied. In case of oversaturation (long queue caused by excessive demand), the performance of the Liu-method degrades as well. For the case of lane changing behind the advanced detector and before the stop bar loop detector, the right lane is usually overestimating while the middle and left lane are underestimating. Furthermore, the effect of stop-and-go traffic leads to perturbation shockwaves in the queue, which causes a too early detection of the change in traffic pattern. The stop-and-go shockwave is misidentified as continuous discharge shockwave, which leads to underestimation.

To overcome these limitations, a new approach with geometric deep learning was examined, presented in chapter \ref{sec.geometric_deep_learning}. The road network topology was abstracted as a graph representation, consisting of nodes and edges. Based on this abstracted graph, a neural network layer called the Graph Attention Layer \cite{velickovic_graph_2017}, introduced in sec. \ref{sec:gat_layer}, uses attention mechanisms to find spatial correlations between the data from single lanes. Furthermore, an encoder-decoder structure (see sec. \ref{sec:enc-dec}) was used to solve the task as a  (time) sequence-to-sequence problem, where the input time sequence from the previous layers gets encoded into hidden states and then decoded again into the output time sequence. In addition to that a temporal attention mechanism was implemented which searches for temporal attention between the current and previous time steps. 
The geometric deep learning approach can handle complex traffic situations better compared to the Liu-method, which results in a more accurate estimation, shown in sec. \ref{sec.DL_results}. The average MAE (mean average error) for all lanes is usually only half as high as the MAE of the Liu-estimation. In situations in which the Liu-method does not perform well, such as lane changing or stop-and-go traffic, the deep learning approach does not suffer. This is because the deep learning approach is able to learn more complex traffic correlations due to spatial and temporal attention mechanisms, which makes this approach superior.  

For the measurement of the performance the mean average error (MAE) was chosen. The mean average percentage error (MAPE) is not a reliable metric to evaluate the performance of the estimation, since it becomes very high when the ground-truth data is almost zero but the estimation not. The MAE is reliable in every case and still good to interpret, since it provides an absolute value (in this case in meters or vehicles). 

Furthermore, the robustness of the deep learning model regarding traffic simulations under different arrival rates was tested in section \ref{sec.DL_robust}. The arrival rate determines how many vehicles arrive on the lane during a certain period of time. When the (statistical) model is not trained on arrival rates where it is tested on, poorly performance can occur. The deep learning model cannot handle situations it has never seen before during training. But in this thesis it was shown that if the model is trained on a high diversity of arrival rates, the number of instabilities can be reduced and the performance can be improved. So a high diversity as well as a high amount of data are beneficial for the generalization of knowledge. 

Finally, it was studied in sec. \ref{sec.Liu_as_input} whether the estimation from the Liu-method is beneficial as an input feature for the geometric deep learning model. The results suggest that including prior knowledge in from of the Liu-method makes the estimations of the deep learning model more stable regarding unrealistic drops of the estimated number of vehicle on the lane below zero and has a similar effect as a regularization mechanism, which avoids overfitting. It is hypothesized that the usage of the Liu-method decreases the amount of necessary training data and saves also computational cost during training. This result justifies the effort for implementation of the Liu-method as supplemental work to the geometric deep learning approach. 

Based on the accurate estimation of the queue length by the geometric deep learning model, the traffic pressure can be calculated directly as presented in section \ref{sec.pressure}. The traffic pressure could then be used as input for an intelligent traffic control algorithm to reduce congestion.

\section{Further Investigations}

\begin{figure}[h]
	\centering
	\includegraphics[width=0.8\textwidth]{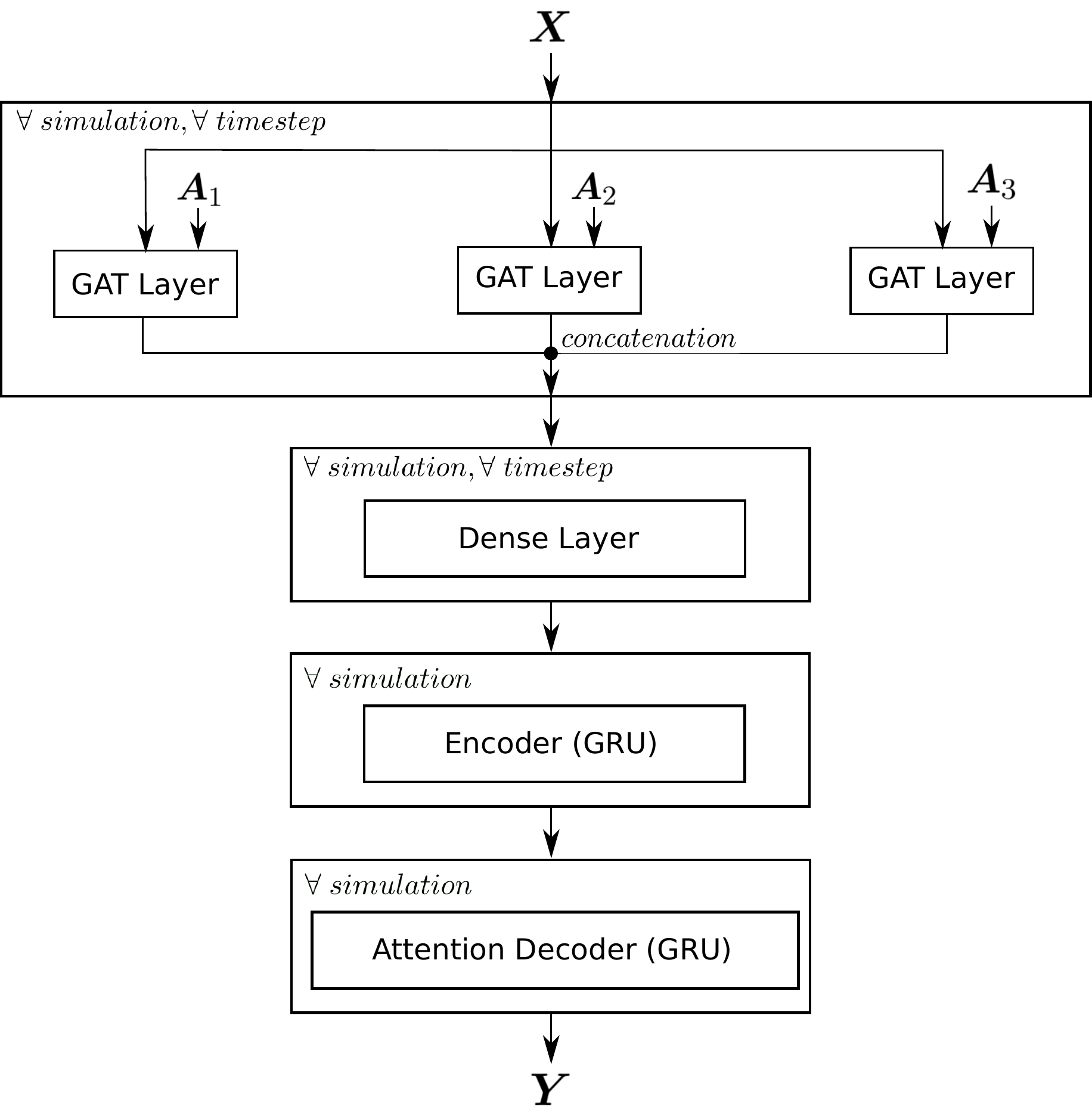}
	\caption[Possible approach for implementation of different spatial correlations] {Possible approach for implementation of different spatial correlations.} \label{bild.multi_channel_GAT}
\end{figure}

Although the results have shown that it is possible to achieve satisfying performance due to geometric deep learning approaches, there is a lot more potential to improve the accuracy. 
For the approach in this work only one kind of spatial correlation is considered. It is not distinguished between upstream or downstream connections, for instance. But a lot of different correlations in a road network exist, like lane changing options, left turn, right turn and straight connections, upstream or downstream connection and so forth. Generally, a lot of different adjacency matrices $\boldsymbol{A}$ can be created and used to learn even more complex correlations in traffic behavior. This would also increase the benefit of the GAT layer even more. Several adjacency matrices have to be build and the GAT layer calculates for each of them the spatial attention separately. One possible solution for this approach is visualized in fig. \ref{bild.multi_channel_GAT}.
The adjacency matrices $\boldsymbol{A}_1$, $\boldsymbol{A}_2$ and $\boldsymbol{A}_3$ have all different graphs from the road network abstracted as source. For instance, $\boldsymbol{A}_1$ contains all the upstream connections, $\boldsymbol{A}_2$ all the downstream connections and $\boldsymbol{A}_3$ the neighboring lanes for lane changing. These informations are processed in three parallel GAT layer and the results can be concatenated. It is expectable that the performance, especially in complicated traffic situations, increases. It may also be possible to process different adjacency matrices in one GAT layer by providing $\boldsymbol{A}$ not as a matrix but as a three dimensional array with the first dimension as the number of different connection types. 
Another interesting point to investigate would be, if the traffic light signals could be represented in form of changing adjacency matrices. The adjacency matrix, or generally the abstracted graph from a road network, represents the lanes which are connected with each other. If the traffic light is green for a particular lane, the lane is connected with other lanes and cars can exchange. If the traffic light is red, no connection is temporarily allowed and the graph (with the adjacency matrix) changes. So it is possible to feed in an adjacency matrix for every time step, which depends on the traffic light data. This might be potential superior to providing a binary signal for each lane, as it is done in this approach. 
Another point, which is also of high interest, is to combine both spatial and temporal attention. In this work, the spatial attention, calculated by the GAT layers, is only considering one time step (or time slice) to calculate the attention coefficients based on the connected lanes. But usually, what happens in the past on the connected lanes is relevant for the current time step on the specific lane. For example if a lot of vehicles arrive in the previous cycle on the upstream connected lane, it is very likely that a lot of this vehicles will arrive on the specific lane in the current cycle. So it is necessary to search for spatial attention under consideration of the previous time steps. 

There are a lot of challenging but also exciting tasks, that have to be solved. But considering, that the attention mechanisms as well as sequence-to-sequence problems are relatively new in the field of deep learning, the possibilities and potential are definitely worth the effort.
%-> implement different kind of spatial correlations (neighbor, upstream downsream; or left, right, straight lane). Show also little graph of possible topology

%-> rather that have parallel gat layer, it could be also possible to to use multi batches for A matrix

%-> it is also possible to include the tls information in the A matrix which changes then every traffic light cycle

%-> develop attention algorithm which combines spatial and temporal attention, since the gat layer gets just data for one timesclice

%-------------------------------------------------------------------------------
% Hier wird automatisch das Literaturverzeichnis erstellt und der Anhang
% (Datei anhang.tex) eingebunden.
%-------------------------------------------------------------------------------
\abspann

\end{document}